\UseRawInputEncoding
\pdfoutput=1
\documentclass[11pt,twoside]{report}
\usepackage{a4wide,caption,epsfig,fancyhdr,url}
\usepackage{subcaption}
\usepackage{graphicx}
\usepackage{enumitem}

\usepackage[pagebackref]{hyperref}

\graphicspath{{./Figs/}}
\newcommand{\SubmissionDate}{\today}
\newcommand{\student}{Tyler Saxton}
\newcommand{\supervisor}{Dr. Dhirendra Singh}
\newcommand{\topic}{Mapping suburban bicycle lanes using street scene images and deep learning}
\newcommand{\school}{School of Computing Technologies}
\newcommand{\program}{Master of Data Science}
\newcommand{\institution}{Royal Melbourne Institute of Technology}

\begin{document}

\title{{\Large\bf \topic}}
\author{
A minor thesis submitted in partial fulfilment of the requirements for the degree of
\\\program\\*[10mm]
\\\student
\\\school
\\Science, Engineering, and Technology Portfolio,
\\\institution
\\Melbourne, Victoria, Australia
}
\maketitle
\thispagestyle{empty}

\chapter*{Declaration}

This thesis contains work that has not been submitted previously, in
whole or in part, for any other academic award and is solely my
original research, except where acknowledged.

This work has been carried out since March 2021, under the
supervision of {\supervisor}.

\paragraph{}
\vspace{5cm}\noindent \\\student \\
\school\\
\institution\\
\SubmissionDate

\pagenumbering{roman}

\chapter*{Acknowledgements}

First and foremost, I would like to thank Dr. Dhirendra Singh for inspiring this research and supervising me throughout the year.  I also greatly appreciate the input provided by Dr. Ron van Schyndel and the ``Research Methods'' class of Semester 1 2021, as I worked to develop a detailed research proposal. \\

To Dr. Sophie Bittinger, Dr. Logan Bittinger, Amy Pendergast, Laura Pritchard, thank you for your encouragement, and your assistance with the editing process.

\chapter*{Summary}

Many policy makers around the world wish to encourage cycling, for health, environmental, and economic reasons.  One significant way they can do this is by providing appropriate infrastructure, including formal on-road bicycle lanes.  It is important for policy makers to have access to accurate information about the existing bicycle network, in order to plan and prioritise upgrades.  Cyclists also benefit when good maps of the bicycle network are available to help them to plan their routes.  This thesis presents an approach to constructing a map of all bicycle lanes within a survey area, based on computer analysis of street scene  images sourced from Google Street View or ``dash cam'' footage.

\chapter*{Abstract}

On-road bicycle lanes improve safety for cyclists, and encourage participation in cycling for active transport and recreation.  With many local authorities responsible for portions of the infrastructure, official maps and datasets of bicycle lanes may be out-of-date and incomplete.  Even ``crowdsourced'' databases may have significant gaps, especially outside popular metropolitan areas.  This thesis presents a method to create a map of bicycle lanes in a survey area by taking sample street scene images from each road,  and then applying a deep learning model that has been trained to recognise bicycle lane symbols.  The list of coordinates where bicycle lane markings are detected is then correlated to geospatial data about the road network to record bicycle lane routes.  The method was applied to successfully build a map for a survey area in the outer suburbs of Melbourne.  It was able to identify bicycle lanes not previously recorded in the official state government dataset, OpenStreetMap, or the ``biking'' layer of Google Maps.


\tableofcontents
\listoffigures
\listoftables

\chapter{Introduction}
\pagenumbering{arabic}

The benefits of ``active transport'', such as walking and cycling, have been well documented in previous studies.  Participants' health may improve due to their increased physical activity.  There are environmental benefits due to reduced emissions and pollution.  And there are economic benefits, including a reduced burden on the health system, and reduced transportation costs for participants \cite{LEE2012219} \cite{RABL2012121}.

Federal and state government policy makers in Australia therefore wish to encourage cycling \cite{federal_policy_2019} \cite{state_policy_2020}.  However, the share of cycling for trips to work in Melbourne is only 1.5\% \cite{melbactive}.  For many commuters, a perceived lack of safety of cycling is a major barrier to adoption.  Other significant factors are the availability of shared bicycle schemes and storage facilities, and the risk of theft \cite{WILSON2018234}.  Cycling infrastructure has a significant impact on real and perceived cyclist safety, and this research project will focus on that issue.  Important safety factors include the presence and width of a bicycle lane, the presence of on-street parking, downhill and uphill grades, and the quality of the road surface \cite{BIKESAFETY} \cite{Teschke2012}.  A comprehensive dataset of cycling infrastructure would help policy makers  prioritize areas in need of improvement to safety.

In Victoria, Australia, the state government publishes a ``Principal Bicycle Network'' dataset to assist with planning \cite{PrincipalBicycleNetwork}, however, during an exploration of the dataset, it was found to be significantly out of date.   (See section \ref{s:datasets}.)  Individual Local Government Areas may produce their own maps of bicycle routes, but availability is inconsistent \cite{vicroads_maps}.

The aim of this research project was to construct a dataset or map of bicycle lanes in a local area, by collecting street scene images at known coordinates, and then using a ``deep learning'' model to detect locations where bicycle lane markings are found.  Detection locations were matched to an existing dataset of roads in the area, to infer bicycle lane routes along stretches of road where markings were consistently found.  If a baseline map of bicycle lane routes can be built this way, then the process could be extended in future to gather information about other significant factors, such as how frequently the bicycle lane is obstructed by parked vehicles, or the presence of debris or damage to the road surface.

Google Street View has been chosen as a source of street scene image data due to its wide geographical coverage, and the accessibility of the data via a public API.  However, a significant limiting factor is that the Google Street View images for any given location might be several years out of date.  Therefore, the use of images collected from a ``dash cam'' was also explored.  A local government that is responsible for building and maintaining bicycle lanes could use dash cameras to gather its own images, at regular intervals, for more up-to-date data.

\section{Research questions}
\begin{itemize}
\item{RQ1: Can a ``deep learning'' model be used to identify bicycle lane markings in street scene images sourced from Google Street View, in at least 80\% of images where those markings appear, with a precision of at least 80\%?}
\item{RQ2: Can the model then be used to correctly detect and map all bicycle lane routes across all streets in a survey area with Google Street View coverage?}
\item{RQ3: Can a similar process be applied to correctly detect and map bicycle lane routes from street scene images collected from dash camera video footage in a survey area?}
\item{RQ4: Can the approach used to map bicycle lane routes from dash camera video footage be re-used to visually survey other details about the infrastructure?}
\end{itemize}

Models from the TensorFlow 2 Model Garden were first trained to detect bicycle lane markings in a custom dataset of Google Street View images.  These initial models generalised well enough to detect bicycle lane markings in dash camera footage, so they were used to gather additional images from the dash camera to include in the dataset for further training and validation.

The remainder of this thesis is laid out as follows:  Chapter \ref{s:literature} is a literature review, providing motivations for the research, and a background in relevant techniques;  chapter \ref{s:methods} describes the approach that was taken;  chapter \ref{s:results} reports the results and discusses the limitations and findings;  chapter \ref{conclusion} discusses the overall conclusions of the research.  Please see Appendix \ref{a:process} for instructions on how to access and use the code and data assets produced as part of this research project.

\chapter{Literature Review}
\label{s:literature}

\section{Motivating literature}

Prior research has clearly shown health, economic, and environmental benefits from active transport.  Lee et al., 2012 \cite{LEE2012219} analysed World Health Organization survey data from 2008, and showed that physical inactivity significantly increased the relative risk of coronary heart disease, type 2 diabetes, breast cancer, colon cancer, and all-cause mortality, across dozens of countries.  Rabl \& de Nazelle, 2012 \cite{RABL2012121} demonstrated that active transport by walking or cycling improves those relative risks for participants.  Moderate to vigorous cycling activity for 5 hours a week reduced the all-cause mortality relative risk by more than 30\%.  They estimated an economic gain from improved participant health and reduced pollution, offset slightly by the cost of cycling accidents.

In Australia, federal and state governments are committed to the principle of supporting active transport through the provision of cycling infrastructure, declaring their commitment through public statements on their official websites \cite{federal_policy_2019} \cite{state_policy_2020}.  Many other governments around the world have adopted similar policies.

Taylor \& Thompson, 2019 \cite{melbactive} surveyed the use of active transport in Melbourne, to establish a baseline of current commuter behaviour.  They found that cycling only accounted for 1.5\% of trips to work in the area.  It could therefore be argued that there is room for improvement.

Schepers et al., 2015 \cite{SCHEPERS2015460} produced a summary of literature related to cycling infrastructure and how it can encourage active transport, resulting in the aforementioned benefits.  The paper found that providing cycling infrastructure that is perceived as being safer does encourage participation.  Other papers such as Wilson et al., 2018 \cite{WILSON2018234} agreed.

Other researchers have examined which factors affect the perceived and actual safety of cycling routes, in a variety of settings.

Klobucar \& Fricker, 2007 \cite{BIKESAFETY} surveyed a group of cyclists in Indiana, USA, asking them to ride a particular route and rate the safety of each road segment along the route, then asking them to review video footage of other routes and rate the safety of those routes, too.  A regression model was created to predict the cyclists' likely safety ratings for other routes.  The creation of the model led to a list of road segment characteristics that were apparently most influential in the area.

Tescheke et al., 2012 \cite{Teschke2012} surveyed patients who attended hospital emergency rooms in Toronto and Vancouver in Canada, due to their involvement in a cycling accident.  Details of the circumstances of each accident were gathered, along with the outcomes.  The dataset was analysed to determine which factors increased (or decreased) the relative odds of a cyclist being involved in an accident.

Malik et al., 2021 \cite{Malik2021} modelled cyclist safety in Tyne and Wear County in north-east England, a more rural setting.

The factors that contribute to cyclist safety vary by locality.  For example, cyclists in one city might be concerned by the hazard of tram tracks, whereas this might not be a relevant concern in another city, or a less built-up area.  The common themes among the aforementioned papers were: the presence and width of a bicycle lane;  the presence of on-street parking; downhill or uphill slopes;  the volume, speed and profile of motor vehicle traffic; the quality of the road surface;  lighting;  and the presence of construction work.

Most of these factors can be influenced by infrastructure and road design.  Therefore, it would be valuable to quantify as many of these factors as possible in a dataset, to assist policy makers in deciding what changes to make, and where, to provide a safer network of cycling infrastructure.

\section{Relevant literature in machine learning}

\begin{figure}[h]
\centering
\includegraphics[scale=0.60]{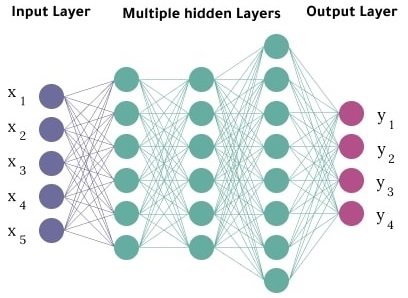}
\caption{Example deep learning model, showing the multiple hidden layers that distinguish it from a ``non-deep'' neural network.  Each arrow in the diagram represents a connection with a ``weight''.  Weights are set and refined during the training process, via a ``backpropagation'' algorithm.  Source: Lofqvist et al., 2020 \cite{diagram}, reproduced under Creative Commons licence.}
\label{fig:deep_learning_diag}
\end{figure}

``Deep Learning'' is a paradigm by which computational models can be constructed to tackle many problems, including visual object recognition.  A Deep Learning model can be trained to perform visual object recognition tasks by supplying it with a ``training'' dataset of images where the objects it must recognise have been pre-labelled.  During training, the model processes its training dataset over and over again, and with each iteration the weights in its multi-layer neural network are refined through the use of a ``backpropagation'' algorithm \cite{deeplearning}.  With a sufficient number of training ``epochs'', the model hopefully develops the ability to detect if and where the objects of interest appear in each image, to an acceptable level of performance.  See figure \ref{fig:deep_learning_diag}.

Tensorflow \cite{TENSORFLOW2016A} \cite{TENSORFLOW2016B} and PyTorch \cite{pytorch} are two dominant frameworks for building, training, evaluating, and applying Deep Learning models.  Within these frameworks, researchers have progressively developed new models for visual object recognition, typically with a focus on either improving the accuracy of the results, increasing speed to allow ``real time'' processing of video streams, or creating models that will work on low-cost hardware.  Significant Deep Learning models considered within this research project include: R-CNN, proposed by Ren et al., 2015 \cite{REN2016};  SSD, proposed by Liu et al., 2016 \cite{ssd};  ResNet, proposed by He et al., 2016 \cite{He_2016_CVPR}; YOLO and subsequent variants, first proposed by Redmon et al., 2016 \cite{YOLOv1}; MobileNetV2, proposed by Sandler et al., 2018 \cite{MobileNetV2}; a nd CenterNet, proposed by Duan et al., 2019 \cite{centernet}.

The Tensorflow 2 Model Garden \cite{zoo} provides access to many of these models, pre-trained on the COCO 17 (``Microsoft COCO: Common Objects in Context'') dataset.  It therefore provides a convenient library of Deep Learning models that have already received a significant amount of training for the general problem of visual object recognition, and can be re-used to recognise new classes of objects through a process of ``transfer learning'' \cite{coco} \cite{transferlearning}.

Semantic image segmentation of street scene images is an important and active area of computer science research.  ``Deep Learning'' models that can understand sequences of images in real time are essential for self-driving vehicles or similar driver-assistance systems, so many papers have focussed on that area.  In order to train a Deep Learning model that specialises in understanding street scene images, a labelled dataset of street scene images is required.  Papers by Cordts et al., 2016 \cite{Cordts_2016_CVPR} and Zhou et al., 2019 \cite{ade20k} announced the publication of the ``Cityscapes'' and ``ade20k'' image datasets, respectively.  These datasets each contain many street scene images, with objects of interest labelled in a format suitable for training deep learning models to understand similar on-road scenarios.  Many papers have used these datasets in order to train new machine learning models.  Chen et al., 2018 \cite{DEEPLAB} is one example of a heavily-cited paper where ``CityScapes'' data has been used to train a ``real time'' model to understand street scene images.  Unfortunately, neither of the datasets has cycling infrastructure labelled.  The ``CityScapes'' dataset has labelled bicycle lanes under the ``sidewalks'' category.  This is useful to train a car not to drive there, but a cyclist might not be legally allowed to ride on a sidewalk designed for pedestrian traffic.  

In another branch of research, deep learning tools have been used to manage roadside infrastructure and assets.  Campbell et al., 2019 \cite{CAMPBELL2019101350} used an ``SSD MobileNet'' model to detect ``Stop'' and ``Give Way'' signs on the side of the road in Google Street View images, to help build a database of road sign assets.  An application called ``RectLabel'' was used to label 500 sample images for each type of sign.  Photogrammetry was used to estimate a location for each detected sign, based on the Google Street View camera's position and optical characteristics and the bounding box of the detected sign within the image.  Zhang et al., 2018 \cite{s18082484} performed a similar exercise, detecting road-side utility poles using a ``ResNet'' model.

In Australia, the ``Supplement to Australian Standard AS 1742.9:2000'' sets out the official standards for how bicycle lanes must be constructed and marked.  Generally, a lane marking depicting a bicycle should be painted on the road inside the bicycle lane within 15 metres before and after each intersection, and at intervals of up to 200 metres\cite{standards}.  Therefore, it is proposed that a Deep Learning model could be trained to detect these markings, similar to how Campbell et al., 2019 \cite{CAMPBELL2019101350} trained a model to detect road signs, and Zhang et al., 2018 \cite{s18082484} trained a model to detect utility poles, using pre-trained models from the TensorFlow 2 Model Garden \cite{zoo} as a starting point.

The use of satellite imagery was also considered.  Li et al., 2016 \cite{ROADNETWORK} showed that a road network could be extracted from satellite imagery using a convolutional neural network (CNN), and this is particularly useful in rural areas where maps are not already available.  However the resolution of publicly available satellite image data would not be sufficient to identify a bicycle lane on a road, and it would not be able to distinguish a standard bicycle lane from a paved shoulder.  Satellite imagery may have a role to play in detecting off-street bicycle tracks in ``green'' parkland area, but it was decided to exclude off-street routes from the scope of this research, and focus on on-road infrastructure.

Aerial photography may be useful, where it is available with sufficient detail.  However it would require a different data source for every jurisdiction.  Ning et al., 2021 \cite{NING2021} had success extracting sidewalks from local aerial photography using a ``YOLACT'' model.  Areas of uncertainty caused by tree cover were filled in using Google Street View images.  The model appeared to rely on a concrete sidewalk having a very different colour to the adjacent bitumen road.  The approach might not be able to distinguish bitumen bicycle lanes.

Aside from formal bicycle lanes, cyclist safety can also be improved by a paved shoulder, or by creating lanes that are wide enough to safely accommodate a vehicle passing a cyclist.  It may be possible to detect and map these arrangements by recognising lane markings and road boundaries.  A common method of detecting lane boundaries is through the combination of a Canny edge detector \cite{canny} and a Hough transformation \cite{hough}.  The OpenCV library provides a frequently used implementation \cite{opencv}.  If the Canny-Hough approach struggles with a poorly defined road boundary or ``noise'' from roadside objects, a Deep Learning approach may help:  Mamidala et al., 2019 \cite{8929655} successfully used a ``CNN'' model to detect the outer boundary of roads in Google Street View images.

During the literature review, one other paper was identified where the authors had applied Deep Learning techniques in the domain of cyclist safety:  Rita, 2020 \cite{rita_2020} used the ``MS Coco'' and ``CityScapes'' datasets to train ``YOLOv5'' and ``PSPNet101'' models to identify various classes of object (Bicycle, Car, Truck, Fire Hydrant, etc.) in Google Street View images of London.  A matrix of correlations between objects was calculated.  This was used to infer the circumstances where cyclists might feel the most safe.  For example, there was a high correlation between ``Person'' and ``Bicycle'' which ``suggests pedestrians and cyclists feel safe occupying the same space''.  This is research did not involve creating maps.

\section{Cycling infrastructure datasets and standards}
\label{s:datasets}

The Victorian State Government in Australia started publishing an official ``Principal Bicycle Network'' dataset on \url{data.gov.au} in 2020 \cite{PrincipalBicycleNetwork}.  It is an official dataset that is intended to help policy makers with their planning.  It includes ``existing'' and ``planned'' bicycle routes.  The dataset only covers formal bicycle lanes, so it generally excludes roads with a simple paved shoulder.  In some country areas, the routes did not appear to be formally marked as standard bicycle lanes.  It was found that some existing bicycle lanes are still marked as ``planned'' in the dataset, or not listed at all.    There is a field to record when each entry was last validated, but the most recent timestamp was in 2014.  The data appears to be incomplete and out of date.  Its scope is limited to the State of Victoria.

``OpenStreetMap'' \cite{OPENSTREETMAP} is a source of crowdsourced map data.  It provides detailed information about road networks worldwide, and the attributes of each road segment.  ``Cycleway'' tags can be used to mark not just where a bicycle lane is, but other interesting attributes, such as whether the cycleway is shared with public transport, whether there is a specially marked area for cyclists to stop at each intersection, and how wide the bicycle lane is.  Given that the data is crowdsourced, the quality and availability of the data may vary by location.

``Google Maps'' provides a ``biking layer'' \cite{bike_layer}.  This includes off-street bicycle paths and on-street bicycle lanes.  It only gives a ``yes'' or ``no'' opinion about whether a route is especially suited to bicycles, with no further information provided.  But that may be of assistance in scouting locations to use in a dataset of Google Street View images.

\chapter{Methods}
\label{s:methods}

This section describes the methodology that was used to address each of the research questions.  To assist with reproducibility, all code has been published on GitHub and FigShare, and the dash camera footage and data has been published on FigShare.  For detailed instructions on how to access the code and data, and how to run the code via the provided Jupyter Notebooks, please refer to appendix \ref{a:process}.

\section{RQ1: Training a model to identify bicycle lanes in Google Street View images}
\label{s:rq1}

\begin{figure}[h]
\centering
\includegraphics[scale=0.550]{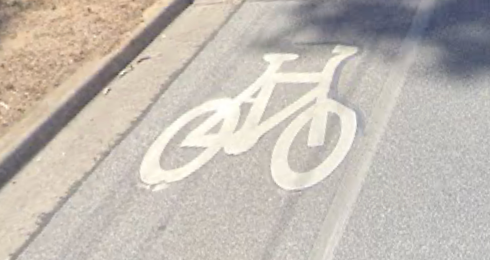}
\caption{Example bicycle lane marking.  Source: Google Street View, Oct 2019}
\label{fig:symbol}
\end{figure}

A review of the relevant Australian standards for bicycle lanes \cite{standards} suggested that the best way to search for bicycle lanes in street scene images would be to focus on the bicycle lane markings that are painted on the road surface.  See figure \ref{fig:symbol} for an example marking.  

This marking is universal across all standard bicycle lanes in Australia, and can distinguish a bicycle lane from other parts of the road.  It may sometimes appear on a green surface, and occasionally it may be partially occluded due to limited space or wear and tear.  Similar markings exist in many other jurisdictions outside Australia.

In order to train a deep learning model to recognise bicycle lane markings in street scene images, it was necessary to first create a dataset of labelled images for training and validation.  This could then be used to apply a variety of pre-trained models from the ``TensorFlow 2 Model Garden'' to the problem, through a process of transfer learning.

\subsection{Creating a labelled dataset of Google Street View images}

The official Australian design standards for bicycle lanes specify that bicycle lane markings should appear within 15m of each intersection \cite{standards}.  Therefore, intersections along known bicycle lane routes are locations where example images are likely to be found.

\subsubsection{Identifying candidate intersections to sample}
\label{s:sample_candidates}

Two possible sources of information about ``known'' bicycle lane routes were considered for use in the sampling process:  The ``Principal Bicycle Dataset'', and XML extracts from the OpenStreetMap database.  These datasets are discussed in section \ref{s:datasets}, with further details of the structure of the OpenStreetMap data provided in Appendix \ref{a:osm_concepts}.  The ``Principal Bicycle Network'' dataset was chosen, as the incumbent official dataset for local policy makers.  Due to the age of the data, it was also less likely to suggest bicycle routes not yet captured in the Google Street View images.  Using OpenStreetMap as a source of ``known'' bicycle lane routes would generalise to other jurisdictions, therefore tools were created to support both approaches, and their operation is described in appendix \ref{aj01} to \ref{aj03}.  Given a list of ``known'' bicycle lane routes from either of these sources, each route can be looked up in an OpenStreetMap database to identify its intersections.

OpenStreetMap data can be downloaded for an individual country via the ``GeoFabrik'' website \cite{geofabrik} and then sliced into a smaller area, if necessary, using the ``osmium'' command line tool \cite{osmium}.  For more background on concepts in the OpenStreetMap data, please refer to Appendix \ref{a:osm_concepts}.  Briefly, an OpenStreetMap XML extract file contains ``ways'', which represent lines on a map, such as road segments, and ``nodes'', which represent points with latitude/longitude coordinates that make up the ``ways'' and thereby describe the shape of each line on a map.  A list of candidate intersections along ``known'' bicycle lane routes can be extracted from OpenStreetMap XML data by looking for ``nodes'' that are common to two or more ``ways'' that are roads and have distinct names.

Generating a list of intersections from the ``Principal Bicycle Network'' dataset involves additional work, because it only lists bicycle lane routes, and does not have any information about intersecting roads.  Each point in the dataset is ``reverse-geocoded'' to find the name of the road using an API call to a local instance of a ``Nominatim'' server, as described in appendix \ref{a:nominatim}.  An OpenStreetMap XML extract for Victoria is then searched for all ``ways'' with a matching name, to find the corresponding road segments in the OpenStreetMap data.  The bounding box around each ``way'' is compared to the bounding box of the ``Principal Bicycle Network'' route, to ensure they roughly overlap, and remove matches for other roads with the same name, in a completely different area.  Once matching ``ways'' have been found for the route, they can be checked for intersections as per the OpenStreetMap process.

For every intersection identified, it is useful to calculate a ``heading'' for the road at the intersection, to help gather Google Street View images that are aligned to the direction of the road.  The Python ``geographiclib'' library can calculate a heading from one point to the next.  It is assumed that the heading at each intersection is the average of the heading from the previous ``node'' and the heading to the next ``node''.

\subsubsection{Downloading sample Google Street View images for the dataset}
\label{s:sample}

Google provides an API for downloading Google Street View images of up to 640x640 pixels, at a cost of \$7.00 USD per 1,000 images.  Volume discounts are available, with a reduced cost of \$5.60 USD per 1,000 image for volumes from 100,001 to 500,000 images.  Beyond that, further volume discounts can be negotiated with Google's sales team \cite{gsv_billing}.

Google Street View images were cached locally, to avoid unnecessary costs when requesting an image that has previously been downloaded.

The Google Street View API allows images to be requested for a location (latitude/longitude) with a desired heading, field-of-view angle, and camera angle relative to the ``horizon''.  A Jupyter Notebook with an ``ipywidgets'' GUI was created to:  Randomly select a sample location from the list of candidate intersections in Victoria;  download four Google Street View images at 0, 90, 180, and 270 degrees relative to the heading of the road, with a 90 degree field of view, and the camera angled downwards 20 degrees towards the ground to focus on any nearby road markings;  allow the operator to press buttons to record which images should be included in the dataset due to the presence of an interesting feature.  Please see appendix \ref{aj04} for screenshots and detailed instructions on how to use the tool.

Bicycle lane markings are not always visible from the middle of an intersection, therefore, options were provided to allow the operator to move the camera up and down the road, 10 metres at a time, to get a better view of a marking that is just visible in the distance.

If the operator could not clearly tell that a white marking in an image was a bicycle lane marking, then the image was not included in the dataset.

The GUI allowed multiple candidate intersections to be assessed per minute.  An initial set of 256 images was collected over the course of approximately 4 hours, based on a random sample of points across the State of Victoria.

\subsubsection{Labelling the dataset}
\label{s:label}

Once a suitable number of Google Street View images containing bicycle lane markings had been collected, they were copied to a single folder and then labelled using the open-source tool ``labelImg'' \cite{labelImg}.  Please refer to appendix \ref{aj05} for more information on how to perform the labelling process.  The ``labelImg'' tool allows the operator to examine each image in a folder, draw bounding boxes around objects of interest in each image, and assign a label to each bounding box to categorise its contents.  The output of the process was one XML file per image, in a format that could be understood by TensorFlow 2 training tools and scripts.

\subsection{Training and evaluating candidate models from the TensorFlow 2 Model Garden}

Training and evaluation of candidate models was conducted on local infrastructure, as described in Appendix \ref{a:computer}.  TensorFlow was configured for GPU-accelerated development, using CUDA and cuDNN libraries from NVIDIA.

The dataset that was collected in section \ref{s:sample} and labelled in section \ref{s:label} was randomly split into ``training'' and ``test'' datasets according to an 80:20 ratio.  There were 205 images in the ``training'' dataset, and 51 images in the ``test'' dataset.

The Jupyter Notebook described in appendix \ref{aj06} was used to download a candidate pre-trained model from the TensorFlow 2 Object Garden and configure it to search for bicycle lane markings based on the collected dataset.  The Jupyter Notebook printed out commands that could be used from the command line, to train the model for a specified number of training steps against the ``training'' dataset, evaluate its performance against the ``test'' dataset, and create a ``frozen'' copy of the model as-at the current level of training.

Three significant factors were found to influence the performance of the model:  The size of the dataset available for ``training'' and ``test'' purposes;  the number of ``epochs'' or ``training steps'' spent training the model on the data;  and the ``confidence score '' threshold, from 0\% to 100\%, at which the model declares that a bicycle lane marking has been detected.

The initial size of the dataset was set based on previous experience in other papers such as Campbell et al., 2019 \cite{CAMPBELL2019101350}, where models had been successfully trained with around 500 images.  An initial dataset of 256 images was gathered, and further images were added later in response to model performance.

During the training process, the TensorFlow 2 training script reported ``loss'' performance metrics every 100 training steps.  This performance metric typically gradually improved as more training was conducted over the dataset, until a point of diminishing returns was reached.  Every 2000 training steps, the training was paused, and an evaluation script was run to test the performance of the model against the independent ``test'' dataset that had been withheld from the training process.  When training reached the point that further training steps did not significantly improve performance, training was halted.

The ``confidence'' threshold was adjusted manually, by reviewing the false positives and false negatives produced by the model, and adjusting it to provide a balance between over-confidence and conservatism.  As part of the evaluation process, copies of each image were written to disk with an overlay showing any detected bounding box and its confidence score.  These outputs were reviewed as part of the threshold tuning process.  The boundary boxes of detections were also checked to ensure that the results were sensible.

Multiple pre-trained models from the TensorFlow 2 Model Garden were trialled, with selections guided by their benchmark ``mAP'' (mean Average Precision) metrics for processing of the ``COCO 17'' dataset.  The process of building a map of bicycle lanes for an area can be considered a ``batch'' process, with no fixed time constraints, therefore mean Average Precision was prioritized over benchmark ``speed''.

The TensorFlow 2 Model Garden provided an efficient way to test multiple models via a single framework.  There are popular object detection models, such as the ``YOLO'' series of models, that are not supported by it.  For the purposes of this research, it was not considered necessary to perform an exhaustive search of all possible models across multiple frameworks.  A different model (e.g. YOLOv3) or a different framework (e.g. PyTorch) could be substituted if required.

Please refer to Appendix \ref{aj06} for instructions on how to run the training process.

\section{RQ2: Building a map of bicycle lane routes from Google Street View images in a survey area}
\label{s:rq2}

In section \ref{s:rq1}, a model was trained to detect bicycle lane markings in one street view image at a time.  To generate a map or dataset of bicycle lanes in an area, the first step is to determine a list of locations for which Google Street View images should be downloaded and processed by the model.  A batch process can then used to process each image and record whether or not a bicycle lane marking was found.  Finally, the list of sample locations where there was a ``hit'' must be correlated to geospatial data about the road network to infer routes, draw them on a map, and compare them to other sources.

\subsection{Sampling strategy for Google Street View images}
\label{s:rq2a}

In section \ref{s:sample}, it was noted that Google charges up to \$7.00 USD per 1,000 Google Street View images, and that the most likely place to find a bicycle lane marker is in the area immediately surrounding an intersection.  To reduce costs, a strategy of sampling only intersections and their immediate vicinity was adopted.  Samples could also be taken at regular intervals along each road, if initial results suggested that this was necessary to compensate for missed detections.

The Jupyter Notebook described in appendix \ref{a08} was used to load an OpenStreetMap XML extract file for the survey area into memory, and then traverse it to produce a batch file with the coordinates of every point within 20 metres of an intersection, at intervals of 10 metres.  The length of the batch file was inspected, to assess potential costs, and then the required images were downloaded via the Google Street View API, if they were not found in the local cache.

It was discovered that some intersections would be missing on roads at the very margins of the survey area, if the intersecting road existed entirely outside the bounds of the OpenStreetMap XML extract for the survey area.  A second extract, capturing bounding box with an extra margin of 200 metres around the survey area, was used to rectify the missing intersection issue.  The 200 metres margin was found to be sufficient to ensure that the adjoining road was included in the extract by the osmium tool.  Roads in the second extract were not considered part of the survey, but data about them was taken into account to ensure that intersections at the margin of the survey area were not missed.

Please see appendix \ref{a09} and \ref{a10} for detailed instructions on how to operate the tools that were used to generate a list of sample locations in a survey area.

\subsection{Applying a detection model to a batch of Google Street View images}
\label{s:rq2b}

\begin{figure}[h!]
\centering
\includegraphics[scale=0.2]{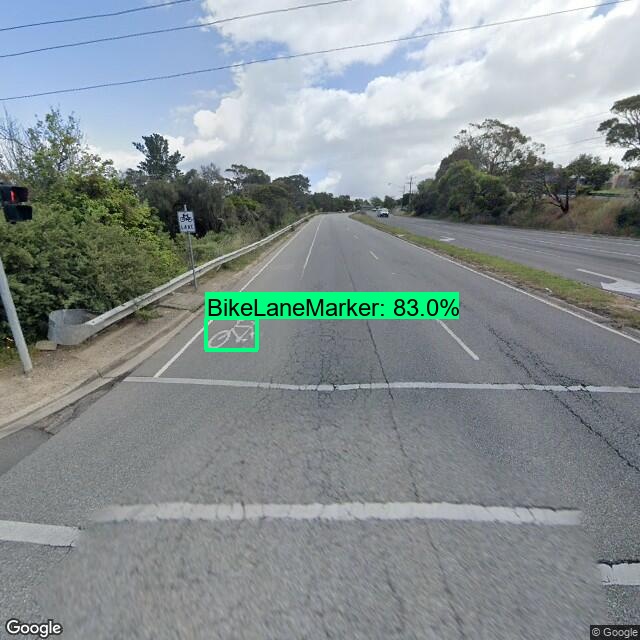}
\caption{Example bicycle lane marking with detection overlay.  Source: Google Street View, Oct 2019}
\label{fig:004a}
\end{figure}

The Jupyter Notebook described in appendix \ref{a10} was also used to apply the detection model to the sampled Google Street View images, and produce an output CSV with the coordinates of every point where a bicycle lane marking was detected.  If one or more bicycle lane markings were detected in an image, a copy of the image was saved to a folder of ``hits'' with a bounding box and confidence score overlay.  An example image with detection overlay is shown in figure \ref{fig:004a}.  If no bicycle lane marking was found, a copy of the image was instead placed into a ``miss'' folder.  Partitioning the result images into ``hits'' and ``misses'' allows the operator to quickly browse the images to check for any false positives or false negatives.

\subsection{Inferring bicycle lane routes from Google Street View detection points}
\label{s:rq2c}

The result of the batch process in section \ref{s:rq2b} is a list of locations where a bicycle lane marking was found, linked to the OpenStreetMap ``way'' IDs and ``node'' IDs that they were sampled from.  The next step is to use those detection points to infer contiguous routes that can be drawn as lines on a map.

Four Google Street View images were sampled at each sample point in the survey, providing a 360 degree view.  An approximate heading was calculated for each road at each intersection, but occasionally that did not appear to match the retrieved images.  Therefore, if a bicycle lane marking was detected at an intersection, it could be difficult to accurately decide which road it belonged to.

The proposed solution was to infer routes from the detection points based on the assumption that if markings were detected at two or more consecutive intersections along a named road, there is a bicycle lane along that segment.  A configuration option was provided to allow for one or more intersections with ``missing'' detections before assuming that the bicycle lane route has been interrupted.

If a street shared an intersection with a road that had a bicycle lane, it would not usually be reported as having a bicycle lane unless it had another intersection with a bicycle lane at the next intersection.

A continuous road may be represented in OpenStreetMap as  multiple ``way'' segments if any of its characteristics change from one segment to the next, such as a change of speed limit.  Before inferring bicycle lane routes along a road from the detection points, it was necessary to link adjacent ``way'' segments from a single named road back up into a single chain.  Otherwise, the option to allow for intersections with ``missing'' detections would not always work as intended.

Each inferred route was drawn as a line in a geojson file, which can be drawn on a map or measured as per section \ref{s:rq2d} below.

This approach was intended to draw a map of bicycle lane routes irrespective of which side of the road the bicycle lane is on.  If additional information is required about roads where bicycle lanes are only present on one side of the road, the process would need to be refined to take into account the heading of the road, the heading of the camera angle, and exactly where in the frame the bicycle lane marking was detected.

\subsection{Comparing results to other data sources}
\label{s:rq2d}

Once a geojson file has been produced to describe the detected bicycle lane routes, it can be drawn on a map in a Jupyter Notebook using the Python ``ipyleaflet'' library.  See appendix \ref{a11}.

The OpenStreetMap XML extract for the survey area can be filtered down to ``ways'' that are roads with the ``cycleway'' tag, and then converted into an equivalent geojson file.  The detected routes and the routes that are tagged in OpenStreetMap can be viewed side-by-side on a map.

\begin{figure}[h!]
\centering
\includegraphics[scale=0.5]{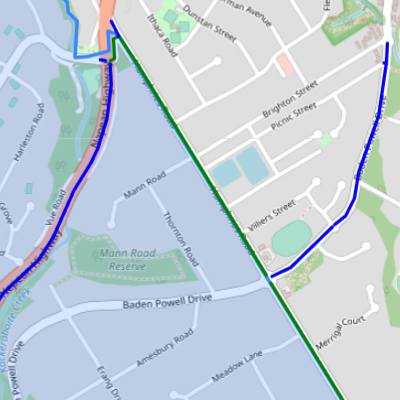}
\caption{Example map comparing detected bicycle lane routes to OpenStreetMap.  Green lines represent routes that are found in both maps.  Blue lines represent detected routes that are not recorded in OpenStreetMap.  Red lines would represent routes that are recorded in OpenStreetMap but not detected, however there are no such routes in this example.}
\label{fig:004}
\end{figure}

To allow more detailed comparisons, the geojson files were drawn simulaneously, in a single process.  For each road where a bicycle lane marking was detected, or tagged in OpenStreetMap, a process ``walks'' down ``nodes'' in order.  If the nodes are part of a ``way'' that has been tagged as a ``cycleway'', their coordinates are added to a line in the ``OpenStreetMap'' geojson.  If a detection point is encountered at two intersections in a row, then the intersection nodes and the nodes in between them are added to a line in the ``detected routes'' geojson.  At the same time, additional geojson files are drawn to show the route sections where both sources agree that there is a bicycle lane, and route sections where one source registers a bicycle lane and the other does not.

The total length of all routes in a geojson file can be calculated by using the Python ``geographiclib'' library to measure the distance between each pair of points along the route.  Therefore, the geojson files can not only be used to compare routes on a map, but to measure the similarities and differences in metres.

The ``Principal Bicycle Network'' dataset is available in a geojson format.  This allows a quick visual comparison with OpenStreetMap and the detected routes.  It is possible to align the ``Principal Bicycle Network'' dataset to the OpenStreetMap data to perform a detailed comparison with measurements, however this was not attempted.  The ``Principal Bicycle Network'' dataset had significant gaps in the survey areas, so significant that quantifying the differences in metres would not have been very meaningful.  Please see the results section \ref{results:rq2}.

\section{RQ3: Building a map of bicycle lane routes from dash camera footage captured in a survey area}
\label{s:rq3}

The use of Google Street View images for detection of bicycle lanes has both advantages and disadvantages to consider.  Google Street View image data is readily available for a wide area across many jurisdictions internationally, and it can be collected without a physical presence on location.  However, usage comes at a cost, which may be a significant factor if mapping is required across a large area.  The image resolution is limited to 640x640 per image, though perhaps this can be worked around by requesting many more images per location, each with a tighter field of view, at additional cost.  It appears that distinct images are only available every 10 metres.  The images may be several years out of date.  The position of the camera cannot be controlled, and it might miss service roads or the other side of a divided road.  With only one image available for each location, a bicycle lane marking might be obscured due to the presence of other traffic.

A local authority might prefer to gather their own footage, to address these shortcomings.  For the third research question, we explore whether bicycle lanes can be detected using dash camera footage instead, perhaps captured from vehicles that must regularly traverse the roads in the area to provide other services, such as waste disposal or deliveries.

\subsection{Choice of dash camera hardware}
\label{s:rq3a}

For these experiments, the ``Navman MiVue 1100 Sensor XL DC Dual Dash Cam'' was self-installed inside a sedan, at a cost of \$529 AUD.  This model was selected because it provided location metadata in NMEA files accompanying the MP4 video recordings.  For each minute of footage recorded, the device produced an MP4 file and a corresponding NMEA file.  The NMEA file provided the latitude, longitude, heading, and altitude at one second intervals.

Video footage was recorded at 60 frames per second, at 1920x1080 resolution.  Only images from the front-facing camera were used.

The camera features a ``wide angle'' lens to capture a good view of the road and its general surroundings, however optical specifications such as focal length were not available from the manufacturer's website.

\subsection{Gathering dash camera footage}
\label{s:rq3b}

On the 3rd of October, 2021, Melbourne set the record for the longest ``COVID lockdown'' in the world \cite{lockdown_record}.  Severe restrictions were placed on residents being more than 5km from home, gradually easing to 10km and then 15km for a period of many months during the course of the research\cite{lockdown_5km}.  Permitted reasons to leave home were restricted.  Therefore it was only possible to experiment with dash camera footage from a confined area.

Approximately 100 minutes of footage was gathered by driving within a local area around the outer metropolitan suburbs of Mount Eliza, Frankston, Langwarrin and Baxter in Melbourne, Victoria, Australia.  Not every road in the area was covered.  Roads were selected for inclusion in the footage to include all known bicycle lanes in the area, based on local knowledge and a review of the OpenStreetMap data.  A variety of other roads were included, to cover roads with and without a paved shoulder, major divided roads and small local roads, and a mix of residential and commercial areas.

The dash camera video footage and corresponding NMEA files were transferred to a computer and processed using the Jupyter Notebook described in appendix \ref{a12}.

For each pair of files, the NMEA file was loaded into memory, then the video file was split into a series of images.  The 60 frame per second footage was reduced to 5 frames per second by only sampling every 12th image.  Sampling images at a higher frequency than this did not appear to be necessary, due to the high similarity between adjacent frames.  Each extracted frame was assigned a location by extrapolating the readings from the two nearest measurements in the NMEA file.  An entry was then added to an output ``batch'' CSV file with the path to the image file, and its location metadata.

\subsection{Training the bicycle lane detection model}
\label{rq3c}

The dash camera footage was initially processed using the original model that was trained exclusively on Google Street View images in section \ref{s:rq1}, without any retraining based on dash camera images.  The original model was able to detect every bicycle lane marking in the test footage, however further training and refinement was required to address some performance issues:

\begin{itemize}
\item{Although the Google Street View-trained model was able to detect every marking during initial trials, it typically only did so when the marking was very close to the camera vehicle.  Further training was apparently needed on footage from the dash camera in order to allow it to detect markings that were further away, but still close enough to be clearly visible.}
\item{There were also issues with false positives that had not been apparent when using the Google Street View model on Google Street View images.  The model would sometimes yield false positives in the following situations:

	\begin{itemize}
	\item{Other white markings on the road, such as turning arrows, diagonal stripes to represent a traffic island, ``keep clear'' signs, or dashed lines advising traffic to ``give way'' at an intersection}
	\item{Pale road defects or other anomalies such as manhole covers}
	\item{Reflections from the bonnet of the camera vehicle}
	\item{Clouds or random objects well outside the boundary of the road}
	\end{itemize}
}
\end{itemize}

\begin{figure}[h]
\centering
\begin{subfigure}{0.3\textwidth}
	\includegraphics[width=\textwidth]{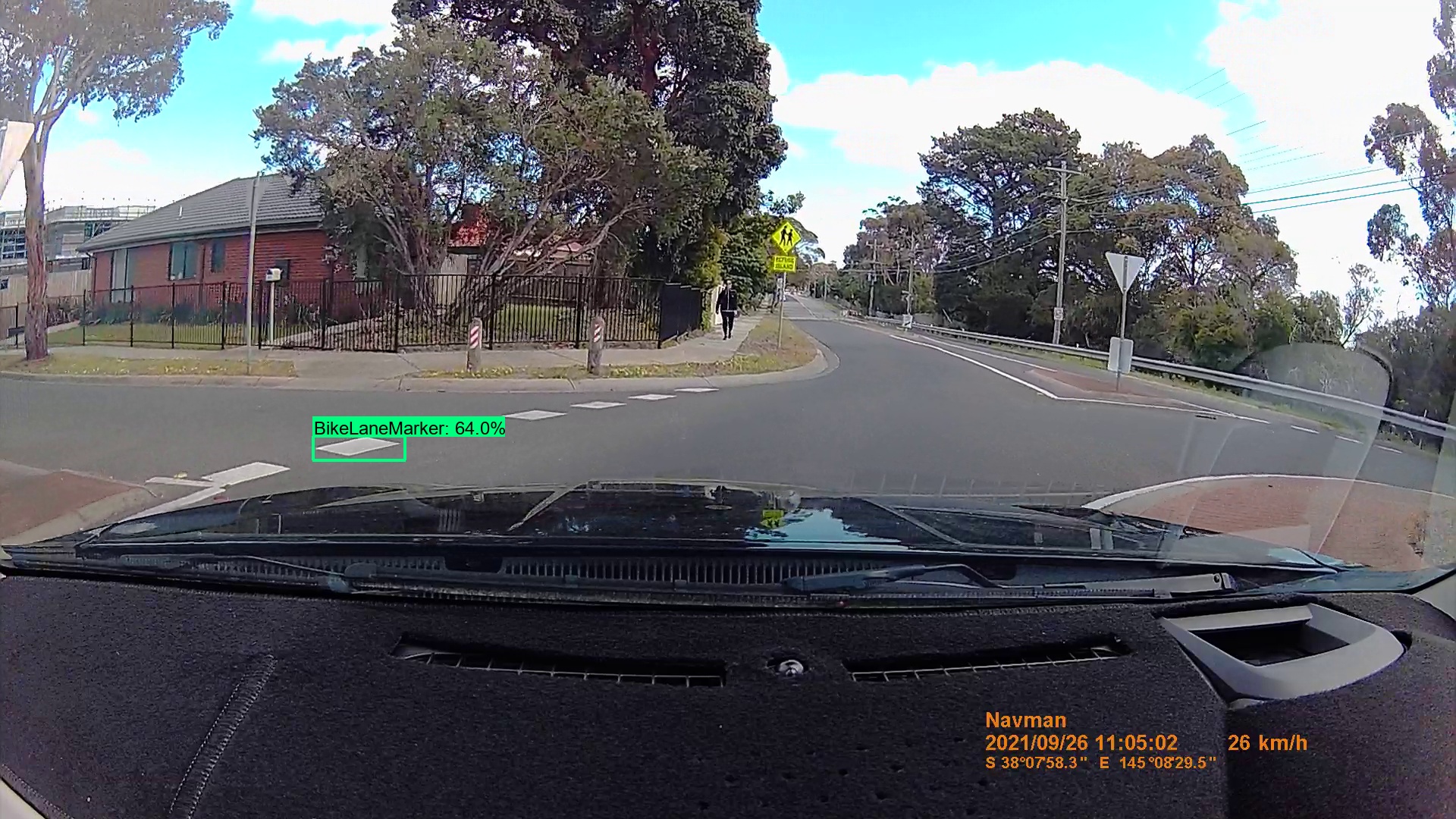}
	\caption{``Give Way'' stripes}
\end{subfigure}
\hfill
\begin{subfigure}{0.3\textwidth}
	\includegraphics[width=\textwidth]{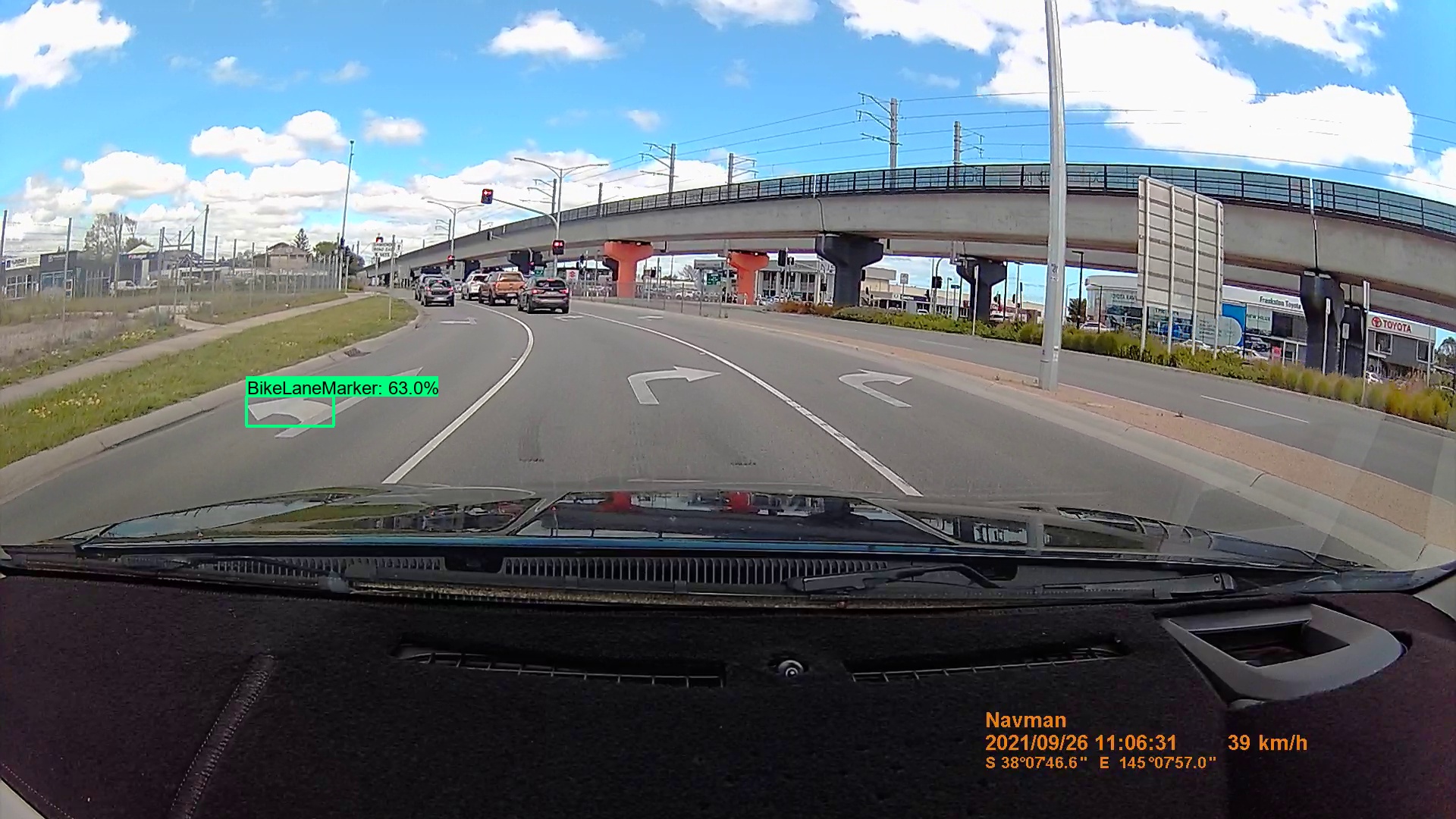}
	\caption{Turning arrow}
\end{subfigure}
\hfill
\begin{subfigure}{0.3\textwidth}
	\includegraphics[width=\textwidth]{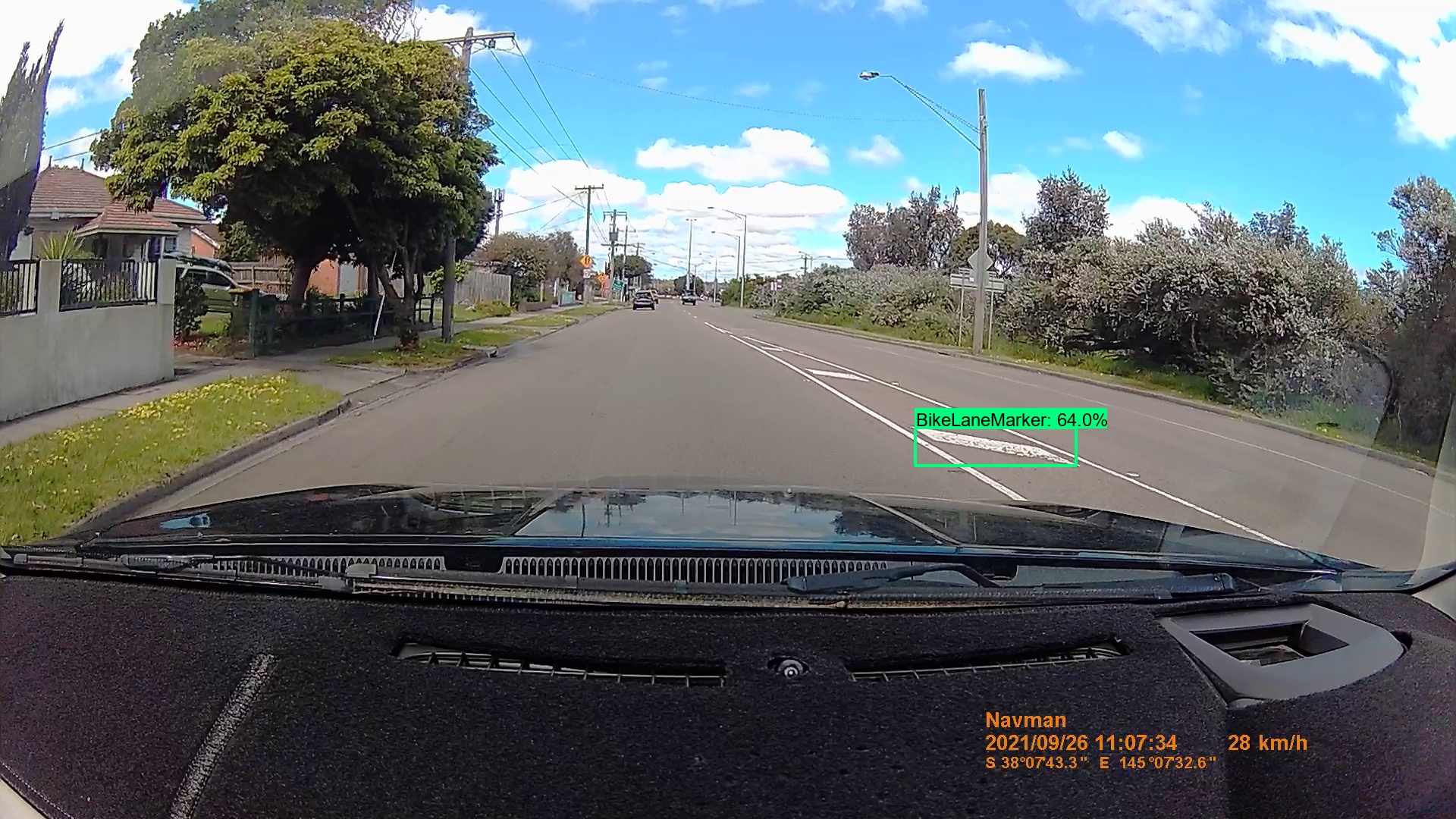}
	\caption{Traffic island}
\end{subfigure}
\hfill
\begin{subfigure}{0.3\textwidth}
	\includegraphics[width=\textwidth]{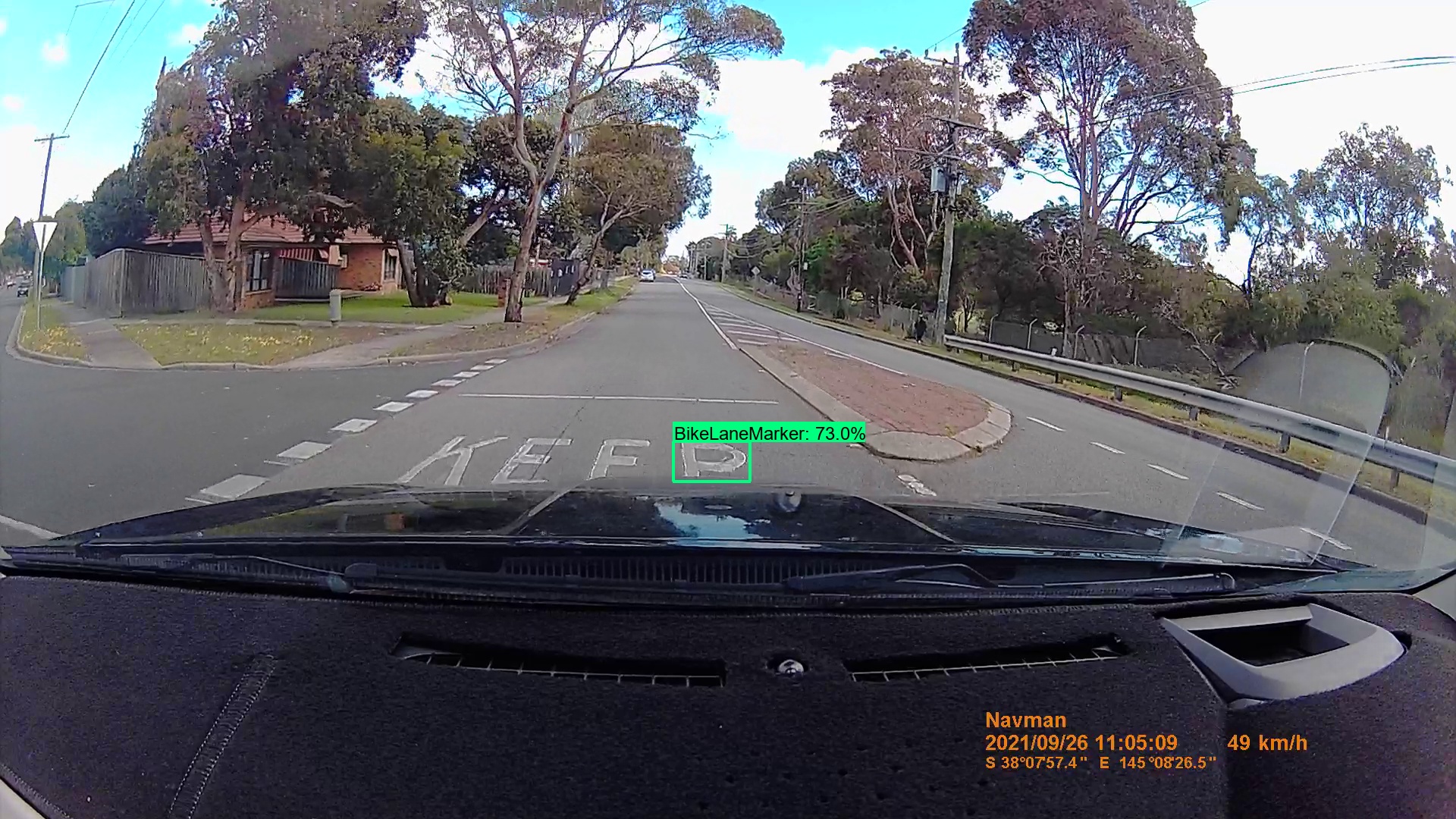}
	\caption{``Keep Clear'' writing}
\end{subfigure}
\hfill
\begin{subfigure}{0.3\textwidth}
	\includegraphics[width=\textwidth]{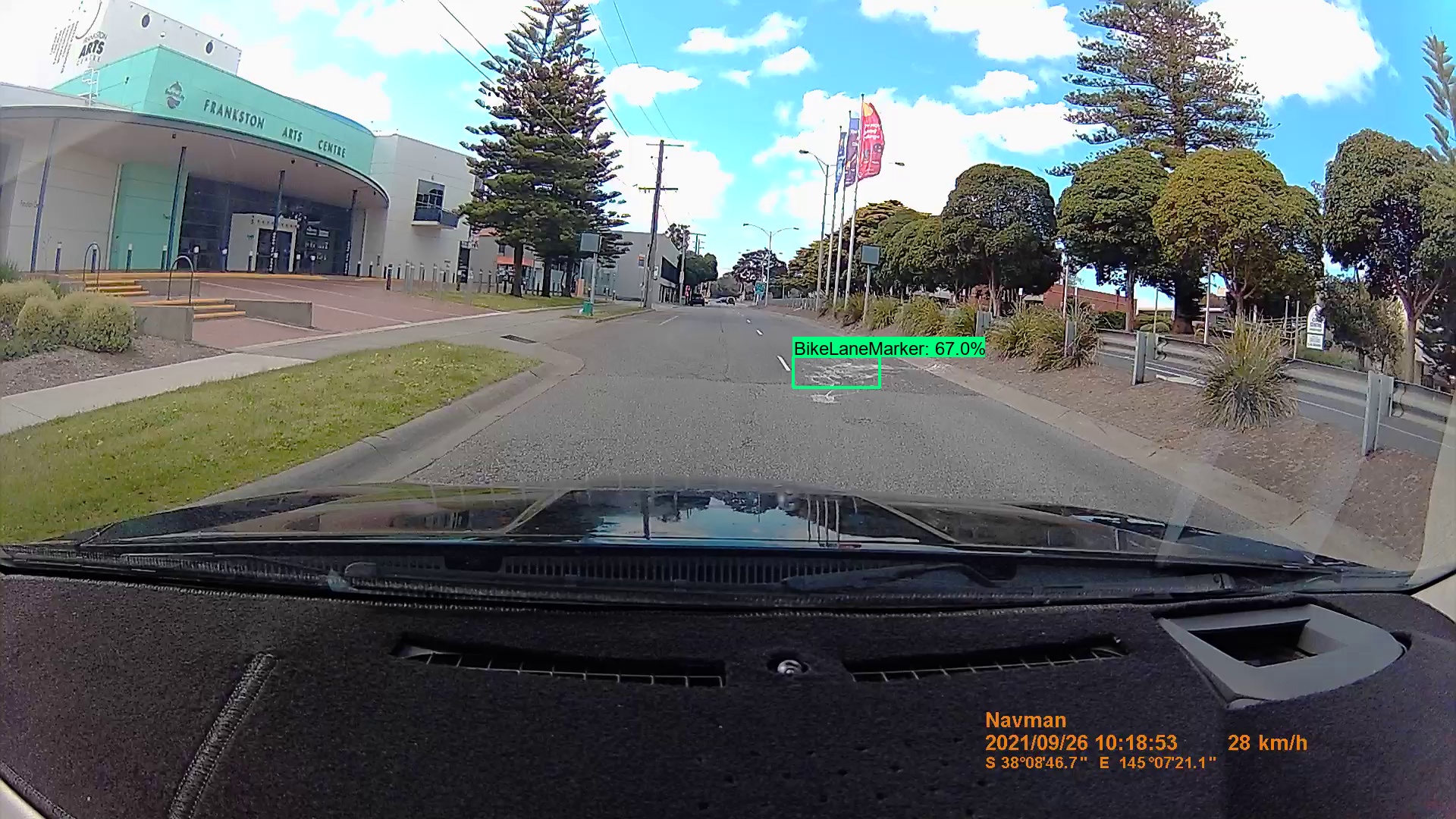}
	\caption{Road surface anomaly}
\end{subfigure}
\hfill
\begin{subfigure}{0.3\textwidth}
	\includegraphics[width=\textwidth]{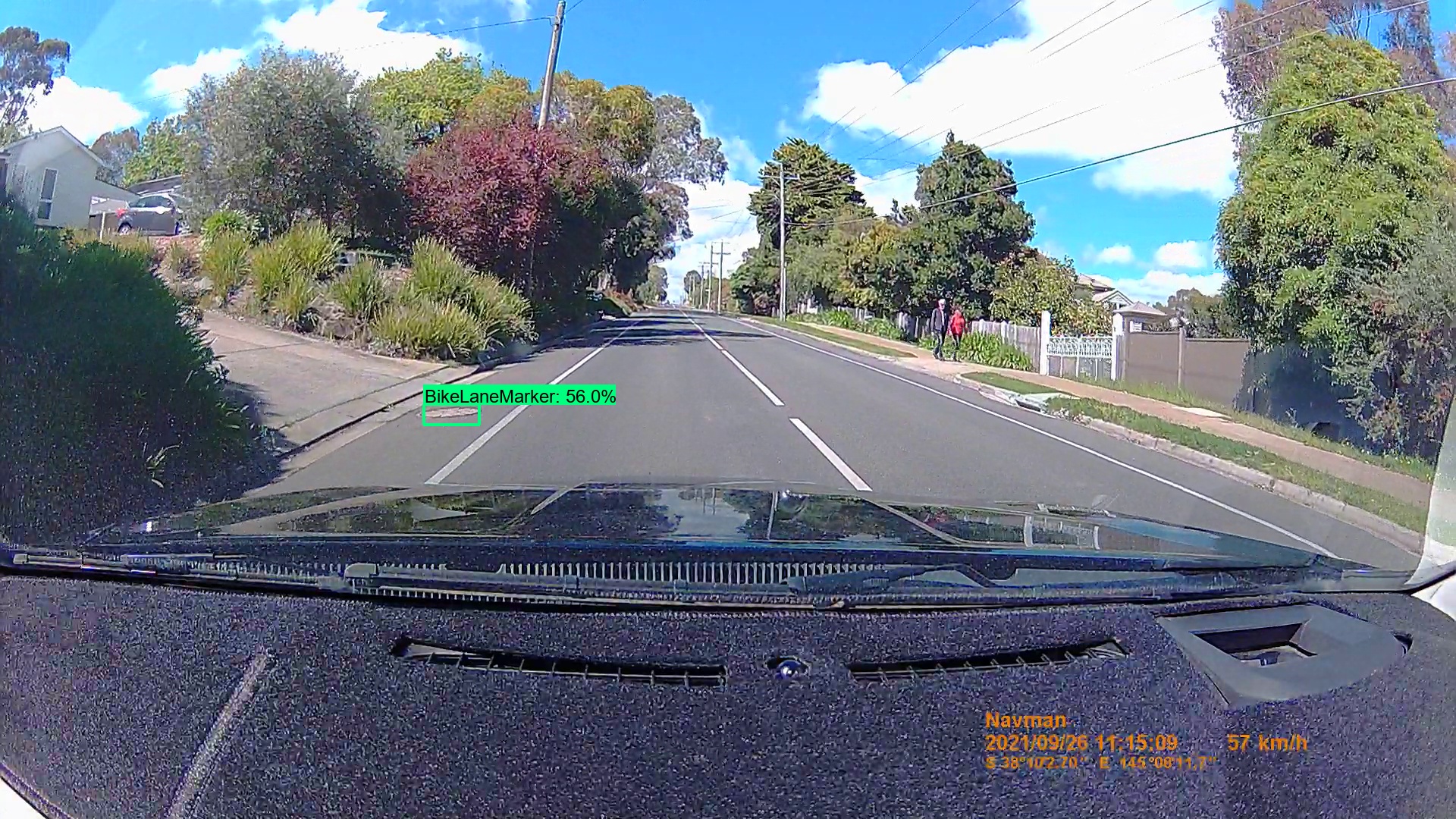}
	\caption{Road surface anomaly}
\end{subfigure}
\hfill
\begin{subfigure}{0.3\textwidth}
	\includegraphics[width=\textwidth]{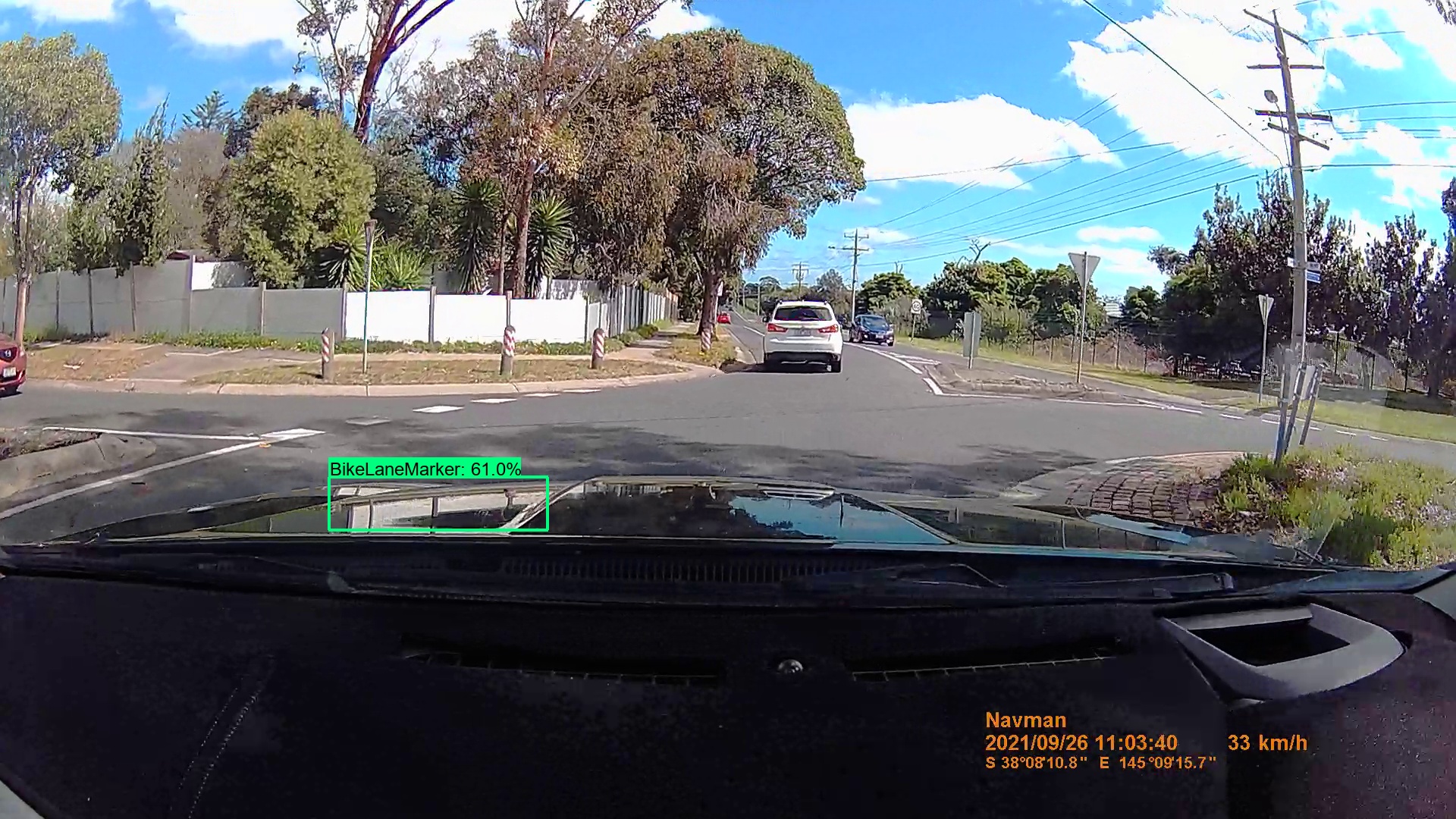}
	\caption{Bonnet reflection}
\end{subfigure}
\begin{subfigure}{0.3\textwidth}
	\includegraphics[width=\textwidth]{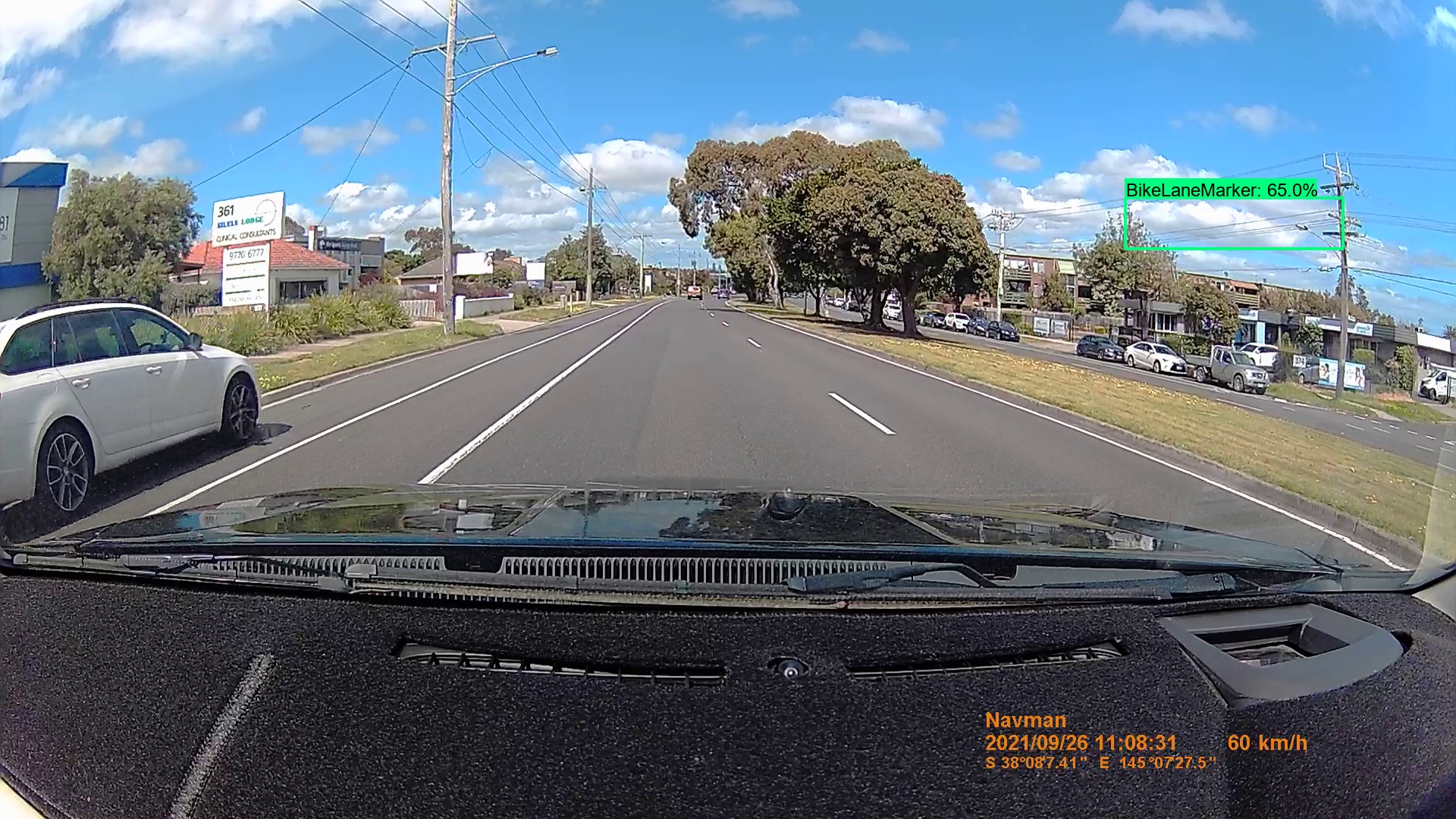}
	\caption{White cloud}
\end{subfigure}
\caption{False Positive Examples}
\label{fig:false_positives}
\end{figure}

See figure \ref{fig:false_positives} for examples.

A decision was made that the detection model needed to be trained with additional data from the dash camera footage.  The footage was divided up into a ``training area'' consisting of footage captured in the Frankston, Langwarrin, and Baxter area, and ``test area'' footage from Mount Eliza.

The Jupyter Notebook documented in appendix \ref{a16} was used to process the dash camera footage from the ``training area'' with the initial detection model that had been trained exclusively on Google Street View images.  Every image that yielded a detection -- whether a true positive or a false positive -- was then labelled and allocated to the ``training dataset'' and ``test dataset'' according to an 80:20 ratio.  A total of 342 images were added to the datasets, supplementing the existing Google Street View images.

The original Google Street View training images were labelled with a single class ``BikeLaneMarker''.  The new training images were labelled with the following additional classes, to encourage the detection model to think of many of the sources of false positives as something other than a bicycle lane marking: ``ArrowMarker'', ``IslandMarker'', ``RoadWriting'', ``GiveWayMarker'', and ``RoadDefect''.

The Google Street View images that had already been included in the dataset were \textit{not} re-labelled to consider the additional classes, though this was considered as an option if required to improve performance.

\begin{figure}[h!]
\centering
\includegraphics[width=0.45\linewidth]{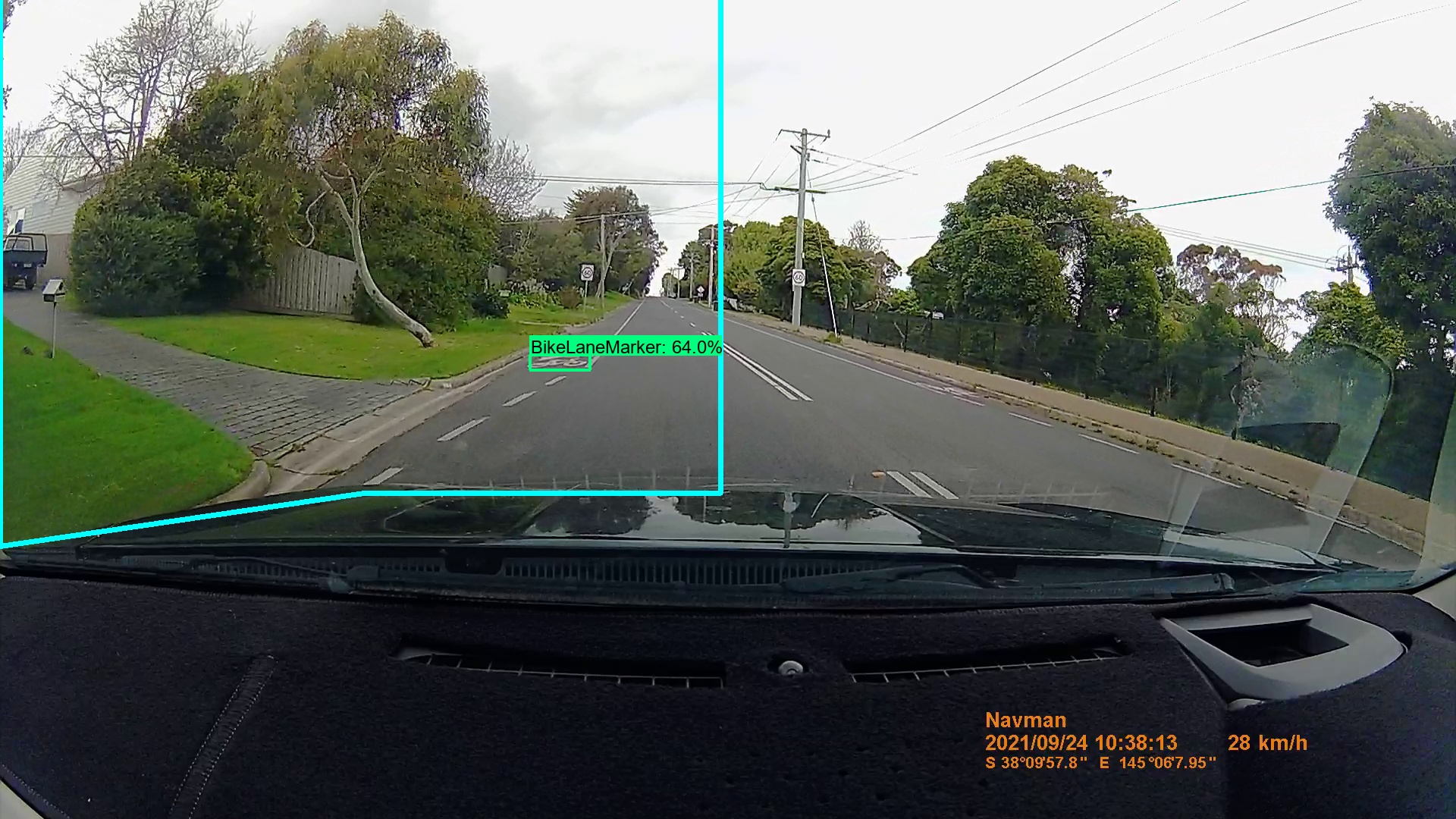}
\caption{Detection mask to exclude car bonnet and right hand side of road.}
\label{fig:006}
\end{figure}

To reduce false positives due to reflections from the bonnet of the camera vehicle, or clouds and other random objects well away from the road, a detection mask was defined so that bicycle lane markings would only count as detections if they fell into an area on the left hand side of the road, in front of the bonnet of the car.  See figure \ref{fig:006} for a visualization of this mask in an example image.

The new images for the enhanced ``training'' and ``test'' datasets were gathered and labelled via the Jupyter Notebook documented in appendix \ref{a13} and the ``labelImg'' tool.  A new model was then trained and evaluated via the Jupyter Notebook documented in appendix \ref{a14}, and its predictions for the ``training'' and ``test'' datasets were analysed in detail using the Jupyter Notebook documented in appendix \ref{a15}.  Finally, the selected model was applied to the dash camera footage from the ``test'' area of Mount Eliza via the Jupyter Notebook documented in appendix \ref{a16}.

\subsection{Mapping dash camera detections and comparing to other sources}
\label{s:rq3d}

Section \ref{s:rq2d} described the process that was used to produce geojson files to compare routes that were detected from Google Street View images to the routes that are recorded in OpenStreetMap.  When sampling images from Google Street View, each sample location can be traced back to a ``way'' and a ``node'' in OpenStreetMap, where the latitude and longitude details came from.  Therefore any detections are already precisely aligned to points in the OpenStreetMap extract.

Images that were collected from dash camera footage take their coordinates from the dash camera's GPS receiver, and they are sampled from locations anywhere along the road.  To infer routes from bicycle lane markings detected in dash camera images, each detection location is mapped to the nearest intersection ``node'' in the OpenStreetMap data.  Once this has been done, the rest of the process is the same as for detections from Google Street View images.

To find the nearest intersection node to a point, the Python ``shapely'' library is first used to find the nearest ``way'' to the point.  The ``shapely'' library appears to use bounding boxes to accelerate this process:  It was significantly faster than a brute-force search of all nodes in the OpenStreetMap data.  Once the closest ``way'' is found, it is traversed to find the nearest ``node'' that is an intersection.  At the same time, the process records the nearest ``node'' of any type, to help with traceability in case the detection occurred a significant distance away from the nearest intersection.

Wherever OpenStreetMap represented a highway or other divided road as multiple parallel ways, the above process was able to distinguish the side of the road that was closest to the camera position.  It is important to gather footage from both sides of a divided road when measuring the differences between the generated map and OpenStreetMap.  Otherwise, OpenStreetMap might count a bicycle lane on two ``ways'' for the divided road, where the generated map only counts one.

\section{RQ4: Surveying other infrastructure details using dash camera footage}
\label{s:rq4}

Further work was conducted to demonstrate how the approach of sampling images, processing them with a model, and correlating the results back to a map, might be re-used to survey other details of interest.  A process was created to map any apparent ``paved shoulders'' on the side of the road in the survey area.  Previous studies such as Klobucar \& Fricker, 2007 \cite{BIKESAFETY} have observed that a paved shoulder can improve cyclist safety, even if it is not formally marked as a bicycle lane.

For a discussion of other areas of potential for future work, please refer to section \ref{s:future_work}.

To identify a paved shoulder, it is necessary to detect lane markings and the edge of the road on the side of the road that vehicles drive on.  For the dash camera footage collected in this research project, this is the left-hand side of the road.

\subsection{Correcting dash camera images for lens distortion}
\label{s:rq4a}

\begin{figure}[h]
\centering
\begin{subfigure}{0.24\textwidth}
	\includegraphics[width=\textwidth]{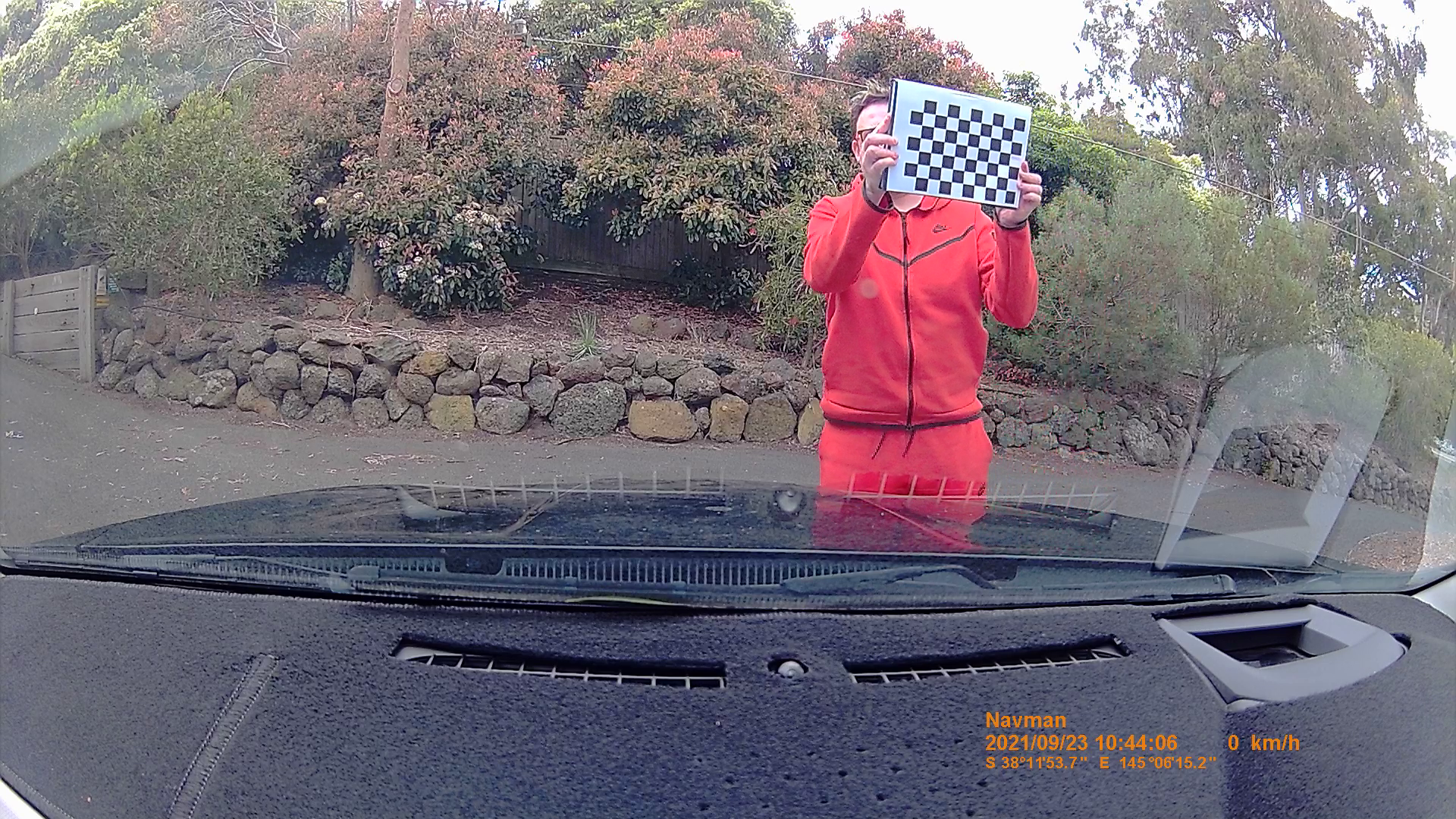}
\end{subfigure}
\hfill
\begin{subfigure}{0.24\textwidth}
	\includegraphics[width=\textwidth]{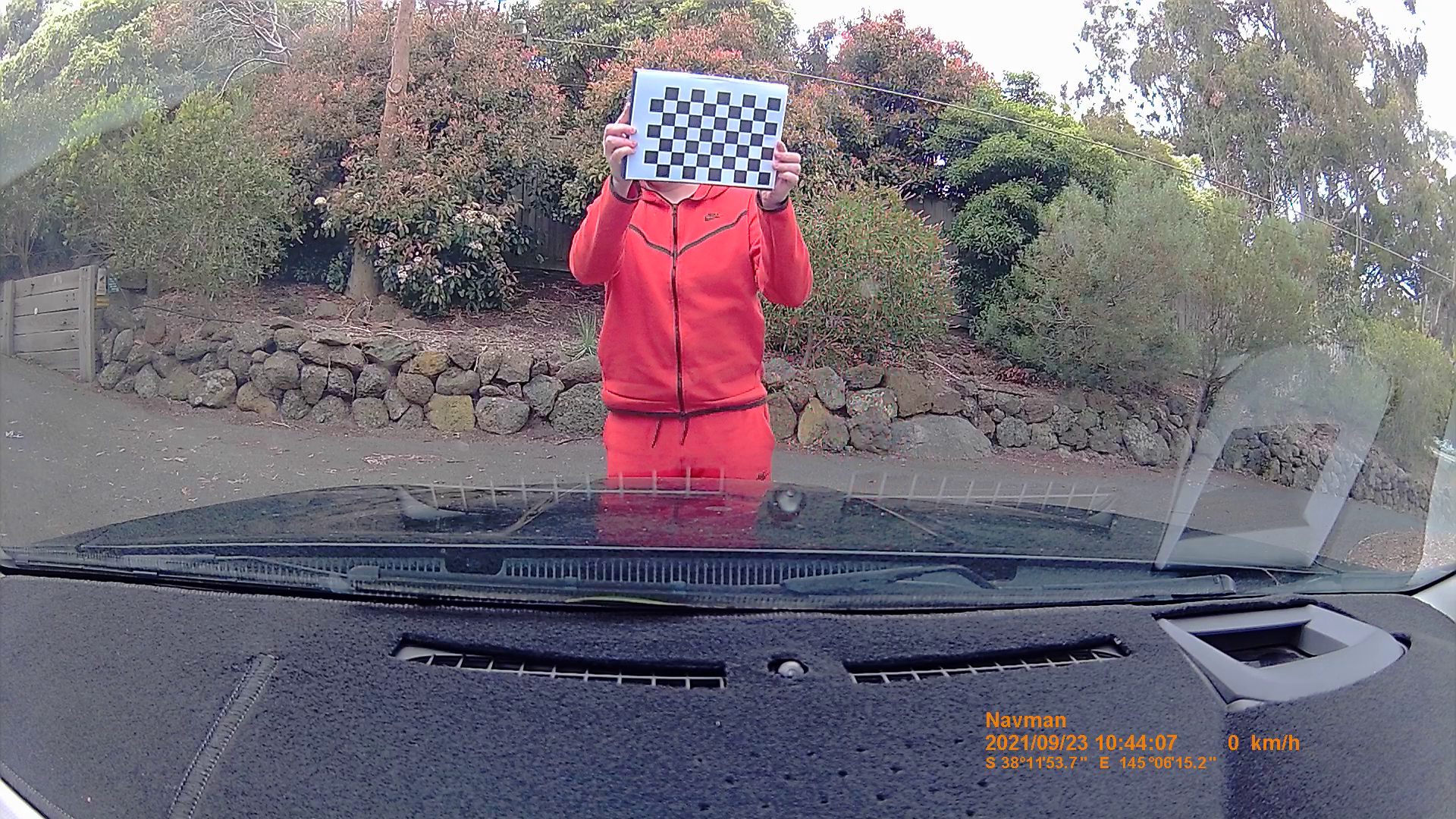}
\end{subfigure}
\hfill
\begin{subfigure}{0.24\textwidth}
	\includegraphics[width=\textwidth]{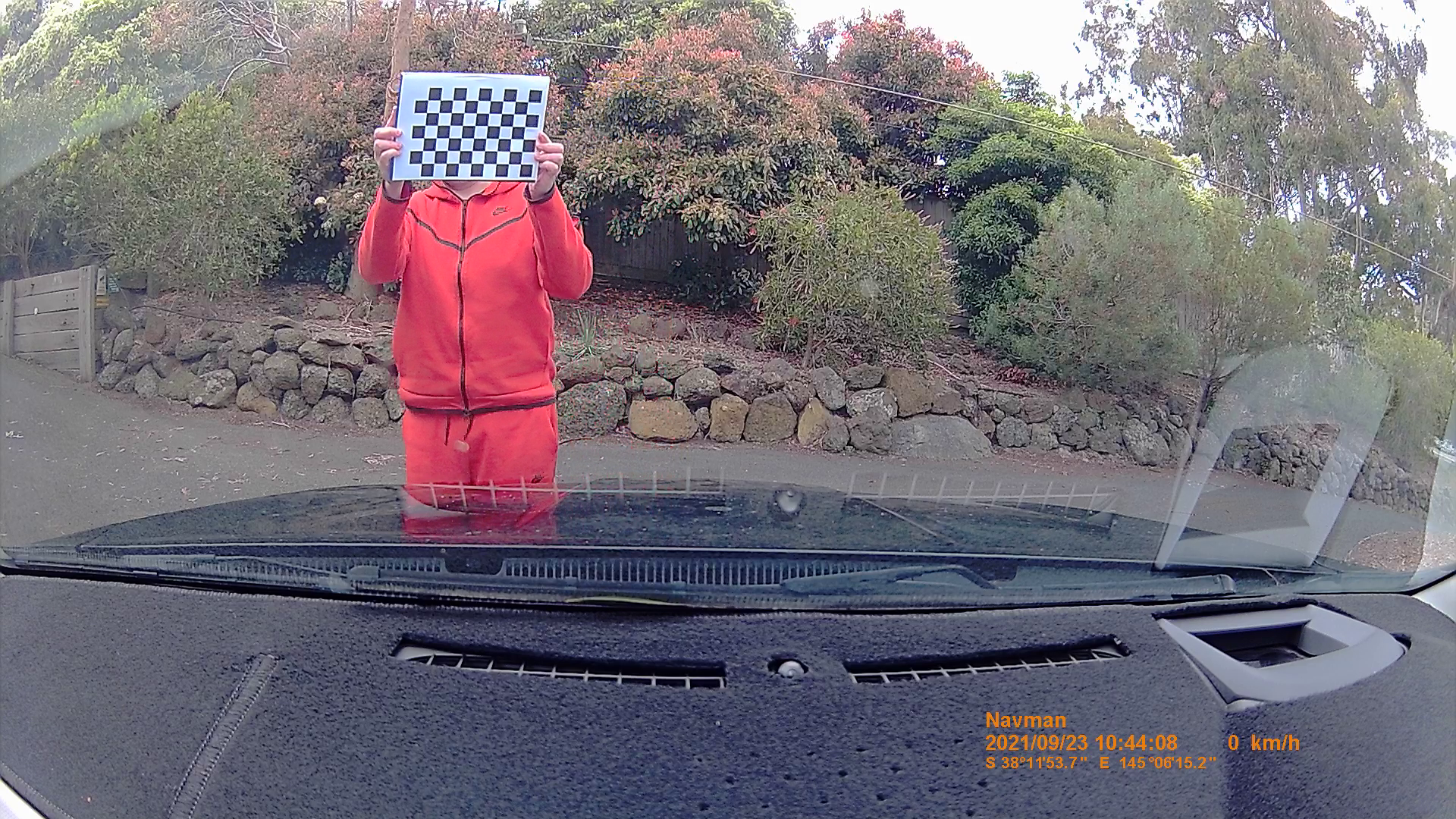}
\end{subfigure}
\hfill
\begin{subfigure}{0.24\textwidth}
	\includegraphics[width=\textwidth]{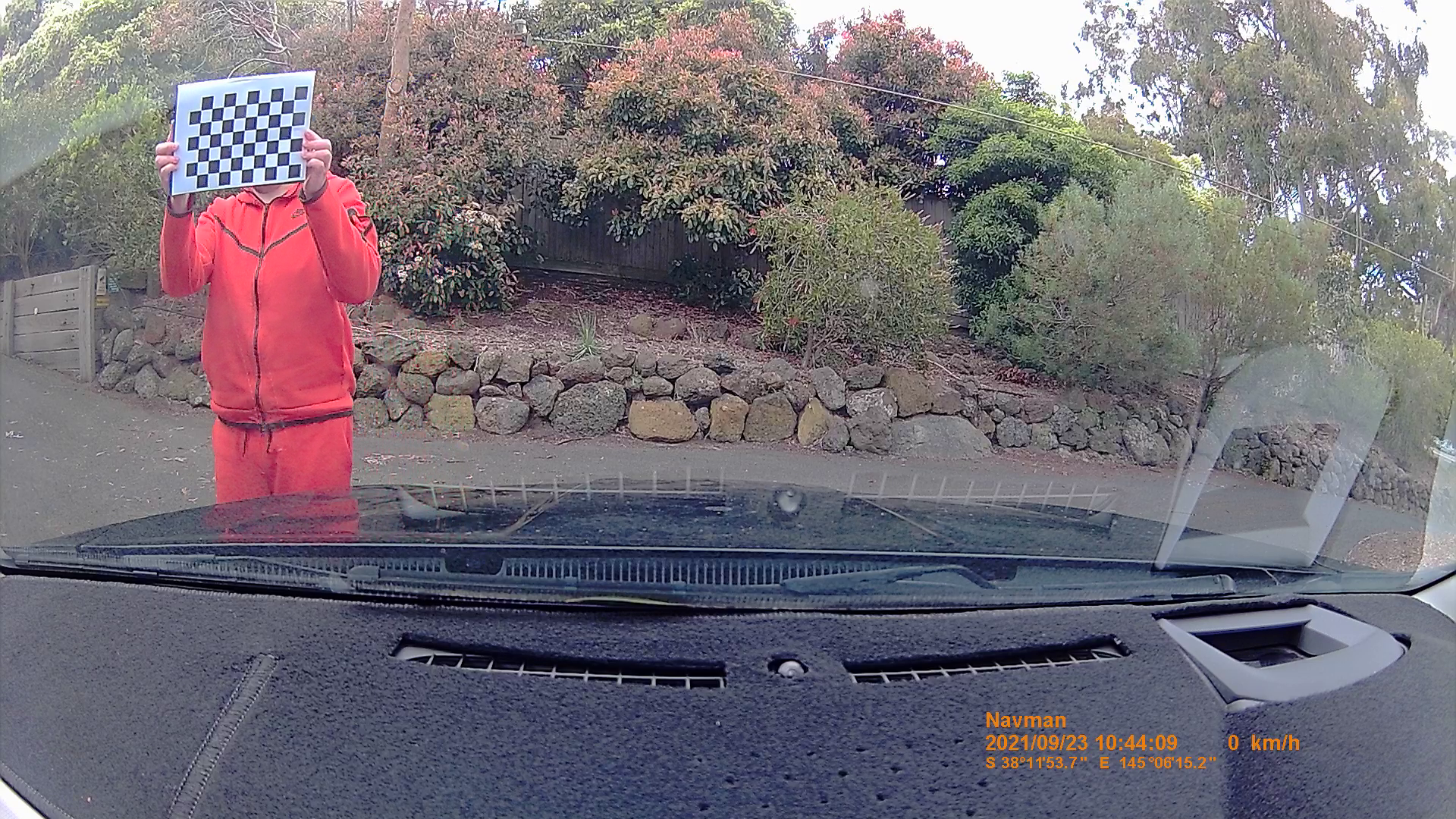}
\end{subfigure}
\caption{Example OpenCV camera calibration images}
\label{fig:calibration}
\end{figure}

Lens distortion is where the optical properties of a camera's lens causes straight lines to appear curved in an image.  The ``OpenCV'' library provides a method to correct distortion for a specific camera \cite{distortion}.  The camera is used to capture a range of images where a special calibration tool appears in frame.  See figure \ref{fig:calibration} for some example images.  An image of a ``chessboard'' pattern of known dimensions is held in front of the camera in different positions.

After the calibration images are collected, they are processed by OpenCV to create a mathematical model of the camera's lens distortion.  With this model, OpenCV can apply a transformation to images from the same camera, to correct for lens distortion.

\begin{figure}[h]
\centering
\begin{subfigure}{0.45\textwidth}
	\includegraphics[width=\textwidth]{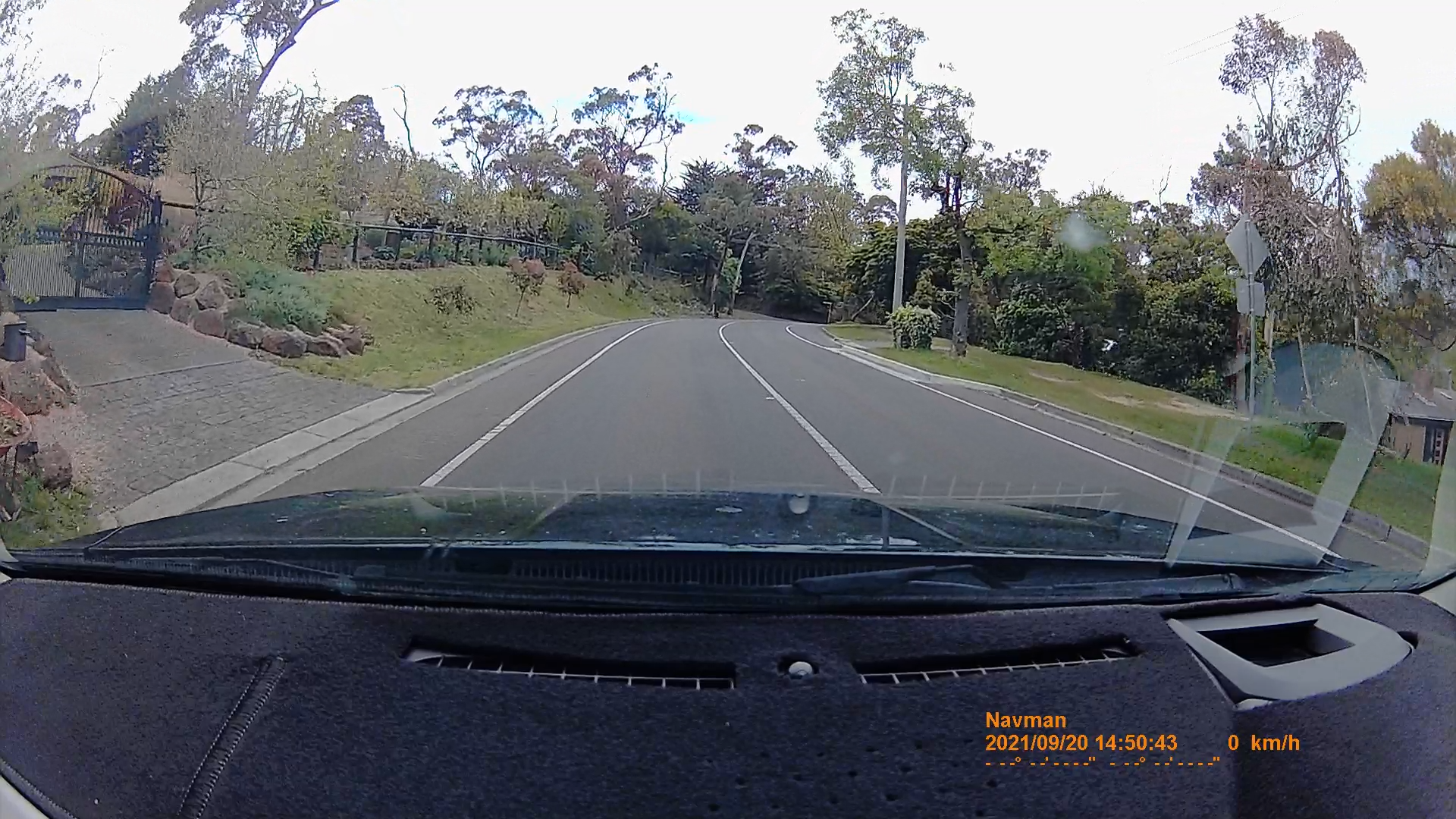}
	\caption{Uncorrected image}
\end{subfigure}
\hfill
\begin{subfigure}{0.45\textwidth}
	\includegraphics[width=\textwidth]{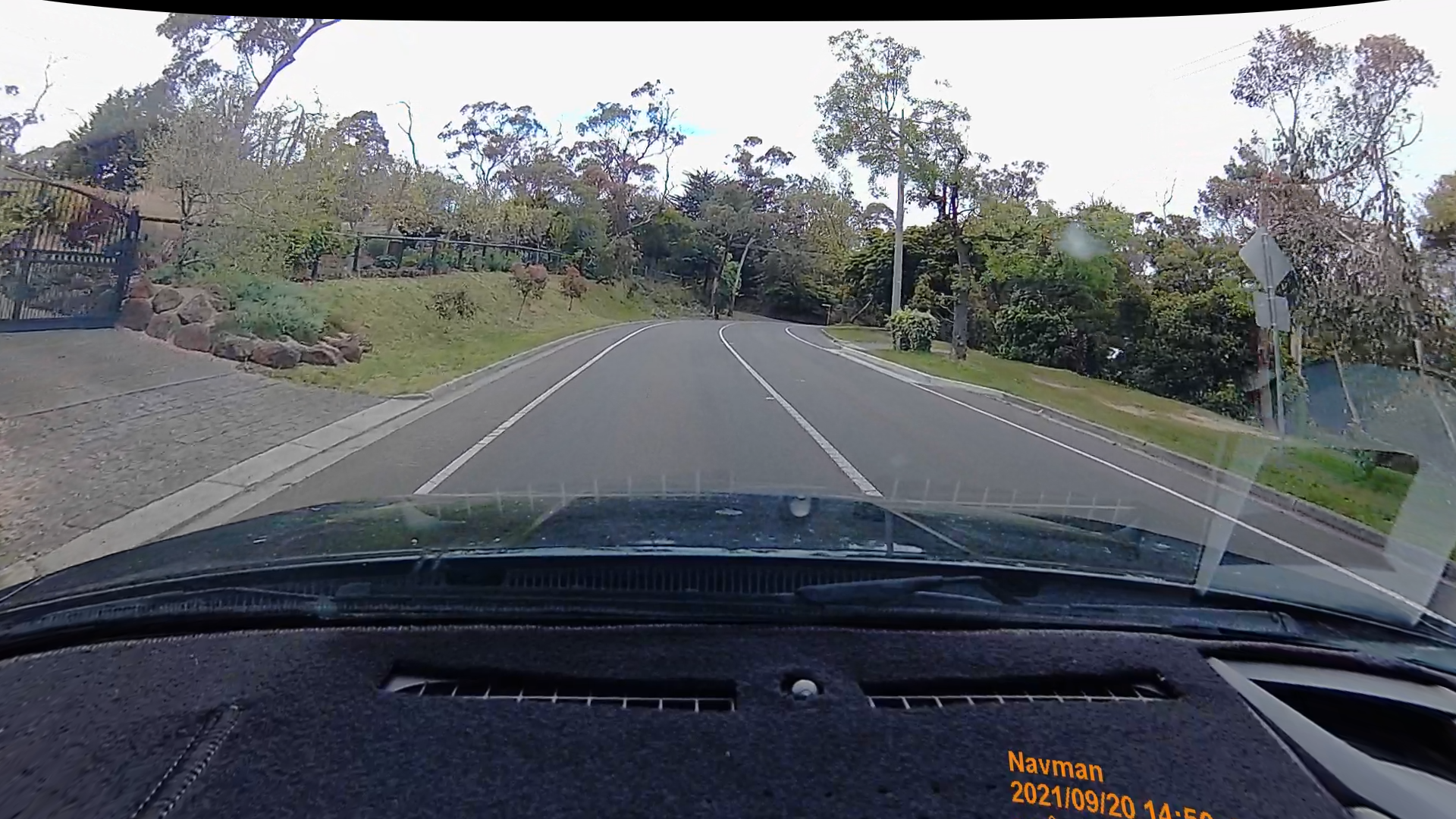}
	\caption{Corrected image}
\end{subfigure}

\caption{Example OpenCV camera calibration images}
\label{fig:008}
\end{figure}

When searching for paved shoulders in a survey area, we are looking for lines in an image that are often straight.  The dash camera being used to collect images was therefore calibrated, and in this exercise, all images were processed to correct for distortion.  See figure \ref{fig:008} for an example raw image from the dash camera, and a corresponding corrected version.

There was not much difference between the corrected and uncorrected images.  The most obvious sign that the correction has been applied in figure \ref{fig:008} is that the top edge of the corrected image is bent downwards slightly, and the location/speed/time information that usually appears in the bottom right corner has been shifted.  Although the operation to correct for lens distortion did not appear to have a significant impact when it came to detecting paved shoulders, it was retained in the pipeline to support any future photogrammetry work to estimate the widths of the lanes.

\subsection{Detecting lane markings}
\label{s:rq4b}

A common solution to the problem of detecting lane markings involves applying a combination of the Canny edge detection algorithm proposed by Canny, 1987 \cite{canny} and the Hough transformation proposed by Hough, 1972 \cite{hough}.  This general approach has been followed in many papers such as Li et al, 2016 \cite{canny_example} and Chai et al, 2014 \cite{canny_example2}, and it is frequently used in capstone projects in the self-driving car domain.  The approach does not involve any machine learning or deep learning techniques.

The Canny-Hough approach typically works as follows:

\begin{enumerate}[label=\alph*.]
\item{Correct the image for lens distortion.}
\item{Apply Canny edge detection to detect edges in an image \cite{canny}.}
\item{Apply a mask to exclude edges outside a triangle representing the area immediately in front of the vehicle.}
\item{Apply a Hough transform \cite{hough} to convert these edges into lines, each line having a slope and an intercept.}
\item{Partition the lines into two groups:  Lines that slope upwards when scanning from left to right, and lines that slope in the opposite direction.  For each group of lines, take the average slope and intercept.  These represent the assumed lane boundaries.  Optionally redraw the image with the detected lines drawn as an overlay, to visualize them.}
\end{enumerate}

\begin{figure}[h]
\centering
\begin{subfigure}{0.3\textwidth}
	\includegraphics[width=\textwidth]{f008_corrected.png}
	\caption{Corrected image}
\end{subfigure}
\hfill
\begin{subfigure}{0.3\textwidth}
	\includegraphics[width=\textwidth]{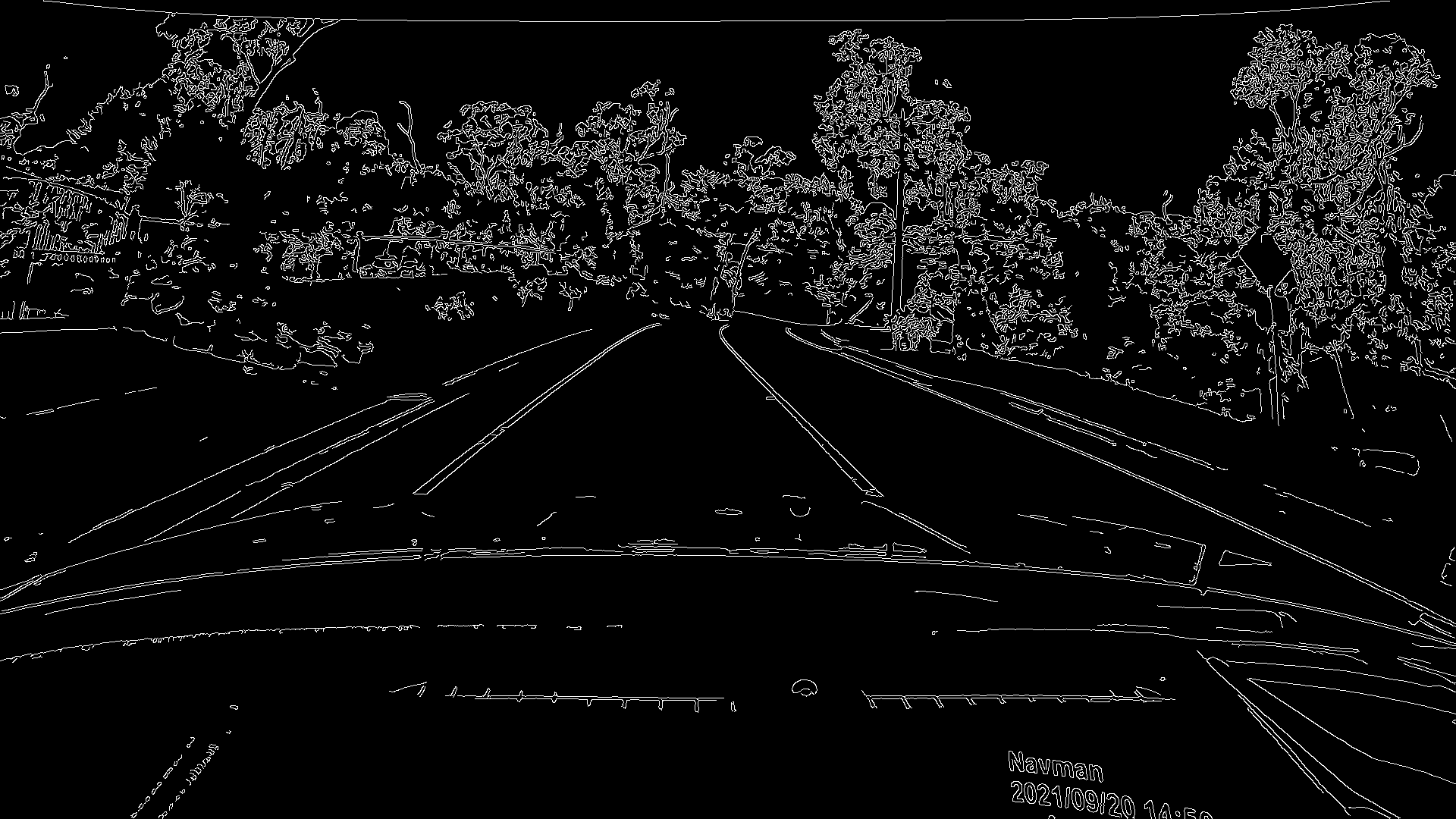}
	\caption{Canny edge detection}
\end{subfigure}
\hfill
\begin{subfigure}{0.3\textwidth}
	\includegraphics[width=\textwidth]{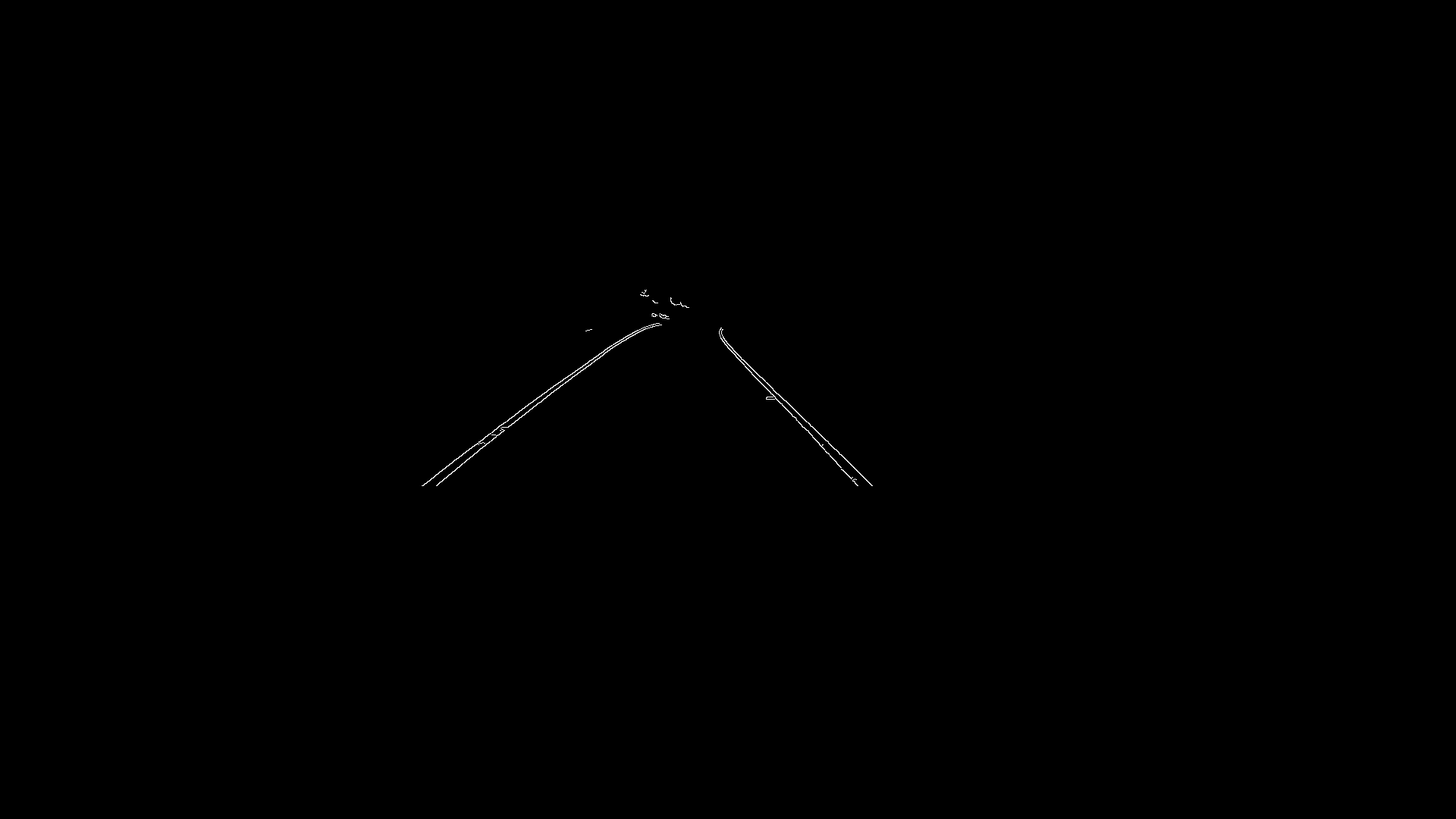}
	\caption{Apply mask}
\end{subfigure}
\hfill
\begin{subfigure}{0.3\textwidth}
	\includegraphics[width=\textwidth]{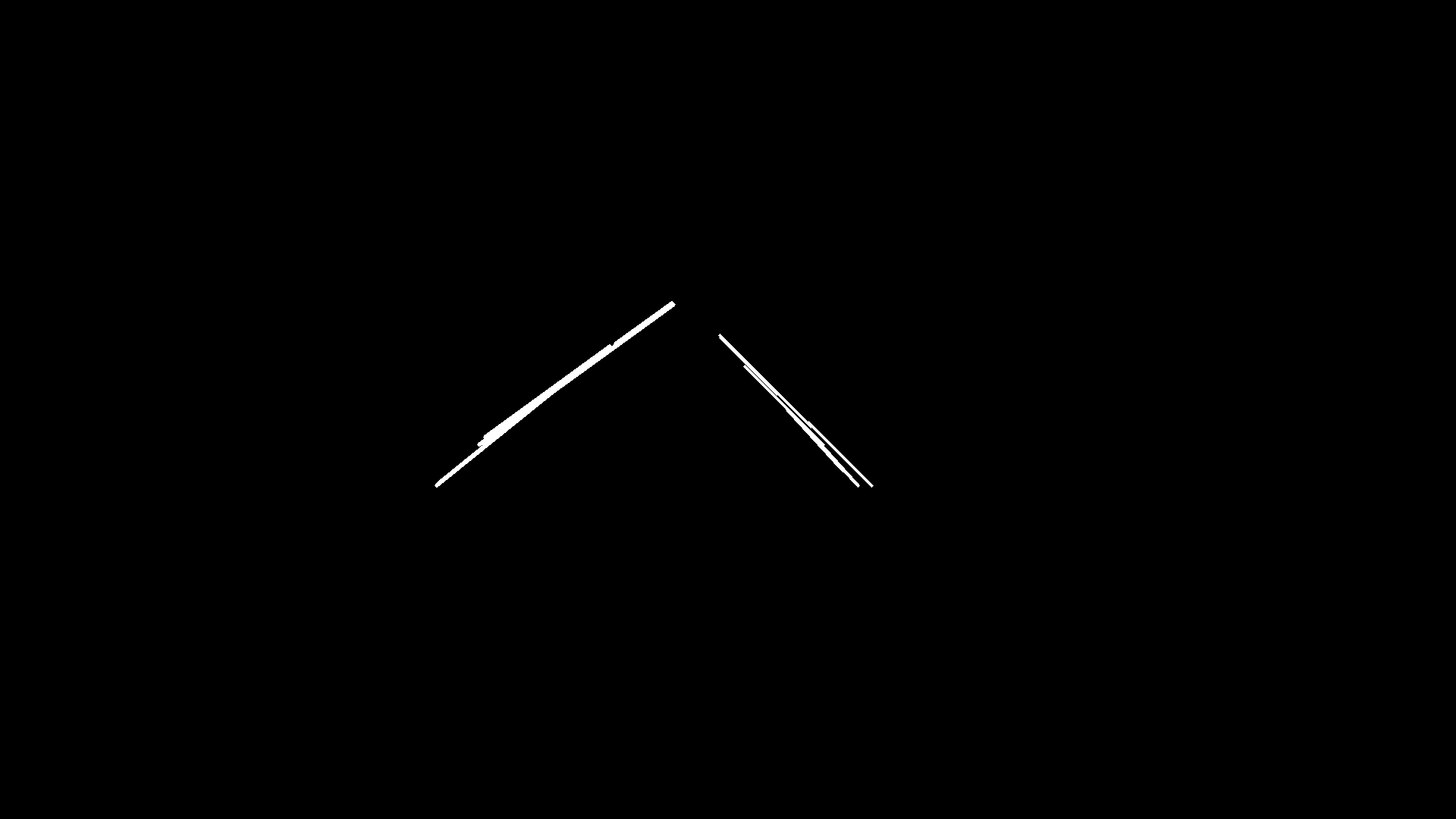}
	\caption{Hough transform}
\end{subfigure}
\hfill
\begin{subfigure}{0.3\textwidth}
	\includegraphics[width=\textwidth]{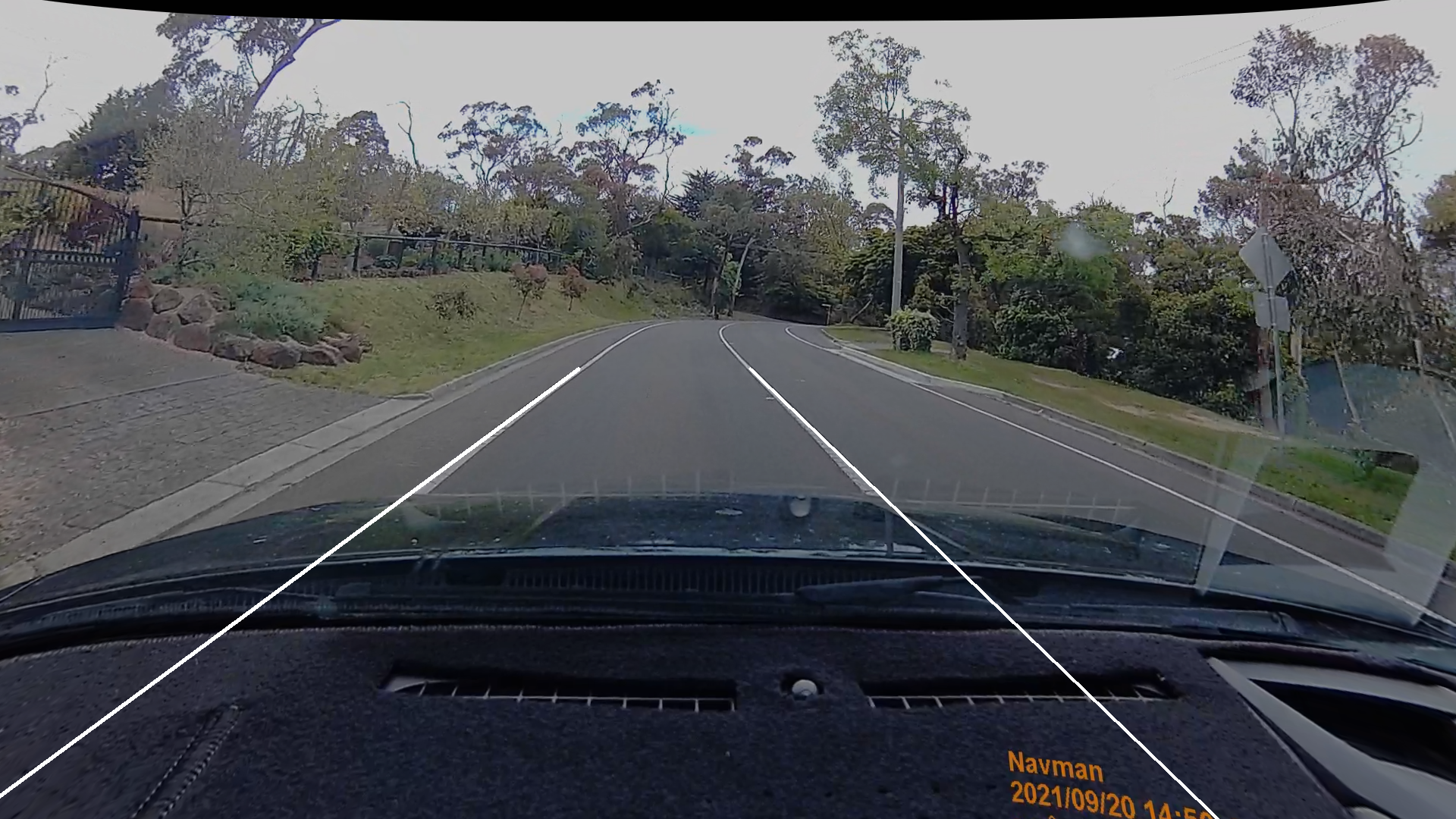}
	\caption{Average lines, superimposed on original image}
\end{subfigure}
\caption{Example sequence of Canny-Hough operations to detect own lane}
\label{fig:009}
\end{figure}

Please see figure \ref{fig:009} for example images to demonstrate the process.
\\

The ``OpenCV'' library provides convenient tools to implement the required transformations.

Typically, the Canny-Hough process is used to detect the camera vehicle's own lane.  In order to detect a paved shoulder, we first want to detect the boundaries of our vehicle's own lane, then perform a second pass at the image where the mask has shifted to focus exclusively on areas further to the edge of the road.

\begin{figure}[h]
\centering
\begin{subfigure}{0.3\textwidth}
	\includegraphics[width=\textwidth]{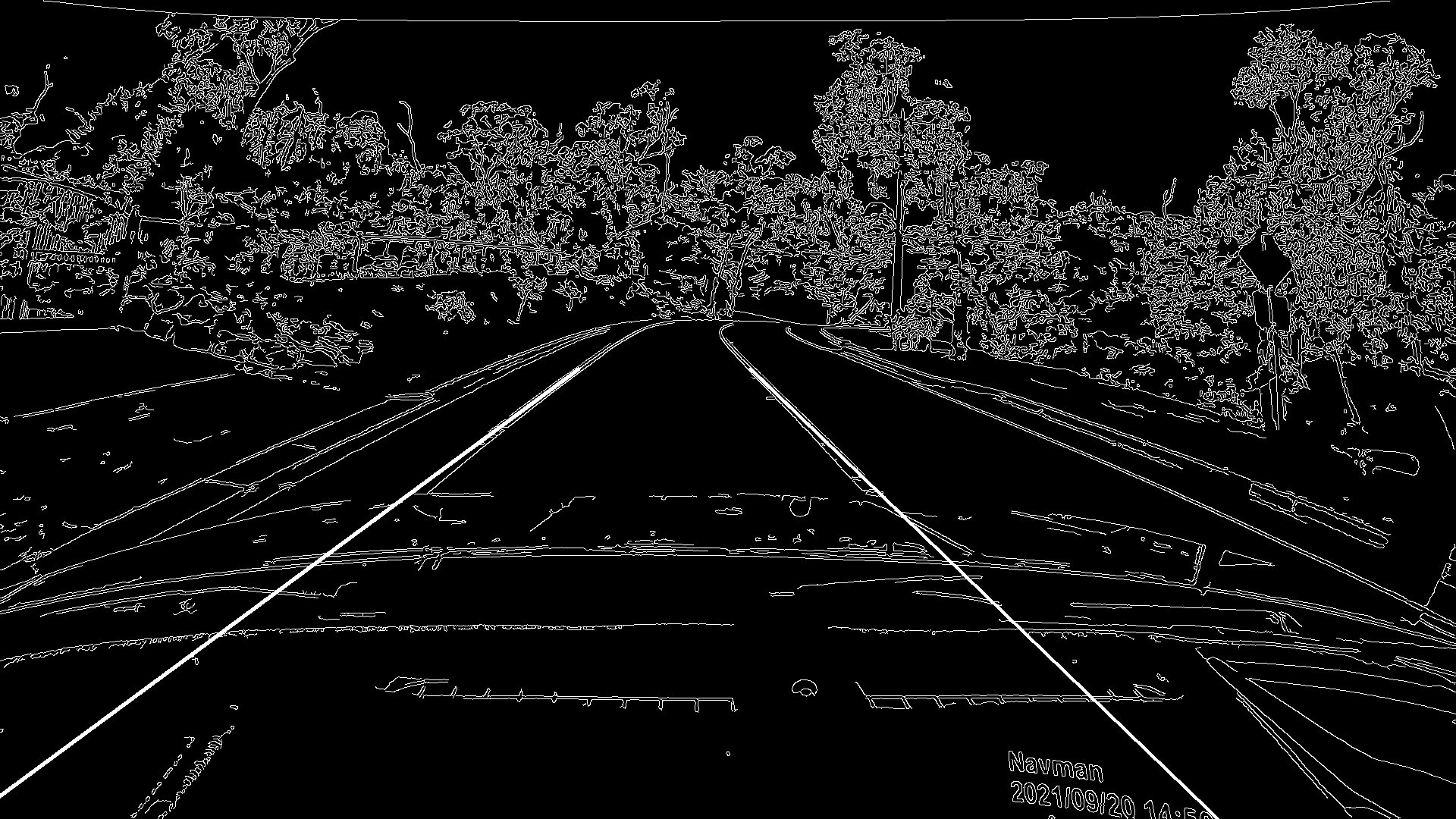}
	\caption{Canny edge detection with previously detected lane}
\end{subfigure}
\hfill
\begin{subfigure}{0.3\textwidth}
	\includegraphics[width=\textwidth]{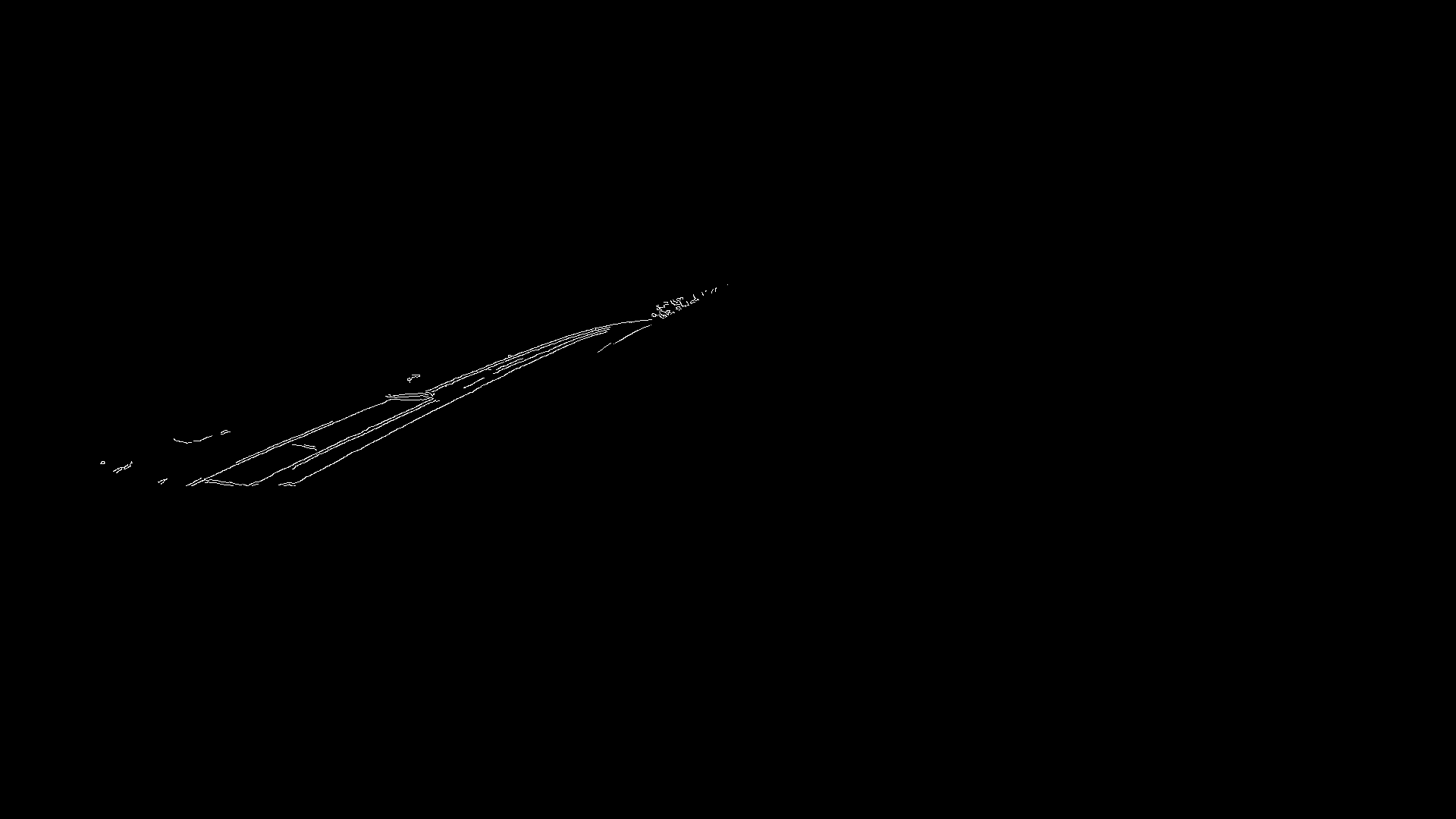}
	\caption{Canny edges with new mask applied}
\end{subfigure}
\hfill
\begin{subfigure}{0.3\textwidth}
	\includegraphics[width=\textwidth]{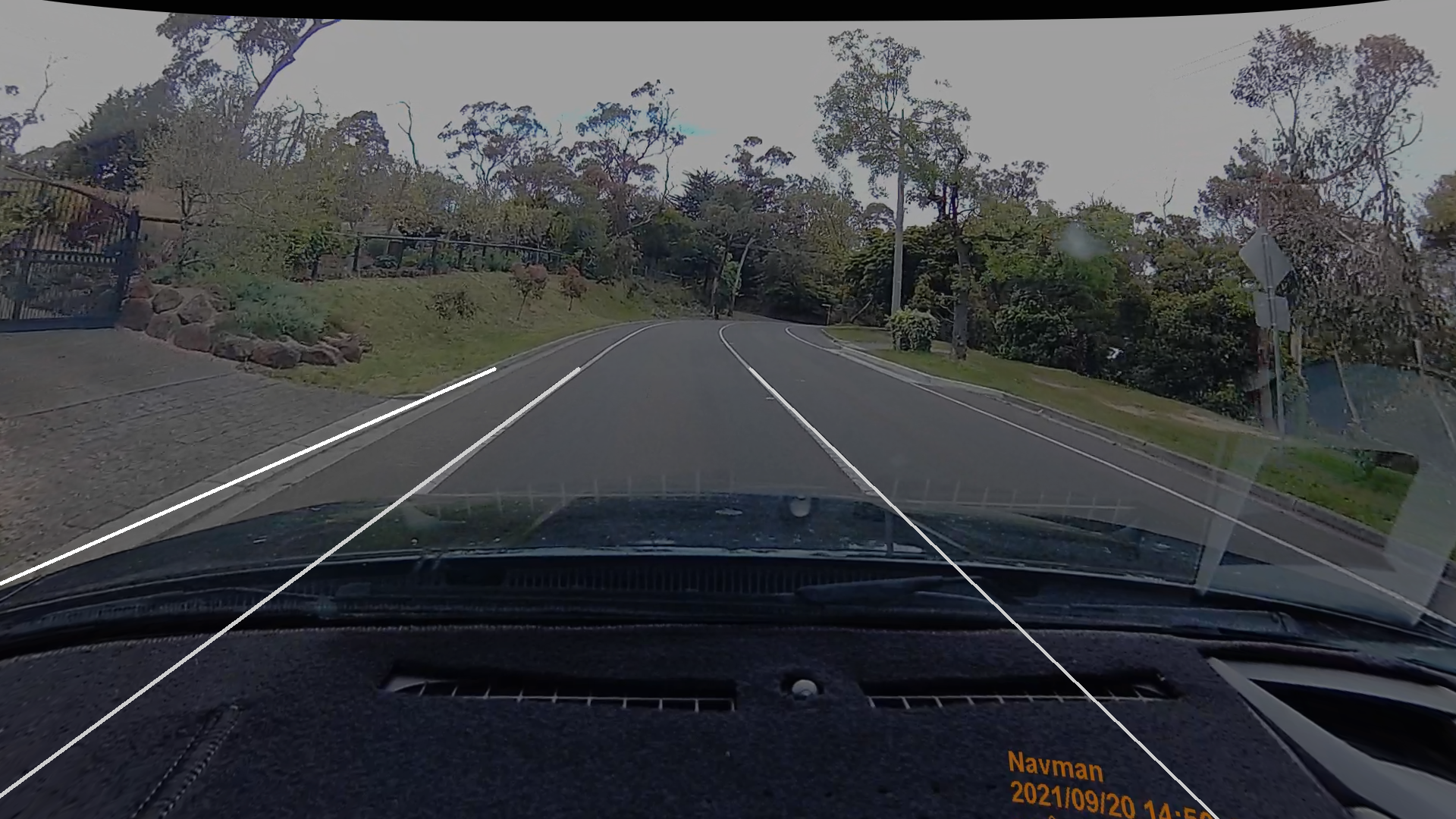}
	\caption{Image with all three detected lines}
\end{subfigure}
\caption{Example sequence of Canny-Hough operations to detect paved shoulder}
\label{fig:010}
\end{figure}

A paved shoulder will be defined by two lines, each with a ``slope'' and an ``intercept''.  One line will shared with a boundary of the camera's own lane.  The other line will be the edge of the road.  To detect this extra line, we apply a mask to the Canny edge detection image just to the left of the camera's own lane boundary that was found in the first step.  We then apply the same Hough transformation to detect lines, and average the ones with the expected ``left'' slope, to find a slope and intercept for the outer boundary of the paved shoulder.  See figure \ref{fig:010} for an example.

If the Canny-Hough approach cannot find another line to the left of the camera vehicle's own lane, it is a strong sign that there is no clearly defined paved shoulder on the road at that location.  If a line is found, it could be a paved shoulder, or it could be a false positive caused by ``noise'' from something on the side of the road, but within the area included by the mask.

\begin{figure}[h]
\centering
\begin{subfigure}{0.3\textwidth}
	\includegraphics[width=\textwidth]{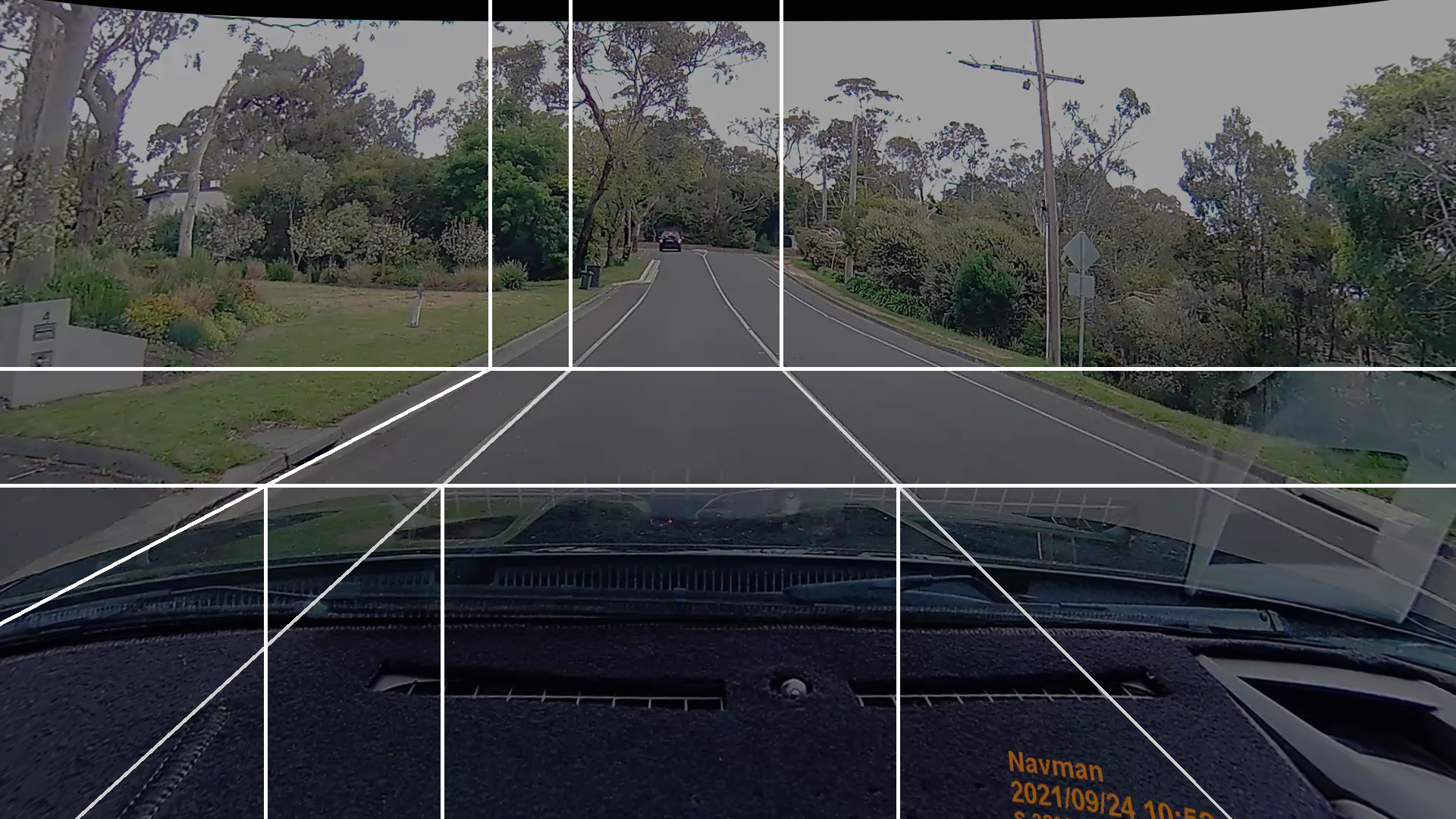}
	\caption{A paved shoulder has been correctly identified}
	\label{fig:011a}
\end{subfigure}
\hfill
\begin{subfigure}{0.3\textwidth}
	\includegraphics[width=\textwidth]{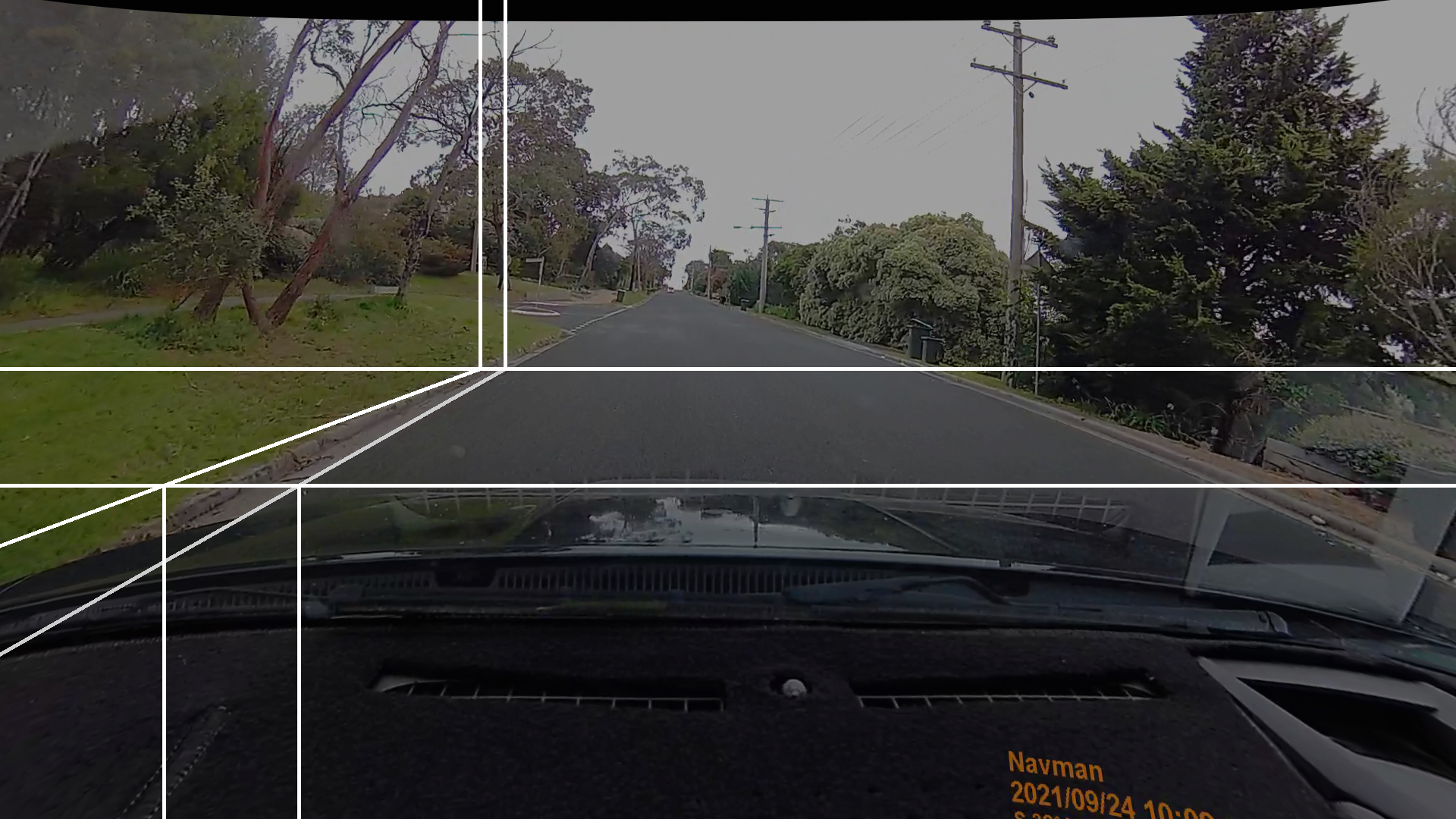}
	\caption{A gutter has resulted in a very narrow shoulder area}
	\label{fig:011b}
\end{subfigure}
\hfill
\begin{subfigure}{0.3\textwidth}
	\includegraphics[width=\textwidth]{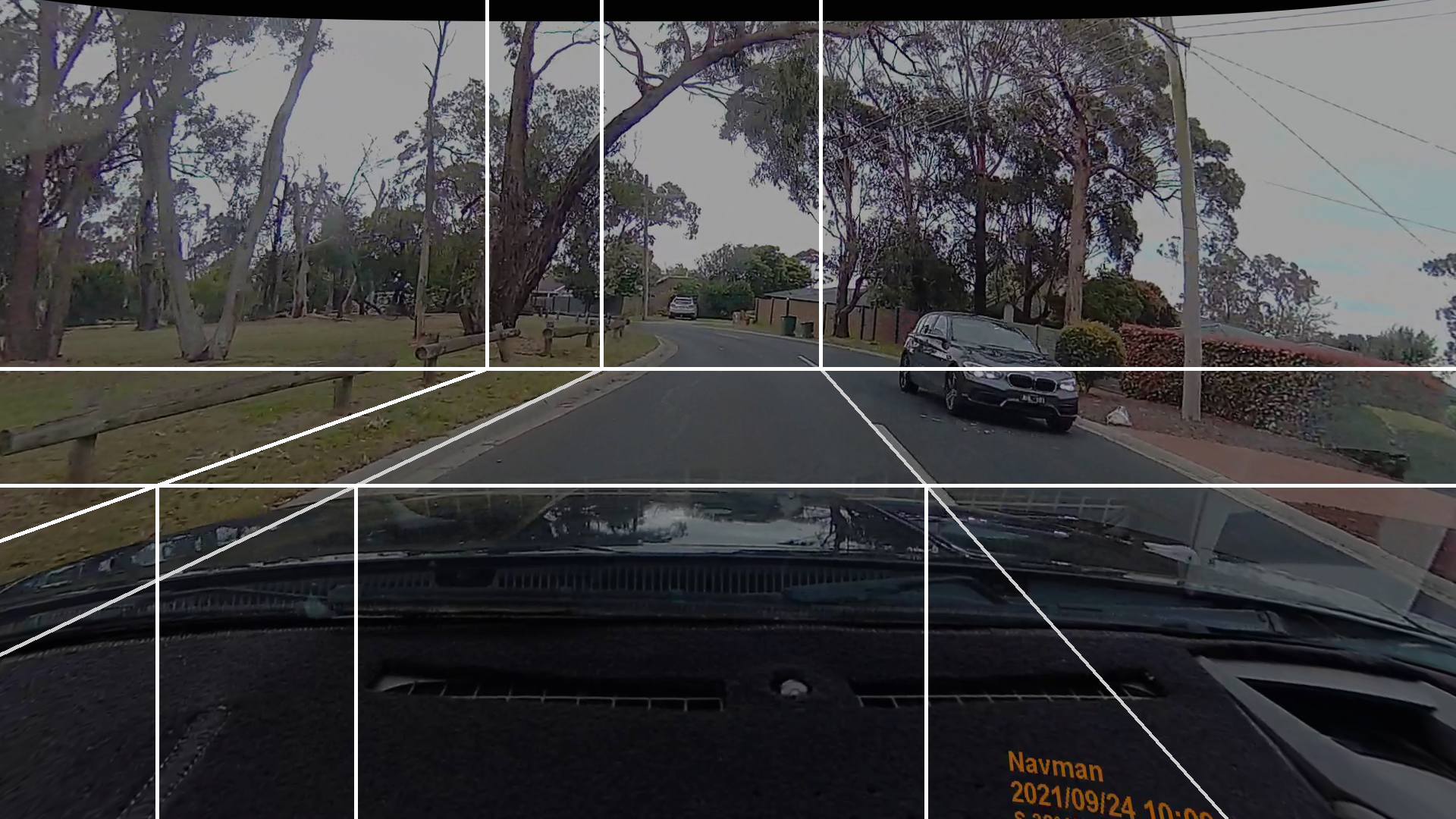}
	\caption{``Noise'' from the straight lines in a fence}
	\label{fig:011c}
\end{subfigure}
\caption{Example paved shoulder detections}
\label{fig:011}
\end{figure}

Please see figure \ref{fig:011} below for example images where lines were detected for a possible paved shoulder.  In each of the example figures, two horizontal lines have been drawn.  The lower horizontal line shows where the detected lines intersect a horizontal line just above the bonnet of the car, the nearest point in the frame with a clear view.  The upper horizontal line shows where they intersect at an arbitrary point further into the distance.  Vertical lines are also drawn to highlight the intersection points.  Together, the horizontal and vertical lines are a visual aid to help assess the width and stability of any detected lanes, as the operator views the images quickly in sequence.

Figure \ref{fig:011a} shows a true positive match, where the boundaries of a paved shoulder have been correctly identified.  Figure \ref{fig:011b} shows a situation where a concrete gutter has resulted in a very narrow area being detected.  This potential for this type of false positive can be mitigated by requiring a minimum width for any detected lane.  Figure \ref{fig:011c} shows a false positive match, where ``noise'' from a low wooden fence parallel to the road has caused the average of lines to the left of the camera's old lane to appear to be a paved shoulder, off-road and in the grass.  This phantom line is not ``stable'' and only appears for a few frames.  This type of false positive can be reduced using metrics that require a paved shoulder to be consistently detected across most images in a road segment, and/or requiring that the slope and position of the line not vary too much from frame to frame.

\subsection{Mapping paved shoulders across a survey area}
\label{s:rq4c}

The Canny-Hough lane detection method can be applied to all sample frames from the survey area, using a similar batch process to the bicycle lane detection process in section \ref{s:rq3} and substituting a different detection model.  The challenge is setting robust criteria for flagging a ``detection'' that will result in a sensible map.  The simplest criteria would be to look for frames where lines were detected for both boundaries of a paved shoulder, but when it was applied to the training area footage, the results were very inconsistent.  On roads where there was a paved shoulder or bicycle lane, the boundaries of that lane remained relatively stable from frame to frame, and the width between the boundaries of the detected area was relatively wide.  There were few ``skipped'' frames where the boundaries were not detected.  In contrast, on roads without a paved shoulder or bicycle lane, there may be many ``skipped'' frames, the slopes of the detected boundaries were inconsistent from frame-to-frame, and the detected boundary lines often had an intersection point much closer to the bottom of the frame.

The following solution was proposed:

\begin{itemize}
\item{Each frame would be linked to its two nearest intersection ``nodes'' in the OpenStreetMap data.  All frames associated with a single stretch of road between two intersections would be assessed as a group.}
\item{If the model fails to find both boundaries of a paved shoulder in more than a threshold percentage of frames along a stretch of road, then it is assumed that there isn't one.}
\item{Frames from locations within 30 metres of an intersection were excluded from consideration, to account for routine tapering of paved shoulders at intersections.}
\item{A horizontal line is drawn across the frame towards a ``horizon'' as per figure \ref{fig:011}.  The width of the space between the boundary lines is measured at this height within the frame, in pixels.  The mean width is calculated across all frames in the stretch of road where boundary lines were detected.  If the mean is less than a threshold number of pixels, then the paved shoulder is too narrow, and it is assumed not to exist.  This accounts for very narrow margins such a gutter that are of no use to a cyclist.}
\item{If the model finds both left and right boundary lines for a possible paved shoulder, use the slopes and intercepts to calculate an (x, y) coordinate relative to the bitmap image where those two boundary lines intercept.  Calculate the standard deviation of the x and y intersection values across all frames in the stretch of road where both boundary lines are defined.  If the standard deviation in either the x or y dimension is greater than a threshold number of pixels, then the boundary lines are considered to be too inconsistent across images.  The stretch of road is assumed not to have a paved shoulder.}
\end{itemize}

The thresholds described above were tuned by analysing the dash camera footage from the ``training'' area and producing a CSV file with one record per frame, in chronological order.  Each frame was mapped to the nearest ``way'' and the two nearest intersection ``nodes'' in the OpenStreetMap data, to associate it with a particular stretch of road, from intersection-to-intersection.  It was also labelled with the name of the road, for traceability.  The required summary statistics were calculated for each group of frames associated with a stretch of road, and these were attached to the CSV file as additional columns.  Finally the footage was reviewed, and using the road names, changes of heading and ``node'' IDs as a guide, each record was labelled to record whether there was really a paved shoulder there or not.

A manual tuning process was performed, filtering the rows based on the column recording the ``ground truth'', to find thresholds that appeared to correctly predict the outcome as often as possible.

The model was then applied to the footage from the Mount Eliza ``testing'' area to see how it performed with the selected thresholds.

To construct a map of detected paved shoulders, the dataset was summarised to a list of distinct combinations of ``way'' ID and intersection ``node'' IDs for each road section where a paved shoulder was predicted.  This information was cross-referenced to the OpenStreetMap data for each ``way'' ID, and a line was ``drawn'' in a geojson file for all ``nodes'' on the way between the two intersection nodes, including the intersection nodes themselves.  The geojson file could then be drawn on map in a Jupyter Notebook using the Python ``ipyleafelet'' library as per sections \ref{s:rq2} and \ref{s:rq3}.

In this exercise, lane detection was attempted from the dash camera footage only, as a demonstration of how the techniques developed to address the first three research questions could be re-used to gather other information visually.  It could be applied to Google Street View images, however it was only attempted with the dash camera footage, where it is easier to ensure that each image is taken with a heading that is consistent with the direction of the road.

\chapter{Results and Discussion}
\label{s:results}

\section{RQ1: Training a model to identify bicycle lanes in Google Street View images}
\label{results:rq1}

Initial performance targets of 80\% recall and 80\% precision were chosen for this research question, to provide a solid baseline before attempting to infer and map routes from the detection points as part of the subsequent phases of research.  To infer routes from detection points with the method proposed in section \ref{s:rq2}, we need to be confident that bicycle lane markings will be detected at most intersections, if present, otherwise the missed detections would cause an incorrect discontinuity in the inferred route.  A recall of  80\% would mean that we can expect to detect the markings most of the time.  We also want to limit false positives, because two adjacent false positives could cause us to infer a route where there is none.  A precision of 80\% would mean that only a small minority of our detections are false positives, and we would be very unlucky to infer a long route spanning multiple intersections that is not real.

An initial dataset of 256 Google Street View images was gathered, with the option to gather more, if necessary, to achieve acceptable performance for the first two research questions.  They were labelled and split into 205 ``training''  images and 51 ``test'' images, according to an 80:20 split.  One of the 51 ``test'' images did not contain any bicycle lane markings.  It was initially included in the dataset in error, however it was retained because it proved useful for quick sanity checks, to ensure the model was not blindly flagging bicycle lane markings in every image.

Five pre-trained object detection models from the TensorFlow 2 Model Garden were trained and evaluated using these ``training'' and ``test'' datasets, through a process of transfer learning.  Each candidate model was trained for 2,000 steps at a time, with training paused every 2,000 steps to evaluate performance against the ``test'' dataset.  Training continued until it yielded no significant improvement to the mean Average Precision of bounding box detection when evaluated against the ``test'' dataset.  

Table \ref{table1} lists the five pre-trained object detection models that were tested, and the mean Average Precision of their bounding box detection when evaluated on the ``test'' dataset after 2,000 training steps.  The ``EfficientDet D1 640x640'' and ``SSD ResNet101 V1 FPN 640x640'' were unable to detect any bicycle lane markings even after 4,000 training steps, so training on them was ceased, to focus on the other three models that were producing more immediate results.

\begin{table}[h]
\centering
\begin{tabular}{|l|r|r|r|}
\hline
\multicolumn{1}{|c|}{\bfseries Pre-Trained Model} & \textbf{Steps} & \textbf{Evaluation mean Average Precision} \\
\hline
CenterNet HourGlass104 512x512 & 2,000 & 0.691116 \\
Faster R-CNN ResNet50 V1 640x640 & 2,000 & 0.476506 \\
SSD MobileNet V1 FPN 640x640 & 2,000 & 0.599674 \\
EfficientDet D1 640x640 & n/a & n/a \\
SSD ResNet101 V1 FPN 640x640 & n/a & n/a \\
\hline
\end{tabular}
\caption{Bounding Box mean Average Precision per pre-trained model (2,000 training steps)}
\label{table1}
\end{table}

\begin{figure}[h]
\centering
\begin{subfigure}{0.45\textwidth}
	\includegraphics[width=\textwidth]{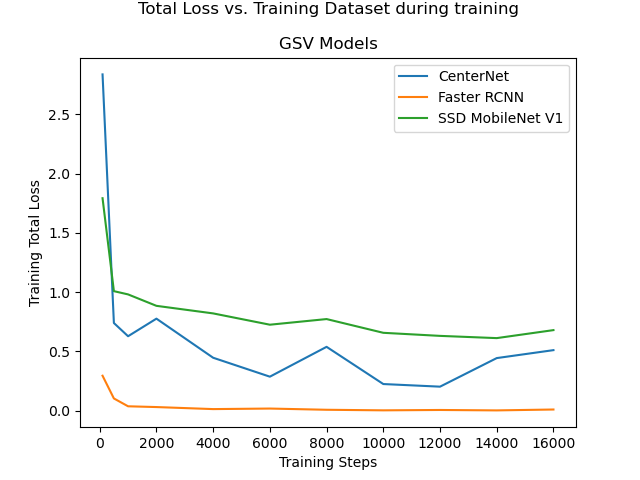}
	\caption{`Training ``Total Loss''}
	\label{fig:rq1a}
\end{subfigure}
\hfill
\begin{subfigure}{0.45\textwidth}
	\includegraphics[width=\textwidth]{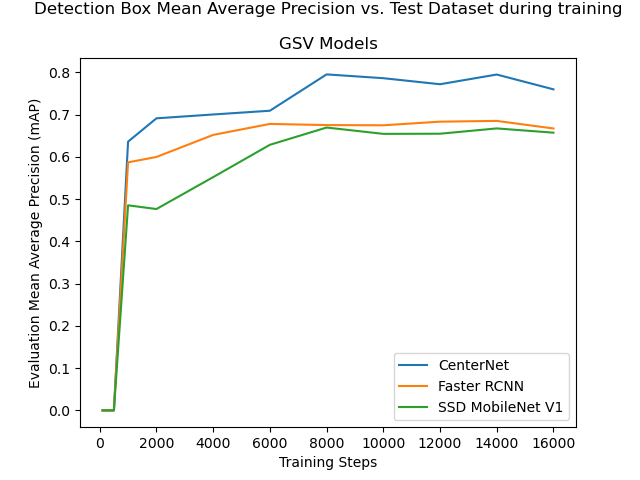}
	\caption{Evaluation ``mean Average Precision''}
	\label{fig:rq1b}
\end{subfigure}
\caption{Training and Evaluation of Detection Models based on Google Street View images}
\label{fig:rq2}
\end{figure}

Figure \ref{fig:rq1a} shows the ``Total Loss'' reported during training of the three remaining models.  All three models reached a point of diminishing returns at 2,000 training steps.

Figure \ref{fig:rq1b} shows the Bounding Box mean Average Precision during evaluation with the ``test'' dataset.  Continuing the training beyond 2,000 training steps yielded small improvements in that metric.  However, the most important measure of success for this model was whether it could correctly identify whether a bicycle lane marking appeared in an image or not.  It was not necessary to know precisely where the bounding box was, so improvements to the Bounding Box mean Average Precision are only useful up to a point.  Using the tool described in appendix \ref{a08}, each of the three models were applied to the ``training'' dataset and the ``test'' dataset, to inspect its predictions for each image.  The images were examined to confirm that the bounding boxes were drawn in a sensible location, and to confirm whether the model had successfully found a bicycle lane marking in each image.  The results are recorded in table \ref{table2}.

\begin{table}[h]
\centering
\begin{tabular}{|l|r|r|r|r|r|}
\hline
\multicolumn{1}{|c|}{\bfseries Pre-Trained Model} & \multicolumn{2}{|c|}{\bfseries Train} & \multicolumn{2}{|c|}{\bfseries Test} \\
& \textbf{False Pos$^n$} & \textbf{False -ve} & \textbf{False Pos$^n$} & \textbf{False -ve} \\
\hline
CenterNet HourGlass104 512x512 & 0 & 11 & 0 & 13 \\
Faster R-CNN ResNet50 V1 640x640 & 1 & 0 & 4 & 0 \\
SSD MobileNet V1 FPN 640x640 & 8 & 30 & 4 & 9 \\
\hline
\end{tabular}
\caption{Errors in the detection of bicycle lane markings in the ``train'' and ``test'' datasets after 2,000 training steps.  (The number of frames with a bounding box drawn in a false position, and the number of frames where the model came to a false negative conclusion.)}
\label{table2}
\end{table}

The ``Faster R-CNN ResNet50 V1 640x640'' model was able to successfully locate a bicycle lane marking in \textit{every single image in ``training'' and ``test'' datasets where one appeared}.  It was not fooled by a solitary image with \textit{no bicycle lane markings} that had been included in the ``test'' dataset.  It produced a bounding box in the wrong position for four images in the test dataset.  It easily met the target performance metrics, with 100\% recall and 92\% precision.

The ``Faster R-CNN ResNet50 V1 640x640'' model, trained to 2,000 training steps, was chosen as the preferred model, and used to create a bicycle lane route maps for research question 2.

The initial dataset of 256 images appeared to be sufficient to get reliable detection results with this model, so no further images were collected from Google Street View for the ``training'' and ``test'' datasets at this stage.  If better performance was required to obtain sensible results when drawing maps of bicycle routes in the next section, we would address that by increasing the number of training steps, the number of training images available, add labels to common sources of false positive results, then re-assess the candidate models after further training.

\section{RQ2: Building a map of bicycle lane routes from Google Street View images in a survey area}
\label{results:rq2}

\subsection{Mount Eliza}
\label{results:rq2a}

The model that was selected in section \ref{results:rq1} was applied to a sample survey area in the vicinity of Mount Eliza.  This area was chosen due to its accessibility:  It was one of very few areas with bicycle lanes that was reachable within the COVID-19 lockdown restrictions that were in place at the time \cite{lockdown_record} \cite{lockdown_5km}.  It was therefore possible to validate the results via an in-person survey, in addition to comparing the results to OpenStreetMap and the ``Principal Bicycle Network'' dataset.  It also allowed for comparison with results from dash camera footage in section \ref{results:rq3}.

A total of 1,113 locations were sampled from the survey area, at points within 20 metres of every intersection.  Four images were downloaded via the Google Street View API at each location, a total of 4,452 images at a cost of approximately \$31 USD.  The exact survey area within Mount Eliza was chosen to include a mix of roads with and without bicycle lanes, ranging from no-through roads to a highway.

\begin{figure}[h]
\centering
\includegraphics[width=0.45\textwidth]{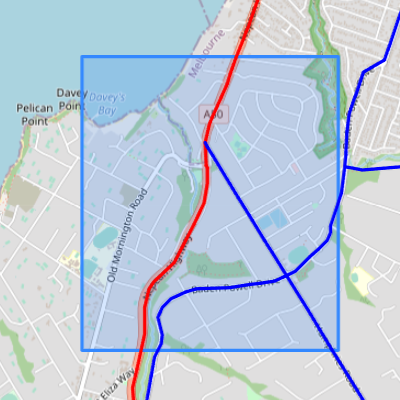}
\caption{Principal Bicycle Network existing and planned routes}
\label{fig:rq2c}
\end{figure}

Figure \ref{fig:rq2c} is a screenshot from an interactive map in the Jupyter Notebook described in appendix \ref{a11}, showing existing and planned bicycle routes in the survey area according to the ``Principal Bicycle Network'' dataset.  The blue bounding box shows the survey area.  The red lines show ``existing'' routes, and the blue lines show ``planned'' routes.  The ``Principal Bicycle Network'' dataset is out of date.  It correctly records that there is an existing route on the Nepean Highway, in red.  However, the long, straight ``planned'' route, Humphries Road, has existed for at least several years, and so has the route to the north-east of it, on Baden Powell Drive.  The planned route on Baden Powell Drive to the west of Humphries Road does not yet exist.

\begin{figure}[h]
\centering
\begin{subfigure}{0.45\textwidth}
	\includegraphics[width=\textwidth]{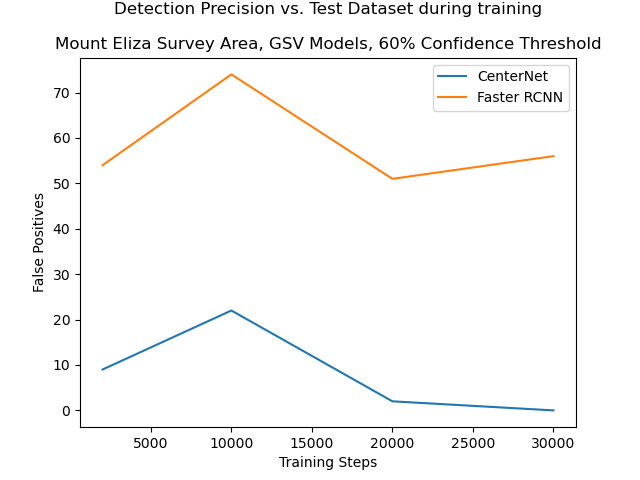}
	\caption{False Positives}
	\label{fig:rq2a}
\end{subfigure}
\hfill
\begin{subfigure}{0.45\textwidth}
	\includegraphics[width=\textwidth]{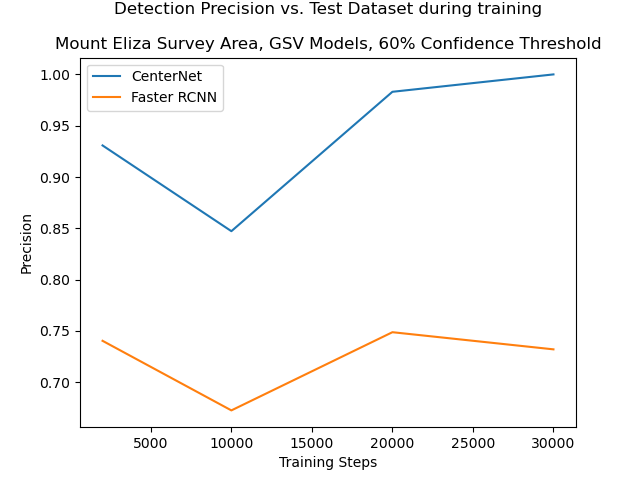}
	\caption{Precision}
	\label{fig:rq2b}
\end{subfigure}
\caption{Extending model training steps to deal with false positives}
\label{fig:centernet}
\end{figure}

In section \ref{results:rq1}, the models were trained for 2,000 training steps, beyond which further training did not appear to yield any improvements when the model was evaluated against the small 51 image ``test'' dataset.  The Faster R-CNN model appeared to be the most promising.  However, when that model was applied to Google Street View images from the Mount Eliza survey area, the initial results were poor.  Precision was significantly lower than it had been in the initial evaluation, with many false positive detections for random objects such as bushes, or leaves on the road.  The models received further training on the original ``training'' dataset, up to 30,000 training steps, and the CenterNet model had the best performance, producing zero false positive detections after 30,000 training steps.  It was therefore used to construct maps from the Google Street View images in the survey areas.  See figure \ref{fig:centernet} for training statistics.

\begin{figure}[h]
\centering
\begin{subfigure}{0.45\textwidth}
	\includegraphics[width=\textwidth]{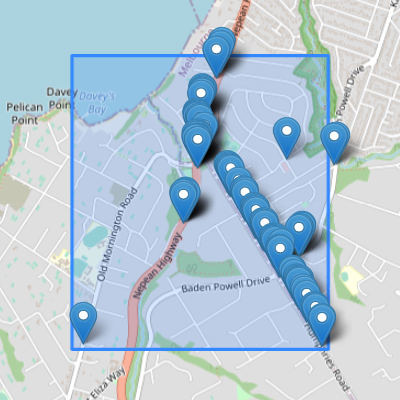}
	\caption{Detection Points within survey area}
	\label{fig:rq2a2}
\end{subfigure}
\hfill
\begin{subfigure}{0.45\textwidth}
	\includegraphics[width=\textwidth]{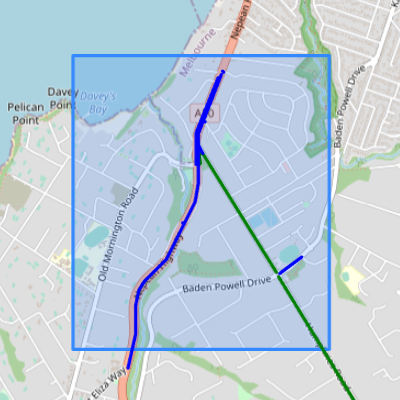}
	\caption{Detected routes c.f. OpenStreetMap}
	\label{fig:rq2b2}
\end{subfigure}
\caption{Map of bicycle lanes in Mount Eliza based on Google Street View images}
\label{fig:rq22}
\end{figure}

Figure \ref{fig:rq2a2} shows the individual points that were detected by the model from Google Street View images sampled near intersections in the survey area.  Figure \ref{fig:rq2b2} shows the bicycle lane routes that were inferred from those detections, according to the method described in section \ref{s:rq2c}.  The routes are colour-coded to highlight areas of agreement and disagreement between the detection model and OpenStreetMap.  Where both the detection model and OpenStreetMap agree that there is a bicycle lane, routes are coloured green.  They agree that there is a bicycle lane on Humphries Road.  Routes that were detected by the model but not recorded with a ``cycleway'' tag in OpenStreetMap are coloured blue.  The OpenStreetMap data is missing the cycleway on the Nepean Highway, and the fragment of the Baden Powell Drive cycleway that was detected near the boundary of the survey area.  Routes that are recorded in OpenStreetMap but were not detected would be coloured red, however there were no such routes in this survey.

Table \ref{table_metres_rq2_me} shows a comparison between the routes detected for Mount Eliza from the Google Street View images to what is recorded in OpenStreetMap, in metres.  The two sources agreed upon 2,344 metres of bicycle lane routes.  The detection model found an additional 2,787 metres that were not recorded in OpenStreetMap, including parts of the divided Nepean Highway where both sides of the road were counted due to them being mapped in OpenStreetMap separately.  Distances are approximate due to the level of precision available when calculating distances between each point in the geojson files.
\begin{table}[h]
\centering
\begin{tabular}{|l|r||}
\hline
\textbf{Route source} & \textbf{Metres} \\
\hline
Detection Model & 5,216m \\
OpenStreetMap & 2,344m \\
Both & 2,344m \\
Detection model only & 2,787m \\
OpenStreetMap only & 0m \\
\hline
\end{tabular}
\caption{Comparison of routes detected in Mount Eliza to OpenStreetMap, GSV images}
\label{table_metres_rq2_me}
\end{table}

Based on an in-person survey of the area, and inspection of the Google Street View images that were downloaded, the detection model correctly detected all bicycle lanes within the survey area, with no false positives.  It identified two routes that had not been recorded in OpenStreetMap.

\subsection{Heidelberg}

\begin{figure}[h]
\centering
\begin{subfigure}{0.45\textwidth}
	\includegraphics[width=\textwidth]{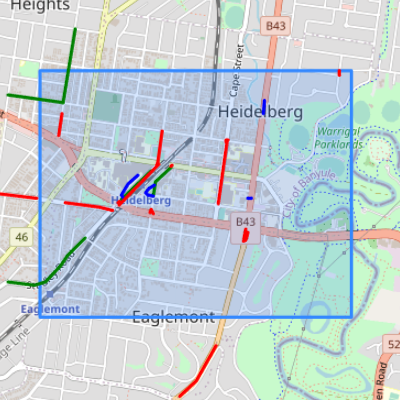}
	\caption{Overview of survey area}
	\label{fig:rq2a-h}
\end{subfigure}
\hfill
\begin{subfigure}{0.45\textwidth}
	\includegraphics[width=\textwidth]{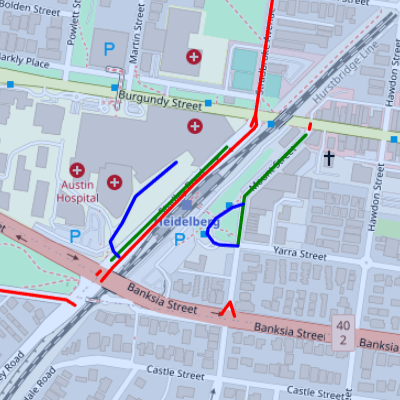}
	\caption{Magnified view}
	\label{fig:rq2b-h}
\end{subfigure}
\caption{Map of bicycle lanes in Heidelberg based on Google Street View images}
\label{fig:heidelberg_good}
\end{figure}

After surveying Mount Eliza with the Google Street View approach, the exercise was repeated for a survey area in Heidelberg, Victoria.  The purpose of this exercise was to assess how well the detection models generalised to a more heavily built-up area, closer to the city.  It included both residential and commercial areas.  See figure \ref{fig:heidelberg_good} for the results.

\begin{figure}[h]
\centering
\begin{subfigure}{0.45\textwidth}
	\includegraphics[width=\textwidth]{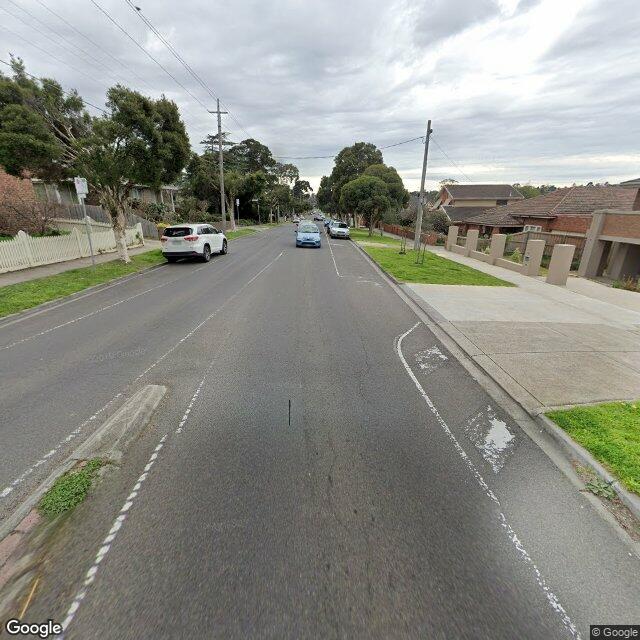}
	\caption{Cape Street, Heidelberg, May 2019}
	\label{heidelberg_a}
\end{subfigure}
\hfill
\begin{subfigure}{0.45\textwidth}
	\includegraphics[width=\textwidth]{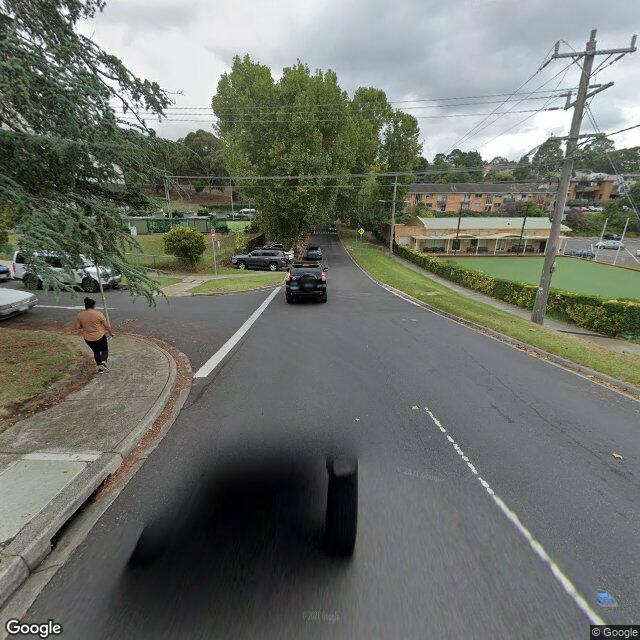}
	\caption{Stradbroke Street, Heidelberg, March 2021}
	\label{fig:heidelberg_b}
\end{subfigure}
\caption{Example Bicycle lane routes that are tagged in Heidelberg but not apparent in the Google Street View images}
\label{fig:heidelberg_osm_errors}
\end{figure}

The red lines in the map again represent routes that are recorded in OpenStreetMap but not detected.  Any red lines outside the bounding box describing the survey area can be ignored.  The Google Street View images for the red lines within the survey area were examined manually, to assess whether the model had produced sensible results based on the input.  Zooming into the red diagonal line on Studley Road near the Heidelberg train station, it can be seen that the model did correctly predict that there was a bicycle lane, but it only marked it on one side of the road, whereas it is tagged separately on both sides of a divided road in OpenStreetMap.  The remainder of the red lines are routes where there is no bicycle lane visible in the Google Street View images, and the model has produced a sensible result.  See figure \ref{fig:heidelberg_osm_errors} for example images from Cape Street and Stradbroke Street.  It is possible that this is caused by out-of-date Google Street View images, however the images for Stradbroke Street were only six months old, and the streets did not seem wide enough to easily accommodate the addition of a bicycle lane.  These routes might be errors in the OpenStreetMap data.  An on-site survey by a local resident with access to the area confirmed this.

The blue lines near the Austin Hospital and Heidelberg Station are false positives.  They are caused by the road being close enough to see a bicycle lane marking on a nearby road.

There is a very short segment where OpenStreetMap records a cycleway on Lower Heidelberg Road near the intersection of Banksia Street.  This route would be barely larger than the nearby intersection, and it is almost certainly an error in OpenStreetMap.

A comparison between the improved detection model results and OpenStreetMap in terms of route lengths can be found in table \ref{table_metres_rq2_h}.

\begin{table}[h]
\centering
\begin{tabular}{|l|r||}
\hline
\textbf{Route source} & \textbf{Metres} \\
\hline
Detection Model & 4,563m \\
OpenStreetMap & 5,317m \\
Both & 2,094m \\
Detection model only & 469m \\
OpenStreetMap only & 2,536m \\
\hline
\end{tabular}
\caption{Comparison of routes detected in Heidelberg to OpenStreetMap, GSV images}
\label{table_metres_rq2_h}
\end{table}

The model was able to identify all bicycle lane routes that were recorded in OpenStreetMap \textit{and visible in the Google Street View images}.  It produced 469 metres worth of false positives, due to bicycle lane markings on one road being visible from multiple points on a nearby road.  An approach using dash camera footage could help to address this issue, by ensuring that the camera is always correctly oriented to face the direction of the road, thereby allowing detection masks to be used to eliminate bicycle lane markings from nearby roads.

\section{RQ3: Building a map of bicycle lane routes from dash camera footage captured in a survey area}
\label{results:rq3}

\subsection{Re-training the model for dash camera footage}

Due to COVID-19 restrictions \cite{lockdown_record} \cite{lockdown_5km}, dash camera footage could only be gathered in a circular area with a 15km radius, much of which was a body of water.  Therefore, the range of areas available to gather dash camera footage for research question 3 was restricted to a handful of outer metropolitan suburbs.  Footage from the Frankston, Langwarrin and Baxter area was used to train a the detection model to work with dash camera footage, and footage from Mount Eliza was used to test it.

Training was conducted as per the approach documented in section \ref{s:rq3}.  Initially, the original model from section \ref{results:rq2} that had been trained exclusively on Google Street View images was used to process the dash camera footage from the training area.  The initial positive matches -- whether true positives or false positives -- were added to the ``training'' and ``test'' datasets according to an 80:20 split.  The resulting ``training'' dataset consisted of 205 Google Street View images and 252 dash camera images.  The ``test'' dataset consisted of 51 Google Street View images and 85 dash camera images.

\begin{figure}[h]
\centering
\begin{subfigure}{0.45\textwidth}
	\includegraphics[width=\textwidth]{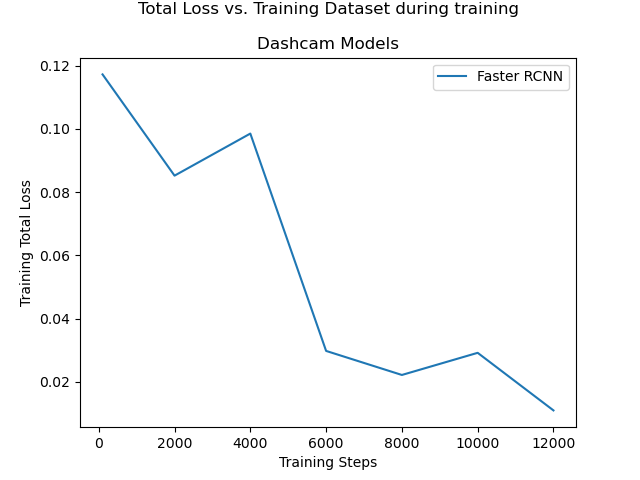}
	\caption{`Training ``Total Loss''}
	\label{fig:rq3a}
\end{subfigure}
\hfill
\begin{subfigure}{0.45\textwidth}
	\includegraphics[width=\textwidth]{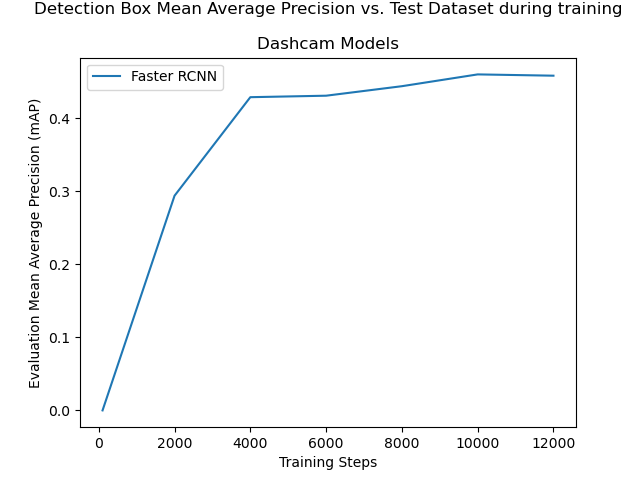}
	\caption{Evaluation ``mean Average Precision''}
	\label{fig:rq3b}
\end{subfigure}
\caption{Training and Evaluation of Detection Models based on dash camera images}
\label{fig:rq3aa}
\end{figure}

Another ``Faster R-CNN ResNet50 V1 640x640'' model was then trained and evaluated using the combined Google Street View + dash camera footage datasets.  To deal with remaining false positives, the dash camera images were labelled with additional classes to distinguish common misleading features from bicycle lane markings, and a detection mask was applied to remove reflections from the bonnet of the car, and distractions from the right hand side of the road.  Please see section \ref{s:rq3} for a full discussion of these enhancements.  The results observed from the model after every 2,000 training steps are recorded in figure \ref{fig:rq3aa} and table \ref{table3}.

\begin{table}[h]
\centering
\begin{tabular}{|l|r|r|r|r|r|r|r|}
\hline
\textbf Model & \textbf{Steps} & \multicolumn{3}{|c|}{\bfseries Train} & \multicolumn{3}{|c|}{\bfseries Test} \\
& & \textbf{Total Loss} & \textbf{False +ve} & \textbf{False -ve} & \textbf{mAP} & \textbf{False +ve} & \textbf{False -ve} \\
\hline
Faster R-CNN & 2,000 & 0.0852 & 1 & 8 & 0.293945 & 3 & 9 \\
Faster R-CNN & 4,000 & 0.0985 & 0 & 8 & 0.428875 & 1 & 2 \\
Faster R-CNN & 6,000 & 0.0298 & 0 & 0 & 0.430906 & 3 & 3 \\
Faster R-CNN & 8,000 & 0.0222 & 0 & 0 & 0.443913 & 1 & 2 \\
Faster R-CNN & 10,000 & 0.0292 & 0 & 0 & 0.46007 & 2 & 2 \\
Faster R-CNN & 12,000 & 0.011 & 0 & 0 & 0.458306 & 0 & 1 \\
\hline
\end{tabular}
\caption{Training and evaluation results for the Faster RCNN model with dash camera footage}
\label{table3}
\end{table}

Figures \ref{fig:rq3a} and \ref{fig:rq3b} show that there were diminishing returns from further training beyond 4,000 to 6,000 training steps, in terms of both the ``Total Loss'' reported in the training process, and the ``mean Average Precision'' of the bounding boxes reported from the evaluation process against the ``test'' dataset.

\begin{figure}[h]
\centering
\begin{subfigure}{0.3\textwidth}
	\includegraphics[width=\textwidth]{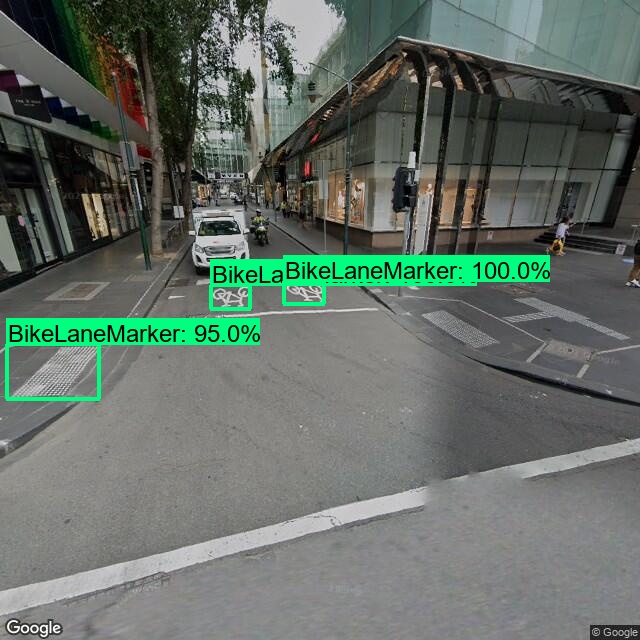}
	\caption{False Positive GSV image example}
	\label{rq3c-a}
\end{subfigure}
\hfill
\begin{subfigure}{0.3\textwidth}
	\includegraphics[width=\textwidth]{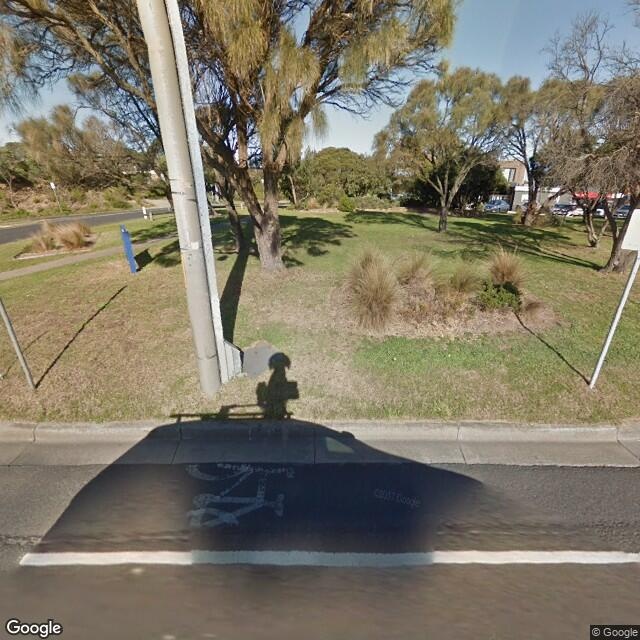}
	\caption{False Negative GSV image example 1}
	\label{rq3c-b}
\end{subfigure}
\hfill
\begin{subfigure}{0.3\textwidth}
	\includegraphics[width=\textwidth]{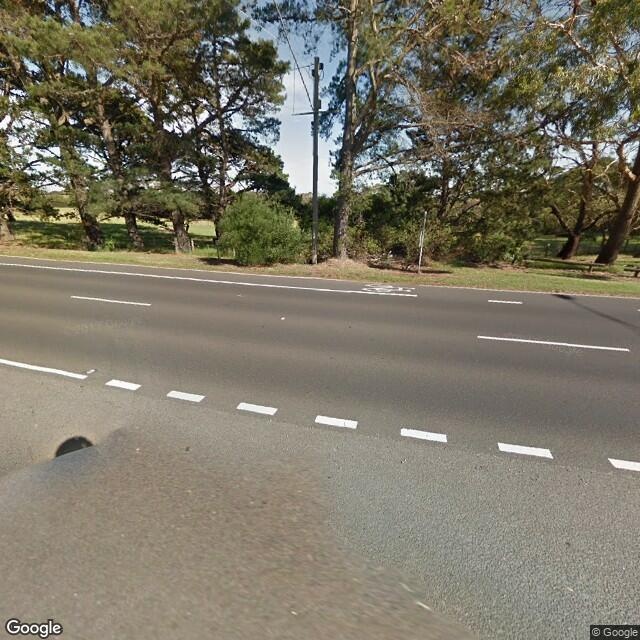}
	\caption{False Negative GSV image example 2}
	\label{rq3c-c}
\end{subfigure}

\caption{False Positive and False Negative results recorded when applying the preferred Faster R-CNN model to the combined Google Street View and dash camera ``test'' dataset, after 8,000 training steps}
\label{fig:rq3ccc}
\end{figure}

The model was applied to the ``training'' and ``test'' images via the tool documented in appendix \ref{a15} to examine the bounding boxes it chose to draw, and whether it believed there was a detection or not.  The results of this review are found in the ``False +ve'' and ``False -ve'' columns reported in table \ref{table3}.  After 6,000 training steps, the model was able correctly detect all bicycle lane images in the ``training'' data, with no false positives, false negatives, or misplaced bounding boxes.  In the test data, there were a total of 3 false positives and 3 false negatives out of 136 test images.  These false results came from  a mix of Google Street View images and dash camera images.  After 8,000 training steps, the performance improved further, in that there was 1 false positive and 2 false negatives when analysing the test data, and these were all for challenging Google Street View images.  See figure \ref{fig:rq3ccc}.

In figure \ref{rq3c-a}, the model has correctly identified two bicycle lane markings in the central business district of Melbourne, but it has also identified a white strip that was installed at a pedestrian crossing to assist people with impaired vision.  A very high confidence of 95\% was assigned to the false positive detection.  To assist the model to deal with this situation better, more training images would be required, with these objects given a label to steer the model away from thinking of them as a bicycle lane marking.

In figure \ref{rq3c-b}, the model has missed a very faint bicycle lane marking that was alongside the camera vehicle, in the shadow of the vehicle.  This was a very challenging image due the the faintness of the marking, and it was common for the models tested in this research project to miss it.

In figure \ref{rq3c-c}, the model has missed a bicycle lane marking on the opposite side of the road, at a challenging angle and distance.  In general, most models struggled with this test image as well.

Training was terminated after 8,000 steps, and the model was then used to conduct a survey from dash camera footage in the ``test'' area of Mount Eliza.  At 8,000 training steps, the only incorrect predictions reported were for Google Street View images.  For this section, the model was being applied to images from dash camera footage, where there were many more frames available to compensate for any missed detections in individual frames.

\subsection{Generating and comparing maps}

\begin{figure}[h]
\centering
\includegraphics[width=0.45\textwidth]{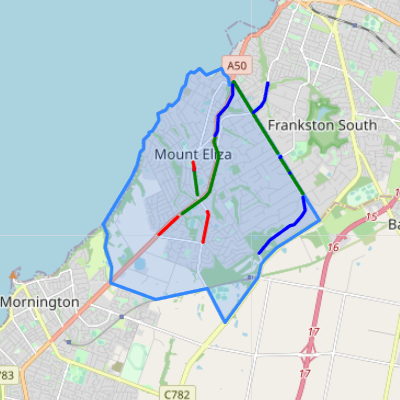}
\caption{Detected bicycle lanes in Mount Eliza based on dash camera images}
\label{fig:rq3}
\end{figure}

Figure \ref{fig:rq3} shows a comparison between the bicycle lanes in Mount Eliza that were detected based on dash camera images, and the routes that are recorded in OpenStreetMap as having bicycle lanes, for roads that were surveyed in the footage only.  The green routes show where there is agreement between the detection model and OpenStreetMap.  The blue routes show bicycle lanes that were detected but not recorded in OpenStreetMap.  The red routes show bicycle lanes that are recorded in OpenStreetMap but not detected.

The model correctly detected three sections that are not recorded in OpenStreetMap.  There is a cycleway on the Nepean Highway that has not been tagged in OpenStreetMap in the northern section depicted on the map.  There is a narrow cycleway on Baden Powell Drive to the north-east of Humphries road.  And there is a cycleway on Two Bays Road to the south-east of the survey area.

There is a short segment in the northern section of Mount Eliza Way where a bicycle lane is tagged in OpenStreetMap but was not detected by the model.  The detection model appears to be returning a reasonable result in this instance, the bicycle lane does not extend beyond the roundabout at Kenaud Avenue.

There is a segment on Wooralla Drive where a bicycle lane is tagged in OpenStreetMap.  This is an error, there is no bicycle lane on that road.

There is a longer segment on the Nepean Highway, leading up to the intersection with Tower Road, where the detection model failed to draw a bicycle route.  This is the only part of the map where the detection model appears to have made an error.  During the survey, the camera vehicle turned into Tower Road and missed that segment.  If the vehicle had continued along Nepean Highway, it might have picked up the next bicycle lane marking and drawn a continuous route.

A comparison between the improved detection model results and OpenStreetMap in terms of route lengths can be found in table \ref{table_metres_rq3_me}.

\begin{table}[h]
\centering
\begin{tabular}{|l|r||}
\hline
\textbf{Route source} & \textbf{Metres} \\
\hline
Detection Model & 13,413m \\
OpenStreetMap & 8,903m \\
Both & 7,484m \\
Detection model only & 5,795m \\
OpenStreetMap only & 9,67m \\
\hline
\end{tabular}
\caption{Comparison of routes detected in Mount Eliza to OpenStreetMap, dash camera images}
\label{table_metres_rq3_me}
\end{table}

Overall, the detection model performed well within the survey area.  It is possible that multiple surveys spread over multiple weeks would have filled in the gap at the edge of the survey area, on the Nepean Highway.  The model detected three route segments that are not recorded on OpenStreetMap, and correctly identified two route segments that appear to be recorded in OpenStreetMap in error.  It missed one route segment that was recorded in OpenStreetMap, at the edge of the survey area, but perhaps repeat surveys or a survey route that extended further down the Nepean Highway may have caught this.   Both OpenStreetMap and the detection model included two routes that are not listed in the ``Principal Bicycle Network'' dataset as existing routes.

\section{RQ4: Surveying other infrastructure details using dash camera footage}
\label{results:rq4}

\subsection{Tuning the Paved Shoulder Detection Model thresholds}

The model proposed in section \ref{s:rq4} was first applied to the training area footage from Frankston, Langwarrin, and Baxter, and then validated against the same test area footage from Mount Eliza as used for research question 3.

It was found that, when processing an individual frame, the proposed method was vulnerable to ``noise'' from objects just outside the paved area, and this led to the use of thresholds described in section \ref{s:rq4c}.  After analysing the observed data and ground truth for the ``training'' area, the following thresholds were set:

\begin{itemize}
\item{A minimum of 80\% of frames in a road segment must have a paved shoulder detected}
\item{The mean width of paved shoulders detected in a road segment, measured at the upper top horizontal line drawn in the overlay, must be at least 75 pixels}
\item{The standard deviation of the X and Y co-ordinates at which the paved shoulder boundary lines would intersect must be less than or equal to 50 pixels}
\end{itemize}

These thresholds were intended to filter out road segments with inconsistent detections, paved shoulders that are too narrow to be used by a cyclist, and false positive detections where the perceived paved shoulder boundaries jump around wildly from frame to frame.

\subsection{Results for the test area}

\begin{figure}[h]
\centering
\includegraphics[width=0.45\textwidth]{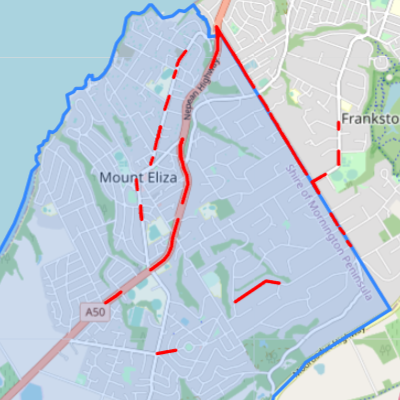}
\caption{Detected paved shoulders in Mount Eliza based on dash camera images}
\label{fig:rq4}
\end{figure}

The map of paved shoulders that was generated for the test area of Mount Eliza from the dash camera footage can be found in figure \ref{fig:rq4}.

The map gives relatively robust results on Humphries Road and the Nepean Highway, where there is a bicycle lane, a special case of a paved shoulder.  Part of the Nepean Highway was not detected, where the lane markings have faded and the ``noise'' from the foliage dragged the boundary of the lane detected for camera vehicle itself too far to the left.

The other lines drawn on the map represent true paved shoulders, but they were easily interrupted by ``noise'' from the side of the road, or parked cars.  Many of the interruptions in the paved shoulder route on Old Mornington Road, to the west of the the Nepean Highway, were attributable to parked cars.

The model did not present any false positives, and it appeared to find most roads with paved shoulders, but the routes drawn were often broken up.

\section{Limitations}
\label{results:discussion}

\subsection{Use of Google Street View images}

The use of Google Street View images to map bicycle lane routes in Mount Eliza was effective, and added value, finding routes that had not already been recorded in either the official ``Principal Bicycle Network'' dataset or the crowdsourced OpenStreetMap database.  Repeating the exercise in the more heavily built-up suburb of Heidelberg was less effective:  There were some false positives, and every real route that was identified was already included in the OpenStreetMap database.

The main limitations of the Google Street View approach are the cost of images and gaps in coverage they can provide.

The Google Street View API appears to be able to provide a distinct snapshot every 10 metres.  A town or suburb can easily have many kilometres of roads, and if every snapshot in a town were sampled, or multiple towns were surveyed, the amount payable to Google for the use of their API could add up quickly.  The proposed solution attempted to mitigate this by only sampling the area immediately around intersections, and making an assumption that a bicycle lane route continues interrupted if detections are missed at only one or two intersections in a row.  If API costs need to be reduced further, then perhaps OpenStreetMap tags could be used to eliminate roads that are for ``local traffic only'', where bicycle lanes might not be as necessary.  Or, the survey could focus on roads that are adjacent to known bicycle lane routes, where continuations of the known routes are waiting to be discovered.

One significant challenge is the recency of the Google Street View images available.  For example, the latest available Google Street View images for the Mount Eliza survey area are dated October 2019.  If any new bicycle lanes had been constructed in the past two years, the model would have been unable to detect them.  The Google Street View approach can only bring a dataset of bicycle lane routes up to date as far as Google's last visit to the area.

The Google Street View camera vehicle typically only provides a view from one lane, on one side of the road.  Bicycle lane markings on the other side of the road may be difficult to spot, especially for a divided road with foliage or other barriers dividing the traffic.  This is especially important if the bicycle lane route map must show which sides of the road a bicycle lane is present.  It may be impossible to detect a bicycle lane on a service road running in the opposite direction to the lane from which the Google Street View snapshot was taken.  Hopefully these challenges are sometimes mitigated by taking samples near intersections from multiple approaches, where some of the bicycle lane markings might be clear.

The model was able to detect bicycle lane markings even when they were moderately far into the distance, despite the 640x640 resolution limit.  Sometimes, the bicycle lane marking was ``smeared'' by artefacts in the Google Street View image, but with enough training examples, the model was able to detect these correctly.

The approach of using a 360 degree view at each intersection to infer routes was vulnerable to false positives where bicycle lane markings were visible from a separate road, nearby.  Using front-facing footage from a dash camera and applying detection masks could resolve this issue.

The survey area of Mount Eliza is in an outer metropolitan area, and the COVID-19 lockdown restrictions have triggered political debate about whether it should even be considered ``metropolitan'' at all.  Most residential roads in the area do not have pedestrian footpaths.  When the model is applied to higher-density suburbs, such as Heidelberg, it is likely to need more images in the ``training'' and ``test'' datasets to cover the additional distractions that might appear on the side of the road.  With more traffic on the roads, it might be much more difficult to detect a bicycle lane marking on the opposite side of a road or intersection.  There might therefore be a much higher reliance on detecting markings immediately in front of and to the left of the camera vehicle, where there is less chance of another vehicle obscuring a bicycle lane marking.

Difficulties in heavily built-up areas might matter less if the bicycle lane routes in those areas have already been well mapped due to their priority and a large number of interested cyclists who could contribute to OpenStreetMap.

\subsection{Use of dash camera images}

The use of dash cameras to gather images would overcome many of the limitations of the Google Street View data.  If the latest available images are too old, then there is freedom to go and collect more at any time.  Images can be gathered from both sides of the road, from any lane or service road required.  The driver can drive conservatively, to limit the degree to which the view is obstructed by a vehicle in front.  Many more snapshots are available:  If the camera vehicle is driving at 50kmph, that equates to just under 14 metres per second.  A camera recording at 60 frames per second would record a frame every 23cm, compared to the 10 metre intervals provided between snapshots by Google Street View.  The limiting factor becomes how much data is meaningful to store and review, and missed detections may be compensated for by analysing nearby frames, or through repeat surveys of the area over time.

The results obtained from the dash camera survey were at least the equal of the Google Street View survey of a smaller sample area.  The dash camera approach has the potential to be more robust, and better able to deal with the technical challenges that limit the Google Street View approach.

The main challenges with the dash camera approach involve logistics.  With Google Street View, you can survey as wide an area as you can afford to download.  With the dash camera approach, someone must install and maintain a dash camera, drive the required routes, and send the data off for processing.  This could be done by interested volunteers for individual areas, or by individual local governments who are aware that they have significant deficiencies in their data.  Perhaps the most promising prospect would be to ``incidentally'' gather footage from local government vehicles that regularly drive most routes for another purpose.  For example, garbage trucks would cover all residential streets on a regular basis.  The business case for collecting, managing, and processing the footage could be justified not only by the value of mapping bicycle lane routes, but by the value of any other information that could be gathered visually and correlated to a location.  For example, a local council might be interested in knowing whether the garbage truck \textit{really} missed someone's bin, or whether the truck is being sent back because the resident actually forgot to put their bin out.  Or they may be interested in monitoring local roads for defects that need to be repaired.

To scale the dash camera approach across multiple cameras, or to reproduce the research, it should be noted that the optical characteristics, resolution, and frame rate of each camera may differ, and placement of the camera in different positions, angles, and vehicles may materially affect the perspective of the images.  These factors might result in the need to gather additional training and test images to ensure that the solution generalises well across cameras.  If a model is being used to correct for lens distortion, each camera must go through its own calibration process, as per section \ref{s:rq4a}.

The method that was used to quickly ``bootstrap'' a set of dash camera images for training and validation based on the Google Street View-trained model worked well in this instance.  But the process of randomly splitting those images into ``training'' and ``test'' datasets would not have ensured the absolute independence of the ``test'' images.  None of the ``test'' images were included in the ``training'' dataset, however some of the ``training'' images may have been very similar, due to them being taken from a nearby location.  If the model had failed to generalise well despite good performance in its evaluation, this may have needed more attention.

It is likely that more ``training'' and ``test'' images would be required to get good results in  inner metropolitan areas, but it was not possible to test this hypothesis.

\subsection{Detection of paved shoulders}

The model that was applied to the problem of detecting paved shoulders had significant limitations.  It was intended as a ``proof of concept'', to show that the general technique used to map bicycle lanes from the bicycle lane markings could be applied to other problems where information can be collected visually.

The model relied on explicit programming of an algorithm, and manually selected thresholds applied to summary statistics based on observations in a training area.  It did not attempt to apply any of the more recent advances in machine learning or deep learning, where the model ``learns'' to solve the problem for itself from the data.  It was able to create an approximate map, without false positives, but the detected routes were often interrupted by parked cars or roadside ``noise''.

These results were achieved by using training data from a suburb that was adjacent to the survey area.  They most likely shared many characteristics.  It might be difficult or impossible to tune the model to work well across a wider variety of settings.

The application of the Canny-Hough approach to lane detection across a variety of environments is challenging, and could be considered a separate research project in itself.  To get a more consistent detection of the road boundary in the face of road-side ``noise'', it would be worth exploring deep learning image segmentation techniques to train a model that can detect the road surface, similar to the approach taken by Mamidala et al., 2019 \cite{8929655}.  A deep learning model could also be trained to detect parked cars, and take them into account:  A parked car should not be allowed to interrupt a detected paved shoulder route, but if there are many parked cars on the shoulder, is it worth recording the route at all?

\subsection{Survey area constraints}

COVID-19 lockdown restrictions imposed significant constraints on the location and variety of survey areas.  Access to dash camera footage was especially limited.  For the Google Street View approach, the restrictions constrained the areas for which an absolute ``ground truth'' could be established via an in-person survey.  Further work would be required, once lockdown restrictions are eased, to show how well the dash camera solutions work in other environments.

During this research, surveys were conducted at a scale of one suburb at a time.  OpenStreetMap data for an area was parsed and then cached in memory for quick access.  It seemed reasonable to partition the work by suburb or ``Local Government Area'', and the methods used worked well at that scale.  Some operations might not scale well if they were applied to a whole state or country at a time.  For example, for every point sampled in dash camera footage, we find the closest ``way'' and then the closest ``node'' in OpenStreetMap, to align the data to the OpenStreetMap version of the shape of each road.  This allows us to identify and measure any differences in routes.  If the process were applied at a state level instead of a suburb level, these searches might take significantly longer.

\section{Opportunities for future research}
\label{s:future_work}

Future work could examine how well the proposed solutions generalise to other survey areas.

If the cost of using the Google Street View API is a concern, then other strategies to reduce API consumption could be investigated.  For example, the current solution takes four images at each location.  How much impact does it have if we only have the front and rear views?  If we approach an intersection from multiple angles, and the angles are not perfectly square, that might result in unnecessary requests for images with headings that are only a few degrees different from ones we already have.  How many API requests can we save by limiting this?  Can we make assumptions about which classes of road we should survey, based on their tags or other information in the OpenStreetMap database?  For example, a ``tertiary'' road might be a good candidate for a bicycle lane, but a minor road designed for local traffic only might not be.  How frequently are bicycle lanes tagged in the OpenStreetMap database for each class of road?  Would it be useful to limit our search to roads that are adjacent to known bicycle routes, where they might provide a useful but unrecorded continuation?

This research focussed on surveying one aspect of the road infrastructure that is known to influence bicycle safety and uptake:  the presence of a bicycle lane.  There are many other factors that could be surveyed visually.  With a more robust way to detect the edge of the road and the each lane, we could not only survey the paved shoulders, but also measure the width of lanes.  We could use another deep learning model to detect vehicles that are obstructing the bicycle lane, or rows of parked cars that present a hazard due to the chance that a door will suddenly open into the bicycle lane, or the chance that a car might suddenly pull out.  We could correlate the data to topographical maps or GPS altitude readings to identify road segments with a significant uphill or downhill slope.  We could look for road surface defects or debris in the bicycle lane that might suggest that maintenance is required.

The solution that was implemented has contributed some general approaches that could be reused for other applications beyond the domain of bicycle safety.  It provided:  a method to sample Google Street View images for a list of locations and review them for possible inclusion in a labelled dataset for the purposes or training or testing a deep learning model;  a method to sample Google Street View images for roads or intersections in a survey area;  a method to infer routes on a map from points where a model has detected something of interest;  and a method to align a route from dash cam footage and its GPS metadata to the closest route in OpenStreetMap to make comparisons.  Any application where there is potential value in recognising something visually, counting incidences, and drawing them on a map, could re-use some of the techniques used in this research project.

\chapter{Conclusion}
\label{conclusion}

A ``deep learning'' model was successfully trained to detect bicycle lane markings based on a dataset of 256 Google Street View images that were randomly sampled from intersections along known bicycle lane routes across the state of Victoria, Australia.  The model was based on the ``Faster R-CNN ResNet50 V1 640x640'' model that is published in the TensorFlow 2 Model Garden, pre-trained on the COCO 17 dataset.  After 2,000 training steps, it achieved 100\% recall and 92\% precision when evaluated against 51 test images that had been withheld during its training.  This easily satisfied the performance targets proposed in research question 2.

The model was applied to Google Street View images across a survey area in the outer metropolitan suburb of Mount Eliza, Victoria, Australia.  The initial results revealed more false positive detections than expected based on the earlier evaluation of the model.  This was addressed by switching to a ``CenterNet HourGlass104 512x512'' model that had been trained to 30,000 steps.  A process was created to correlate the detection points to OpenStreetMap data about the road network, to infer and map bicycle routes.  All bicycle routes in the area were successfully detected, including routes not previously identified in OpenStreetMap or the ``Principal Bicycle Network'' dataset, satisfying the aims of research question 2.  There were no false positive routes.  When the same process was re-applied to the more heavily built-up suburb of Heidelberg, the results were not as strong, with some  inferred routes that were not real.  Additional training data may be required to improve performance, and future research should examine whether better results could be obtained using dash camera footage.

The initial model that was trained on Google Street View images generalised well enough to detect many of the bicycle lane markings in dash camera footage collected from a training area.  These detections were used to gather dash camera images that were labelled and added to the ``training'' and ``test'' datasets, to further improve the performance of the model when processing dash camera footage.  The final model was able to correctly identify all bicycle lane routes in the survey area, from dash camera footage, including routes not previously recorded, with no false positive routes.  It therefore satisfied the aims of research question 3, and it shows promise working in other outer metropolitan areas where bicycle lane routes are not already well mapped.  It was not possible to test the model in more heavily built-up areas, due to movement restrictions in place as part of the government response to COVID-19.

Finally, it was found that the approach that was used to map bicycle lane routes from dash camera footage could be re-used to visually survey other details about the road infrastructure.  As a demonstration, a Canny-Hough based model was created to detect paved shoulders on the side of the road, suitable for a cyclist to use, to maintain some separation from other traffic.  The model was able to create and draw a map.  However there were issues with gaps in the routes drawn, due to interference from parked cars and other roadside ``noise''.  With future research, it would be possible to plug in a deep learning model to improve performance.  Or to plug in a different model to detect something else entirely, for another application.  This satisfied research question 4.


\appendix

\chapter{Process documentation}
\label{a:process}

To assist with reproduction of results, code is provided at GitHub at:

\url{https://github.com/tylerrmit/minor_thesis.git}

Data necessary to reproduce the results, including dash camera footage collected during the experiments, can be found in an archive on \url{figshare.com} under the Digital Object Identifier (DOI):

\url{10.25439/rmt.16862518}

The code consists of a series of Jupyter Notebooks for high-level operation of the process, supported by Python classes for implementation details.  The Jupyter Notebooks are numbered to help the user through the proper sequence, and they are documented below.

Dash camera MP4 footage and NMEA metadata that was collected during the experiments are provided on FigShare, along with calibration footage and other files.  These are intended to help reproduce the results of the dash camera experiments, however a full process is provided for anyone who wishes to assemble their own equipment and collect data in a different area.

\section{Jupyter Notebooks}
\label{aj}

All Jupyter Notebooks listed below follow a convention that immediately after a title and description, there is a single ``Config'' cell where \textit{all} customization or configuration for the notebook can be made.  E.g. there may be a parameter that can be adjusted to survey a different town or suburb.  Shortly after the ``Config'' cell, any required Python modules are imported, and then then any variables that are derived from the ``Config'' variables (e.g. paths to files or directories) are defined.  To run the Jupyter Notebooks in general:

\begin{itemize}
\item{Review the description and documentation}
\item{Review and update the ``Config'' cell}
\item{Run all cells in the notebook, or run the cells one at a time if specifically directed by the notebook}	
\end{itemize}
Wherever an operation is expected to take a long time, progress bars are provided.

The notebooks are designed with the assumption that the ``jupyter notebooks'' command will be run from the ``minor\_thesis'' top-level directory that was checked out from GitHub as the current working directory.  They expect to find a ``data\_sources'' subdirectory for a lot of the data and configuration files, and a ``jupyter'' subdirectory where the notebooks live.

\subsection{01\_Reverse\_Geocode\_PBN}
\label{aj01}

We create a dataset of Google Street View images containing bicycle lane markings by examining images of intersections along known bicycle routes, because the markings are consistently found at intersections.  To identify candidate intersections to survey for sample data, we can use either the ``Principal Bicycle Network'' dataset or OpenStreetMap ``cycleway'' tags as our source of known bicycle routes.

If we wish to use the ``Principal Bicycle Network'' dataset as the basis for our sampling strategy, we use notebooks 01 and 02.  If we wish to use OpenStreetMap data, we can use notebook 03 instead.

The purpose of notebook 01 is to apply ``reverse-geocoding'' to all points in the ``Principal Bicycle Network'' dataset, to help identify intersections where we might look to gather Google Street View images containing bicycle lane markings, for our dataset.  The process takes latitude/longitude coordinates, and finds the street name and town/suburb/city, which we can use to search for intersections in the next step.

It requires a copy of the ``Principal Bicycle Network'' dataset from \url{data.gov.au} in geojson format, and connection details for a local Nominatim server as per appendix \ref{a:nominatim}.

It will create the file ``pbn\_exploded.geojson'' as its output.  This notebook can be skipped by using the copy provided in the archive on FigShare, or by using OpenStreetMap as the driver for sampling intersections along known bicycle lane routes, via notebook 03.

\subsection{02\_Identify\_Candidate\_Intersections\_PBN}
\label{aj02}

This notebook takes the reverse-geocoded ``Principal Bicycle Network'' data from notebook 01, and correlates it to OpenSteetMap data to find the intersections.  It outputs a file ``pbn\_intersections.csv'' for use in the sampling process in notebook 04.

This notebook requires a local Nominatim server.  It also requires an OpenStreetMap XML extract for Victoria -- see the notes at the top of the notebook for instructions on how to obtain that.  It can be skipped by the copy of ``pbn\_intersections.csv'' provided in the archive on FigShare, or by using OpenStreetMap as the driver for sampling intersections along known bicycle lane routes, via notebook 03.

\subsection{03\_Identify\_Candidate\_Intersections\_OSM}
\label{aj03}

This notebook creates a list of intersections along known bicycle lane routes using OpenStreetMap ``cycleway'' tags as its source of information.  It is an alternative to using notebooks 01 and 02 that should generalise well for other states and countries, where the ``Principal Bicycle Network'' dataset does not provide coverage.

It requires an OpenStreetMap XML extract for Victoria.  See the notes at the top of the Jupyter Notebook for instructions on how to obtain that.   It outputs a file ``osm\_intersections.csv'', which can also be found in the archive on FigShare.

This notebook is provided as an option to help reproduce the results outside of Victoria, however it was the ``Principal Bicycle Network'' approach in notebooks 01 and 02 that was actually used.

\subsection{04\_Gather\_Dataset\_GSV}
\label{aj04}

This notebook provides a GUI for sampling Google Street View images at intersections along known bicycle routes (the output of notebook 02 or 03).

Update the ``Config'' to give it the name of the file containing a list of locations to sample from, usually ``pbn\_intersections.csv'' or ``osm\_intersections.csv''.  Then run it from top to bottom.  The last cell contains a GUI using ``ipywidgets''.

You will need to provide a Google Street View API key that is linked to your Google Account and billing info, as per \url{https://developers.google.com/maps/documentation/streetview/get-api-key}.  The API key is stored in a text file ``apikey.txt'' in the parent directory of the directory from which Jupyter Notebooks was run.  (The directory that also contains a copy of the ``minor\_thesis'' directory that was checked out from GitHub.)

You may need to enable ipywidgets in Jupyter Notebooks before running the ``jupyter notebooks'' command.  Please find instructions on the commands to do that online, they depend on whether Anaconda is being used, or MacOS, etc.

\begin{figure}[h]
\centering
\includegraphics[scale=0.25]{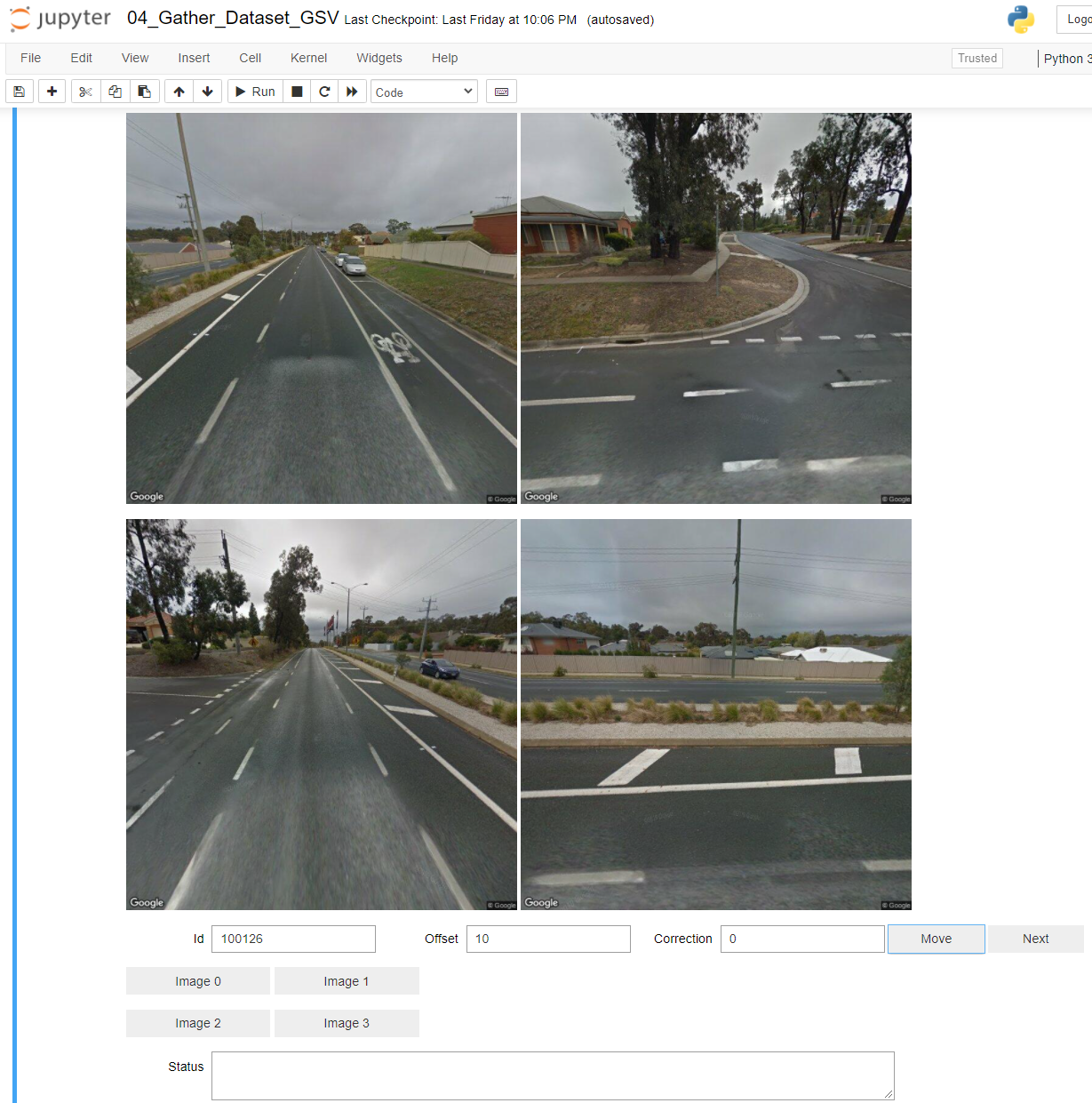}
\caption{GUI for gathering Google Street View image dataset}
\label{fig:gui}
\end{figure}

See figure \ref{fig:gui} for a screenshot of the GUI.

Four images are arranged on screen, with corresponding buttons ``Image 0'' through ``Image 3''.  The top-left image is ``Image 0'' and represents the forward-facing view at the sample location.  The bottom-right is ``Image 2'' and represents the rear-facing view.  The other two images show what appeared alongside the camera vehicle.

The GUI will automatically open with the first randomly-selected sample point.  Examine the four images, and for each image where a bicycle lane marking appears, press the corresponding button to record the ``hit'' in the output CSV ``hits.csv''.  This will mark the image for inclusion in the dataset in the next stage of the process, when images are ``labelled''.  In the example in figure \ref{fig:gui}, a bicycle lane marking is clearly visible in the top-left image, and the operator would hit the ``Image 0'' button to record it.

By default, the ``Offset'' and ``Correction'' text boxes are set to zero.

If you see a bicycle lane marking off in the distance in the forward direction, and you want to move the camera closer to it, enter ``10'' in the ``Offset'' text box, and then press ``Move''.  The ``Status'' box will briefly mention that the next image is being loaded, and then the images will update to a camera position 10 metres down the road.  To move in the reverse direction, enter a negative offset.

Moving forward or backward down the road depends on having the camera set at the correct heading for the road in the first place.  If the camera appears to have been placed so that the ``forward'' direction is at an angle to the road, correct it by entering  a number of degrees in the ``Correction'' text box, and then pressing ``Move''.  The orientation will shift.  Once you are happy the orientation is roughly aligned with the road, you can then enter an offset to move the camera to a better position.

When you are finished trying to obtain sample images from this location, press the ``Next'' button to move to a new sample location.

With practice, multiple sample images can be obtained per minute.  If there is no sign of any bicycle lane marking in the area, even in the distance, then you can quickly skip to the next location by pressing ``Next''.  It is not uncommon for the ``Principal Bicycle Network'' dataset to flag a route in a country town such as Warrnambool as an existing on-road bicycle route, only to find that all Google Street View images appear to show an informal paved shoulder, with no bicycle lane markings anywhere along the road.  If you can see bicycle lane marking in the immediate area, hit a button for each matching image and move on.  If you can see one off in the distance, move the camera 10 or 20 metres to capture it.

Every set of images that is downloaded from Google Street View will be cached in the ``gsv'' directory under ``data\_sources''.

The output of this notebook is a ``hits.csv'' file in ``data\_sources'' containing a list of cached Google Street View images where suitable bicycle lane markings were found.  You can monitor the size of ``hits.csv'' as you go along, to keep track of how many images you have found for your dataset.  Once you have enough images, proceed to notebook 05 for the labelling process.

\subsection{05\_Copy\_GSV\_Images\_For\_Labelling}
\label{aj05}

In notebook 05, we take the Google Street View images previously identified, copy them to a single folder, label them with the ``labelImg'' tool, and split them into ``training'' and ``test'' folders.

The key ``Config'' items are a version suffix to give to the dataset, e.g. ``V1'', and the percentage of the dataset that should be held in reserve for ``testing'' of models and not included in the ``training'' data, e.g. ``20'' for 20\%.

Do not run all cells of this notebook at once.  You will need to pause in the middle to run the external ``labelImg'' tool, from \url{https://github.com/tzutalin/labelImg}, or a suitable alternative.

In the first part of the notebook, the code will take every image that was flagged in ``hits.csv'' during the notebook 04 process, and copy them all into a single ``dataset'' folder.

After those cells have run, the notebook will tell you to stop and run the ``labelImg'' tool to ``label'' the data.  This involves telling ``labelImg'' where to find the new ``dataset'' folder with all the images, and then, one image at a time, you draw a bounding box around any object of interest, and give it a label, e.g. ``BikeLaneMarker'' for the bicycle lane markings we are training the model to look for.  See the instructions in the notebook and the official documentation for ``labelImg'' for more details.

Once the labelling is done, you can proceed with running the final cell in the notebook, which will randomly select a percentage of the labelled images in the ``dataset'' file and move them to a ``test'' directory, with the version suffix specified in the ``Config''.  The remainder will be moved to an equivalent ``train'' directory.  The labelled images in the ``train'' and ``test'' directories are ready to be turned into ``tfrecord'' files suitable for TensorFlow processing in the next notebook.

\subsection{06\_Model\_Training\_Setup}
\label{aj06}

The purpose of notebook 06 is to download and compile the required code to support TensorFlow Object Garden development, and to set up ``tfrecord'' files for training and validation of the models.

This notebook is worth running one cell at a time, to watch for errors about missing dependencies, and address them with ``conda install'' or ``pip install'' as required.

Cells 4 and 5 will download the required TensorFlow Object Garden code from GitHub, and then run the required ``protoc'' commands to set it up.

Cell 6 will run a validation script to check that TensorFlow development has been set up correctly.  Not all errors will necessarily mean that the setup is broken, but watch its output to make sure you are not missing any dependencies.

The final two cells will create a ``label map'' file, which must include all the labels that were used in ``labelImg'', and then convert the image and label data in the ``train'' and ``test'' dataset directories into ``tfrecord'' format, for use with TensorFlow.

\subsection{07\_Model\_Training\_GSV}

Notebook 07 allows you to download a pre-trained model from the TensorFlow 2 Object Garden, and configure it for training on the custom dataset.

\url{https://github.com/tensorflow/models/blob/master/research/object_detection/g3doc/tf2_detection_zoo.md}

In our case, we ran this notebook to configure multiple candidate models from the garden, before settling on the ``CenterNet HourGlass104 512x512'' model based on its performance.

In the ``Config'' section, give the URL and Name of the pre-trained model you wish to download, from the table on the above ``TensorFlow 2 Detection Zoo'' webpage.  You can also specify the name you'd like to give to a model when it has been trained and you want to create a ``frozen'' model from it, e.g. ``centernet\_V1''.

Then run the notebook from top to bottom.  The notebook will download the model, and update its configuration to look for our custom objects, instead of the ``COCO 17'' objects it was originally trained to detect.  In the results of the very last cell, it will print three commands that you can run manually, from a command prompt.  It is possible to update the notebook to run the commands from the notebook, but generally you get better, immediate feedback about the progress if you run them yourself from a command prompt.

Run the first command to train the model on our ``train'' dataset for 2000 steps.  Every 100 steps, it will give you statistics how long it is taking to perform each step, on average, and ``loss'' statistics to show how well it thinks it is performing with the ``train'' data.  You can re-run this command with a higher number of training steps to continue the training from where it left off.

The second command allows you to evaluate how the model performs with the ``test'' data that was withheld from training.  It can give you statistics such as ``mean Average Precision'' (mAP) and ``Total Loss''.  How did its predictions compare to the actual labelled bounding boxes in the test data?

The third command allows you to make a smaller ``frozen'' copy of the model.  A ``frozen'' model is useful if you want to:

\begin{itemize}
\item{Keep a smaller copy of the model as-at a certain number of training steps}
\item{Copy the model to a different system where it will be applied}
\end{itemize}

Use notebook 07 to download a pre-trained model and set it up for training, then run the training commands recommended by it until you are happy with the evaluation statistics and think it's ready to apply.  If the model does not perform well, you might need to review the dataset (possibly adding more images) or increase the number of training steps (if that is improving the performance) or switch to a different pre-trained model from the ``zoo''.

The next notebook can help show how the model is performing beyond the statistics, and give insight into what it is struggling with.

The frozen models that were produced by this notebook and then used to produce the results in the report have been included in the archive on FigShare.

\subsection{08\_Apply\_Model\_to\_Train\_and\_Test\_GSV}
\label{a08}

The ``evaluation'' script in notebook 07 will take a model and apply it to the ``test'' data, and produce performance statistics as output.  If you want to \textit{see} how it is performing, notebook 08 can help.  It processes the data from the ``train'' and ``test'' directories, and outputs images that have been sorted into ``hits'' and ``miss'' directories depending on whether a bicycle lane marking was detected, and it will draw a bounding box and confidence score around any detection.

In the ``Config'' section, give it the name of the model, which will either be the ``pretrained\_model\_name'' from notebook 07, or the ``frozen\_name'' if you created a ``frozen'' copy of the model.  Give it the version suffix of the dataset, e.g. ``V1''.  And then the ``confidence threshold'' at which the model should assume that an object has been detected, e.g. ``0.55'' means it will score a ``hit'' if its confidence is 0.55 or greater.

Run it from top to bottom, and it will produce output in \url{detections/train_XXX} and \url{detecions/test_XXX} where \url{XXX} is the version suffix of the dataset.  Within those directories, there will be a \url{hits} folder and a \url{miss} folder, containing output images.  You can review the \url{hits} to see if it accurately placed bounding boxes around the detect objects.  You can review the \url{miss} folder to see the images where it failed to make a detection, where one was expected.  Were there any unusual challenges with the image?  Is it reasonable that the object might not have been clearly visible?  Would it help to add more images like this to the dataset?

\subsection{09\_Filter\_OSM\_to\_Local\_Area}
\label{a09}

Notebooks 01 to 08 prepare and train a model to detect bicycle lane markings.  Next, we need to choose a ``survey area''.  We will use OpenStreetMap data for that area to identify locations to sample -- near each intersection -- and then we will download the images and process them with our model.

The purpose of Notebook 09 is to create a geojson file describing the shape of a local area, such as a town or suburb, based on government-issued data.  We can then use this to extract OpenStreetMap data for that specific area, and use it as our ``survey area''.

In the ``Config'' section, choose a suburb name as the ``locality\_name''.  It will expect to load the government-issued ``LGA\_boundaries\_VIC.geojson'' file from \url{data.gov.au} and reduce it down to a geojson name for the selected suburb.

Download the OpenStreetMap data for Australia from \url{download.geofabrik.de} as instructed in the notebook.

If you run the notebook from top-to-bottom, it will output two suggested ``osmium'' commands.  These are designed to reduce the full ``country-level'' extract of OpenStreetMap data down to data for the survey area.  The first command will reduce the data to the exact official shape of the suburb.  The second command will create an extract for a slightly bigger bounding box, to help identify intersections near the margin on the survey area.

Install the ``osmium'' tool and run the commands suggested by the notebook to create the required OpenStreetMap extracts.

OpenStreetMap extracts that were produced for the survey areas using this process have been included in the archive on FigShare.

\subsection{10\_Apply\_Model\_to\_Survey\_Area\_GSV\_Images}
\label{a10}

This notebook will identify sample points near intersections in the survey area, download Google Street View images, apply the model, and record any detected bicycle lane markings in the output ``detection\_log.csv''.

If you wish to control Google Street View API costs, please run this notebook one cell at a time.

In the first phase, it will use the OpenStreetMap extracts for the survey area, from notebook 09, to create a list of sample points within a certain number of metres of each intersection.  The ``margin'' parameter lets you specify how many metres either side of each intersection you want to sample.  If it is set to 0, the process will only take samples from the middle of each intersection.  If it is set to 10, it will also take samples 10 metres either side of the intersection.  The recommended setting is 20, because the Australian standards specify that a bicycle lane marking should be within 15 metres of every intersection, and the Google Street View API apparently only gives a distinct image every 10 metres.

At the end of the first phase, the notebook will report how many sample points were identified for margins of 0m, 10m and 20m.  The process will be taking four images per sample point, so multiply the proposed number of sample points by four to get an estimate of the cost.  If there are too many sample points, consider using the ``osmium'' tool with its ``--bbox'' option to extract a smaller rectangular area within the suburb as a sample, instead of processing the whole town.

If you proceed with the rest of the notebook with the proposed sample points, it will download the required images from Google Street View, process them with the model, record the detections to a subdirectory for the ``locality\_name'' within the ``detections'' directory, and create geojson files to be used with the maps in notebook 11.  It will infer detected routes from the detection locations according to some rules, and compare these to what OpenStreetMap said with its ``cycleway'' tags.  As part of the comparison, it will tell you how many metres of bicycle lane routes were detected, how many metres were tagged in OpenStreetMap, how many metres they agree, and how many metres each source had a route that the other one did not.

The ``detections'' folder for the ``locality\_name'' will also have ``hits'' and ``miss'' folders where output images (with bounding box overlays) can be examined.  It can be helpful to quickly flick through the images in the ``miss'' folder to see if there are many clear bicycle lane markings that were missed.  It can also be helpful to flick through the ``hits'' folder to see if there were many false positives, and what might have been causing them.

The results of applying this notebook to the survey areas in Mount Eliza and Heidelberg have been included in the archive on FigShare, excluding the large ``miss'' folders.  To understand which points were sampled, see the ``batch'' file for the survey area in the archive.  The ``detection\_log.csv'' file records a summary of images where a bicycle lane marking was detected, and the ``hits'' folder provides images with detection overlays.

\subsection{11\_Map\_Bicycle\_Lanes\_GSV}
\label{a11}

Notebook 11 produces interactive maps to view and compare bicycle lane routes for the survey area:

\begin{itemize}
\item{Map 1 shows each detection as a point on the map}
\item{Map 2 shows the bicycle lane routes that were inferred from the detections}
\item{Map 3 shows bicycle lane routes that are tagged as ``cycleway'' routes in OpenStreetMap}
\item{Map 4 shows a comparison between the detected routes and OpenStreetMap, with different colours to represent where they agree, and where one had a route section that the other did not.}	
\end{itemize}

Set the survey area name in the ``Config'' section, and run the notebook from top to bottom.  You can zoom in and out and pan around each map as required.  The ``Config'' section allows you to set a default zoom level.

The geojson files that were produced for the survey areas are included in the archive on FigShare.  Therefore, the maps can be examined directly by setting up an environment with the latest code for GitHub and archive data from FigShare, and running this notebook.

\subsection{12\_Split\_Dashcam\_Footage}
\label{a12}

The next group of notebooks are dedicated to detecting bicycle lane routes from dash camera footage instead of Google Street View images.

Notebook 12 expects to find a folder within ``data\_sources'' containing MP4 video files and associated NMEA files.  These came from a ``Navman MiVue 1100 Sensor XL DC Dual Dash Cam'' that records one MP4 video and one NMEA file per minute.

In the ``Config'' section, specify the name of the folder containing the footage, this is effectively the ``survey area''.  You can also specify how many image frames you want to sample, up to the number of frames in the original footage.  E.g. you can reduce 60fps footage down to 5 frames per second.

Run this notebook from top to bottom.  It will split the video files into images on disk, and create a file ``metadata.csv'' with the location for each image based on the NMEA data.

The first progress bar shows how many of the video files have been processed.  Subsequent progress bars show progress through each individual video file as it loads.

\subsection{13\_Copy\_Dashcam\_Images\_For\_Labelling}
\label{a13}

The first time you are processing dashcam footage, it is recommended that you skip straight to notebook 15 and attempt to detect bicycle lane markings with an existing model that was trained with Google Street View images.  That model will not necessarily have ideal performance on the dashcam images, which are taken from a different perspective, with different equipment, at a different resolution.  But running a first pass with the GSV model can help to quickly identify images that could be taken from the dashcam footage and added to the training and validation dataset.

Use notebook 15 to process footage from a ``training'' area.  Then come back to this notebook 13.

In the ``Config'' directory, set the version suffix for the previous GSV dataset that you wish to add to, e.g. ``V1''.  Then set the version suffix for the new dataset version you wish to create e.g. ``V2''.  Then start running the first few cells.

Any ``hits'' from the initial training -- whether they are true positives or false positives -- will be read from the ``detection\_log.csv'' file by notebook 13, and copied to a ``dataset'' folder.

Stop running the notebook, and label the images in the ``dataset'' folder with ``labelImg'' as directed.  To avoid false positives, it was found to be helpful to label additional classes such as turning arrows, traffic islands, and give way markings, to avoid confusing their simple white markings with bicycle lane markings.  See the notebook instructions for further details.

Once the images in the ``dataset'' directory have been labelled, continue running the notebook.  Existing ``train'' and ``test'' images from the previous dataset will be copied into this new version, according to their original splits.  Then any new images from the dashcam will be split into the ``train'' and ``test'' directories to join them.  Finally, ``tfrecord'' files will be created from them, for TensorFlow to use.

\subsection{14\_Model\_Training\_Dashcam}
\label{a14}

Once new ``train'' and ``test'' datasets have been constructed, with dashcam images supplementing the original Google Street View images, we can train a new model.

This notebook 14 is equivalent to notebook 07, but for the new dataset that includes dashcam images.  We select a pre-trained model that we want to work from, download it, and run the recommended training, evaluation, and model-freezing commands.

Beware that if you give exactly the same ``pretrained\_model\_name'' as you did for GSV training, it will try to continue training that model where you left off.  You may prefer to give a different ``pretrained\_model\_name'' to start the training from scratch.

The frozen model that was used to survey dash camera images in Mount Eliza has been included in the archive on FigShare.

\subsection{15\_Apply\_Model\_to\_Train\_and\_Test\_Dashcam}
\label{a15}

This notebook 15 is equivalent to notebook 08, but for the new dataset that includes dashcam images.  Run this notebook if you want to visualize the results of applying the model-in-training to the ``train'' and ``test'' datasets.

\subsection{16\_Apply\_Model\_to\_Dashcam\_Footage}
\label{a16}

This notebook 16 is equivalent to notebook 09, instead of sampling Google Street View images near intersections, we are processing every image we extracted from the video footage in notebook 12.

Based on the findings from cycling the ``training area'' footage through notebook 15 $\rightarrow$ 13 $\rightarrow$ 14 in a loop, some enhancements were made to the model to reduce false positives.  See the notebook and ``Methods'' section of this document for further details.

In the ``Config'' section, you can specify the folder/location being surveyed, and the minimum confidence score required to flag a detection.  Specify the name of the model or frozen model you wish to use, and the dataset version prefix to ensure the correct label map file is used.

The ``mask'' option allows you to specify which part of the frame to run detections on, instead of the whole frame.  This will exclude detections in areas where we do not expect to see a bicycle lane marking, and avoid some potential false positives.  The mask boundary will be drawn on the output images in the ``hits'' and ``miss'' folders, so you can see what was considered by the detection model.

There are some further options to control how many detections are required in a general area before a detection is counted for the purposes of inferring a route.  This is to avoid false positives that occur in a single frame with nothing detected in adjacent frames.  By default, each detection requires two further detections within 50 metres, but a minimum of 10 metres away to avoid duplication when the camera is motionless.

The end result of this notebook is a ``detection\_log.csv'' with all detection points, and ``detection\_log\_filtered.csv'' where one-off detection points that are not supported by nearby detections are excluded.

A copy of the output of this process for the Mount Eliza test area has been included in the archive on FigShare, excluding the ``miss'' folder.

\subsection{17\_Convert\_Detection\_Log\_to\_GeoJSON}
\label{a17}

In notebook 17, we take the ``detection\_log\_filtered.csv'' output from notebook 16, and correlate it to the OpenStreetMap data, infer bicycle lane routes, map them, and compare to the routes that are tagged as ``cycleway'' in the OpenStreetMap data.  Differences and agreement between routes are measured in metres, and geojson files are created in order to produce maps in the next notebook.

Check the ``Config'' section to make sure the correct dashcam image folder has been specified for the survey area, along with the ``locality'' name for the OpenStreetMap extract.  Then run the notebook from top to bottom.

\subsection{18\_Map\_Bicycle\_Lanes\_Dashcam}
\label{a18}

This notebook 18 is equivalent to notebook 11, except it is producing maps to view and compare routes that were detected from dash camera footage, instead of Google Street View images.

\begin{itemize}
\item{Map 1 shows the bicycle lane routes that were inferred from the detections}
\item{Map 2 shows bicycle lane routes that are tagged as ``cycleway'' routes in OpenStreetMap, where the route was covered by the dash camera footage}
\item{Map 3 shows a comparison between the detected routes and OpenStreetMap, with different colours to represent where they agree, and where one had a route section that the other did not.}	
\end{itemize}

The geojson files that were generated for the Mount Eliza test area from dash camera footage have been included in the archive on FigShare.  Therefore, the maps can be examined directly by setting up an environment with the latest code for GitHub and archive data from FigShare, and running this notebook.

\subsection{19\_Calibrate\_Dashcam}
\label{a19}

The final group of notebooks were used to demonstrate how the general technique of applying a model to images and correlating the result to existing geospatial data could be used to build a dataset about road infrastructure.  In this example, we looked at whether we could detect a ``paved shoulder'' to the left of the camera vehicle's lane, where a cyclist might ride their bike with some separation from other vehicles.  Future work could be looking at estimating lane widths, detecting parked cars or other obstacles in the bicycle lane, or detecting hazardous road surface defects.

Notebook 19 is used to calibrate a model to correct images from the dash camera to account for optical distortion.  Follow the instructions in the notebook (and the reference material linked in the notebook description) to print out a calibration tool that looks like a chessboard.  Record footage of this calibration tool being held up in front of the dash camera at different positions in the frame, at different distances from the camera.

Update the ``Config'' to point to the folder where the calibration videos can be found, and record the number and size of the squares as printed on the calibration tool.  Then run the notebook.

The notebook will split the video into images at 1 frame per second, then follow a standard OpenCV calibration process to produce a model.  The model will then be saved as a ``dashcam\_calibration.yml'' file in the ``data\_sources'' directory.

At the bottom of the notebook, you will see an example of an original image from the dashcam, and a corrected image where the model has been applied to make straight lines appear straighter.

\subsection{20\_Detect\_Paved\_Shoulders\_Dashcam}
\label{a20}

This notebook will re-process the video footage from a survey area, just as notebook 12 did.  But this time:

\begin{itemize}
\item{The model to correct for optical distortion, from notebook 19, will be applied, then...}
\item{A ``lane detection'' model will be applied, to detect the camera's own lane and then the next lane to the left, if any}
\item{The detected lane lines will be overlaid onto output images so that they can be visualized}
\item{For each road segment, from intersection to intersection, an assessment is made as to whether there seems to be a paved shoulder, based on whether one was found in most images, whether it was wide enough, and whether the boundaries were stable enough across the frames in the group.  See the ``Methods'' section and the notebook for further explanation.}	
\end{itemize}

The first significant output of this process is the ``metadata\_with\_summary.csv'' file, which is recorded in the ``split'' subdirectory under the footage folder.  It contains one record per frame, with the summary statistics for the road segment that are used to decide whether there might be a paved shoulder along that segment, or not.

Then, a geojson file is created, to allow us to draw on a map where the paved shoulder detection criteria was met in ``metadata\_with\_summary.csv''.  This is mapped in notebook 21.

\subsection{21\_Map\_Paved\_Shoulders\_Dashcam}
\label{a21}

This notebook draws the detected ``paved shoulders'' from notebook 20 on a map, so that they can be visualized.  Most bicycle lanes should also appear as paved shoulders, so it is useful to compare this output to the other maps in notebooks 18 and 11.

\subsection{22\_Apply\_Model\_To\_Video\_Stream}
\label{a22}

This notebook is used to create demonstration videos where a dash camera video is loaded, a model is applied to all frames, and then an output video is created with a detection overlay.  Sample outputs are available in the FigShare archive under the ``demos'' directory.

\chapter{OpenStreetMap XML Concepts}
\label{a:osm_concepts}

The use of OpenStreetMap XML extract data is fundamental to the solutions proposed in this research project.  In this appendix, we briefly describe some key concepts in the data, and provide example XML data to demonstrate them.

\section{Ways}
\label{osm:ways}

A ``way'' in the OpenStreetMap XML data is a line that can be drawn on a map.  Typically, it will represent a road segment, as per the example in figure \ref{xml:way_st}.  However, it could also be the path of a natural feature such as a creek or coastline.  It could be an off-street walking track or bicycle trail.  Or it could be the boundary of a reserve or an estate.

\begin{figure}[t]
\centering
\begin{verbatim}
  <way id="26662301" version="32" timestamp="2020-08-27T04:26:02Z">
    <nd ref="30204323"/>
    <nd ref="638346068"/>
    ...
    <nd ref="2117131454"/>
    <nd ref="638346153"/>
    <tag k="cycleway:left" v="shared_lane"/>
    <tag k="highway" v="tertiary"/>
    <tag k="maxspeed" v="60"/>
    <tag k="name" v="Humphries Road"/>
    <tag k="sidewalk" v="right"/>
    <tag k="surface" v="asphalt"/>
  </way>
\end{verbatim}
\caption{Sample OpenStreetMap XML ``way''}
\label{xml:way_st}
\end{figure}

Each ``way'' has a unique ``id'', followed by a list of ``node references'' and ``tags''.

The ``node reference'' links to a ``node'', which has a latitude/longitude position.  Therefore, the node reference list effectively describes path of the ``way'' if it were drawn on a map.

The ``tags'' describe characteristics of the way.  In figure \ref{xml:way_st}, the ``way'' is a ``tertiary'' road named ``Humphries Road'' where the speed limit is 60 (kmph) and there is a bicycle lane and a sidewalk.

A ``way'' can be only a \textit{part} of a longer road.  If different sections of a road have different properties, the road will be broken up into multiple ``ways''.  E.g. a change of speed limit would split the road into two ``ways''.

\clearpage
\section{Nodes}
\label{osm:nodes}

In a separate section of the OpenStreetMap XML file, each of the ``nodes'' are listed.  A ``node'' is a point on a map, usually just with a latitude and longitude, but occasionally they can have tags to describe special properties, such as the presence of traffic signals as in figure \ref{xml:way_nodes}.\\

\begin{figure}[h]
\centering
\begin{verbatim}
  <node id="30204322" version="18" timestamp="..." lat="-38.1655191" lon="145.1016428"/>
  <node id="30204323" version="21" timestamp="..." lat="-38.1667063" lon="145.1017474">
    <tag k="highway" v="traffic_signals"/>
  </node>
  <node id="30204324" version="18" timestamp="..." lat="-38.1674697" lon="145.101785"/>	
\end{verbatim}
\caption{Sample OpenStreetMap XML ``nodes''}
\label{xml:way_nodes}
\end{figure}

\section{Intersections}
\label{osm:intersections}

If a ``node'' is shared by two more ``ways'', it may be an intersection.  By checking the name of each ``way'' we can see whether it is an intersection, or just a ``node'' at which single road was split into two ways due to a change in a characteristic.

Therefore, to find ``nodes'' that are intersections, look for ``nodes'' that are shared by ``ways'' with multiple distinct names.

\section{Cycleways}
\label{osm:cycleways}

For the purposes of this project, if a ``way'' has a tag that begins with ``cycleway'', then OpenStreetMap believes that it has some sort of bicycle lane on at least one side of the road.  This was consistent with how real bicycle lane routes were tagged in the survey areas.  The ``bicycle'' tag typically appears to mean that bicycles are \textit{permitted}, but it may not mean that there is a dedicated bicycle lane.

OpenStreetMap relies on crowdsourced data, and contributors might not always follow the same conventions on how to use each tag.

\chapter{Computer Environment}
\label{a:environment}

\section{Machine Learning workstation}
\label{a:computer}

This workstation was used for all machine learning activities.  TensorFlow 2.3 GPU acceleration was enabled via NVIDA cuDNN and CUDA drivers.

\begin{itemize}
\item{Windows 10}
\item{Anaconda 3}
\item{Intel(R) Core(TM) i9-9900K CPU @ 3.60GHz}	
\item{32GB RAM}
\item{NVIDIA GeForce RTX 2080 Ti (12288MB dedicated, 16341MB shared}
\item{1TB SSD}
\end{itemize}

Details of the Anaconda environment, including the packages and versions used, can be found in the GitHub repository in the ``environment.yml'' file.

\section{Nominatim server}
\label{a:nominatim}

This Linux server ran on a separate machine, as a virtual host.  Its sole purpose was to run a Nominatim server for the reverse-geocoding and geocoding services required by some Jupyter notebooks.

\begin{itemize}
\item{Ubuntu 20.04 running on Windows 10 under Hyper-V}
\item{AMD Ryzen 7 3700X 8-core processor 3.6GHz}
\item{16GB (32 GB for windows host)}
\item{100MB SSD}
\end{itemize}

The server was loaded with OpenStreetMap data for Australia sourced from the GeoFabrik website \cite{geofabrik} according to the Nominatim installation instructions \cite{nominatim_install}.


\clearpage
\bibliographystyle{IEEEtran}
\bibliography{references.bib}
\end{document}